\documentclass[preprint,5p]{elsarticle}

\usepackage{siunitx}
\usepackage{float}
\usepackage{subcaption}
\usepackage[utf8]{inputenc}
\usepackage{hyperref}
\usepackage{amsmath}
\usepackage{orcidlink}
\usepackage{bibentry}
\usepackage{lineno}

\usepackage{acro}
\acsetup{list/template=longtable, case-sensitive=false, use-id-as-short=true}

\DeclareAcronym{dl}{
  short = DL ,
  long  = Deep Learning ,
}
\DeclareAcronym{ml}{
  short = ML ,
  long  = Machine Learning ,
}
\DeclareAcronym{ai}{
  short = AI ,
  long  = Artificial Intelligence ,
}
\DeclareAcronym{ssd}{
  short = SSD ,
  long  = Single-Shot Multibox Detector ,
}
\DeclareAcronym{fpn}{
  short = FPN ,
  long  = Feature Pyramid Network ,
}
\DeclareAcronym{fpga}{
  short = FPGA ,
  long  = Field Programmable Gate Array ,
}
\DeclareAcronym{gpu}{
  short = GPU ,
  long  = Graphical Processing Unit ,
}
\DeclareAcronym{cpu}{
  short = CPU ,
  long  = Central Processing Unit ,
}
\DeclareAcronym{asic}{
  short = ASIC ,
  long  = Application-Specific Integrated Circuit ,
}
\DeclareAcronym{tpu}{
  short = TPU ,
  long  = Tensor Processing Unit ,
}
\DeclareAcronym{ncs}{
  short = NCS ,
  long  = Neural Compute Stick ,
}
\DeclareAcronym{dpu}{
  short = DPU ,
  long  = Deep Learning Processor Unit ,
}
\DeclareAcronym{relu}{
  short = ReLU ,
  long  = Rectified Linear Unit ,
}
\DeclareAcronym{sgd}{
  short = SGD ,
  long  = Stochastic Gradient Descendent ,
}
\DeclareAcronym{int8}{
  short = INT8 ,
  long  = 8-bit integer ,
}
\DeclareAcronym{fp32}{
  short = FP32 ,
  long  = single-precision floating-point ,
}
\DeclareAcronym{fp16}{
  short = FP16 ,
  long  = half-precision floating-point ,
}
\DeclareAcronym{trt}{
  short = TRT ,
  long  = TensorRT ,
}
\DeclareAcronym{tf-trt}{
  short = TF-TRT ,
  long  = TensorFlow TensorRT ,
}
\DeclareAcronym{tflite}{
  short = TFLite ,
  long  = TensorFlow Lite ,
}
\DeclareAcronym{map}{
  short = mAP ,
  long  = mean Average Precision ,
}
\DeclareAcronym{iou}{
  short = IoU ,
  long  = Intersection over Union ,
}
\DeclareAcronym{tp}{
  short = TP ,
  long  = True Positive ,
}
\DeclareAcronym{tn}{
  short = TN ,
  long  = True Negatives ,
}
\DeclareAcronym{fp}{
  short = FP ,
  long  = False Positive ,
}
\DeclareAcronym{fn}{
  short = FN ,
  long  = False Negative ,
}
\DeclareAcronym{YOLO}{
  short = YOLO ,
  long  = You Only Look Once ,
}
\DeclareAcronym{cnn}{
  short = CNN ,
  long  = Convolutional Neural Network ,
}
\DeclareAcronym{vram}{
  short = VRAM ,
  long  = Video Random Access Memory ,
}
\DeclareAcronym{ann}{
  short = ANN ,
  long  = Artificial Neural Network ,
}
\DeclareAcronym{ps}{
  short = PS ,
  long  = Processing System ,
}
\DeclareAcronym{pl}{
  short = PL ,
  long  = Programmable Logic ,
}
\DeclareAcronym{som}{
  short = SoM ,
  long  = system on-module ,
}
\DeclareAcronym{fps}{
  short = FPS ,
  long  = frames per second ,
}
\DeclareAcronym{ip}{
  short = IP ,
  long  = Intellectual Property ,
}

\journal{Engineering Applications in Artificial Intelligence}

\begin{document}

\begin{frontmatter}

\title{Benchmarking Edge Computing Devices for Grape Bunches and Trunks Detection using Accelerated Object Detection Single Shot MultiBox Deep Learning Models\tnoteref{t1}}

\tnotetext[t1]{This manuscript is available in SienceDirect via \url{http://dx.doi.org/10.1016/j.engappai.2022.105604}}

\author[1,2]{Sandro Costa Magalhães\orcidlink{0000-0002-3095-197X}}\ead{sandro.a.magalhaes@inesctec.pt}\corref{corr1}

\author[2]{Filipe Neves Santos\orcidlink{0000-0002-8486-6113}}\ead{fbsantos@inestec.pt}

\author[3]{Pedro Machado\orcidlink{0000-0003-1760-3871}}\ead{pedro.machado@ntu.ac.uk}

\author[1,2]{António Paulo Moreira\orcidlink{0000-0001-8573-3147}}\ead{amoreira@fe.up.pt}

\author[4,5]{Jorge Dias\orcidlink{0000-0002-2725-8867}}\ead{jorge.dias@ku.ac.ae}

\cortext[corr1]{Corresponding author}

\affiliation[1]{organization={INESC TEC -- Instituto de Engenharia, Tecnologia e Ciencia}, addressline={Campus da FEUP, Rua Dr. Roberto Frias s/n}, city={Porto}, postcode={4200-465}, state={Porto}, country={Portugal}}

\affiliation[2]{organization={Faculty of Engineering, University of Porto}, addressline={Rua Dr. Roberto Frias s/n}, city={Porto}, postcode={4200-465}, state={Porto}, country={Portugal}}

\affiliation[3]{organization={Computational Intelligence and Applications group (CIA), Department of Computer Science, School of Science and Technology, Nottingham Trent University}, addressline={Clifton Campus}, city={Nottingham}, postcode={NG11 8NS}, country={United Kingdom}}

\affiliation[4]{organization={Khalifa University Center of Autonomous Robotics Systems (KUCARS), Khalifa University of Science, Technology and Research (KU)}, postcode={127788}, state={Abu Dhabi}, country={United Arab Emirates}}

\affiliation[5]{organization={Department of Electrical Engineering and Computers, University of Coimbra}, adressline={Rua Silvio Lima}, city={Coimbra}, postcode={3030-290}, state={Coimbra}, country={Portugal}}


\begin{abstract}
\textbf{Purpose:} Visual perception enables robots to perceive the environment. Visual data is processed using computer vision algorithms that are usually time-expensive and require powerful devices to process the visual data in real-time, which is unfeasible for open-field robots with limited energy. This work benchmarks the performance of different heterogeneous platforms for object detection in real-time. This research benchmarks three architectures: embedded GPU -- Graphical Processing Units (such as NVIDIA Jetson Nano \SI{2}{GB} and \SI{4}{GB}, and NVIDIA Jetson TX2), TPU -- Tensor Processing Unit (such as Coral Dev Board TPU), and DPU -- Deep Learning Processor Unit (such as in AMD-Xilinx ZCU104 Development Board, and AMD-Xilinx Kria KV260 Starter Kit).
\textbf{Method:} The authors used the RetinaNet ResNet-50 fine-tuned using the natural VineSet dataset. After the trained model was converted and compiled for target-specific hardware formats to improve the execution efficiency.
\textbf{Conclusions and Results:} The platforms were assessed in terms of performance of the evaluation metrics and efficiency (time of inference). \acp{gpu} were the slowest devices, running at \SIrange{3}{5}{FPS}, and \acp{fpga} were the fastest devices, running at \SIrange{14}{25}{FPS}. The efficiency of the \ac{tpu} is irrelevant and similar to NVIDIA Jetson TX2. \Ac{tpu} and \ac{gpu} are the most power-efficient, consuming about \SI{5}{\watt}. The performance differences, in the evaluation metrics, across devices are irrelevant and have an F1 of about \SI{70}{\percent} and \ac{map} of about \SI{60}{\percent}.
\end{abstract}

\begin{graphicalabstract}
\includegraphics{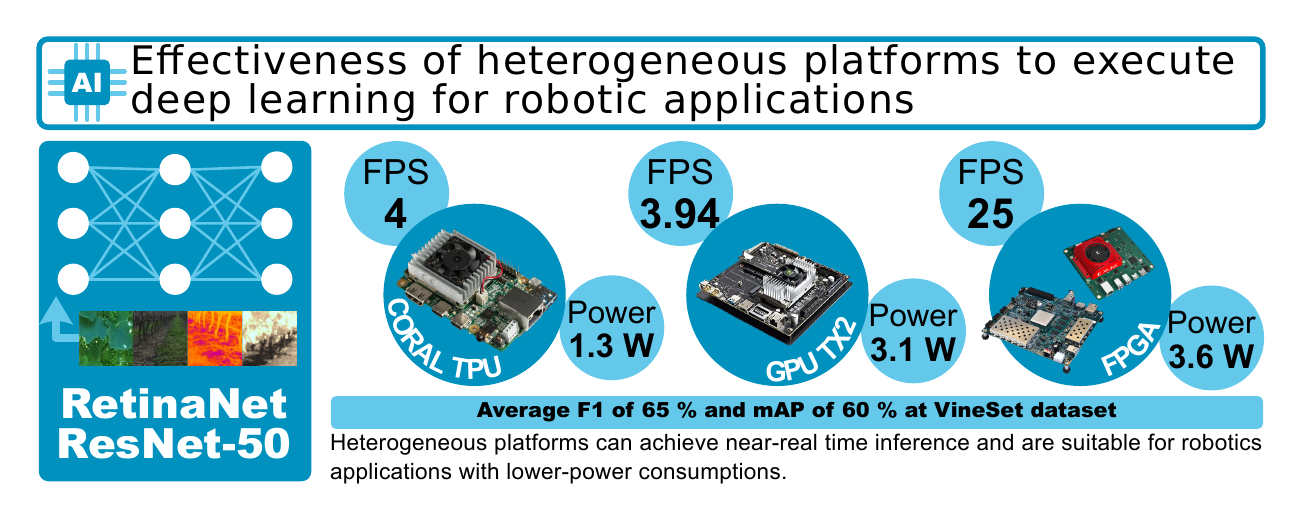}
\end{graphicalabstract}



\end{frontmatter}


\section{Introduction}\label{sec:introduction}

Computer vision classifiers are largely explored in multiple robotics systems, such as agricultural ones. These systems allow robots to perform visual localisation by visually detecting natural landmarks like tree trunks \cite{Mendes2016} or to detect objects for other purposes such as grasping or harvesting \cite{Magalhaes2021,Moreira2022}.

The rise of \acf{ai} and the continuous generation of big data is creating computational challenges. \acfp{cpu} are not enough to efficiently run state-of-the-art \ac{ai} algorithms or process all the data generated by a wide range of sensors. World-leading processing technology companies (such as NVIDIA, AMD, Intel and ARM) have been looking closely into the new requirements. They have been pushing the boundaries of technology to deliver efficient and flexible processing solutions. 

Heterogeneous computing refers to the use of different types of processor systems in a given scientific computing challenge. 

Heterogeneous platforms are composed of different types of computational units and technologies. Such media can be composed of multi-core \acp{cpu}, \acp{gpu} and \acp{fpga} acting as computational units and offering the flexibility and adaptability demanded by a wide range of application domains \cite{SicaDeAndrade2018}. These computational units can significantly increase the overall system efficiency and reduce power consumption by parallelising concurrent operations that require substantial \ac{cpu} resources over long periods. 

Accelerators like \acp{gpu} and \acp{fpga} are massive parallel processing systems that enable accelerating portions of code that are parallelisable. Combining \acp{cpu} with \acp{gpu} and \acp{fpga} help improve the efficiency (speed of executing algorithms) by assigning different computational tasks to specialised processing systems. \acp{gpu} are optimised to perform matrix multiplications in parallel, which is the major bottleneck in video processing and computer graphics. Nevertheless, \acp{gpu} also introduce hardware and environmental limitations (e.g. high-power consumption and architectural limitations) \cite{Intel2020b}. \acp{cnn} are massively parallel in their nature and not suitable for matrix representation because each neuron can be considered a node containing several sequential mathematics operations. Despite of very optimised to execute parallel operations, \acp{gpu} architecture is inspired by \ac{cpu}. \Acp{asic} are synthesized \acp{fpga}' designs that aim to optimise and specify the operations executions. \Acp{asic} are more compact and, if designed for processing \ac{cnn} algorithms, so fast as \acp{fpga}. \Acp{asic} can be designed to work as single devices or connected to external systems.


In \ac{dl} applied to visual problems, \acp{cnn} are the most common \acp{ann}. These networks' architecture is mainly composed of sequential convolution layers that are trained to extract relevant features from images. \Acp{cnn} are frequently applied to classification, object detection and segmentation problems. In the scope of object detection, the most used \acp{cnn} architectures are \ac{ssd} \cite{Liu2016}, faster R-\ac{cnn} \cite{Wang2019}, and \ac{yolo} \cite{Redmon2016,Redmon2018}. Faster R-\ac{cnn} is the most precise model to detect objects but processes the image in two stages, making inference slower. \ac{ssd} and \ac{yolo} are both single-shot architectures, i.e., they only process the image once using feature maps, repositioning the object bounding boxes, and making their classification. {Some authors have been exploring single-shot architectures to detect fruits and other objects in open-field environments \cite{Magalhaes2021,Sozzi2022,Zhao2022,Olenskyj2022,Terra2021,Magalhaes2022}. Inside this group of architectures, \ac{yolo} models are undoubtedly the most common deep neural network \cite{Magalhaes2021,Sozzi2022,Zhao2022,Olenskyj2022}. Because, They are fast and can achieve near real-time speed easily under regular computing hardware \cite{Zhao2022}, without big degradation of the metric when compared with other equivalent \acp{ann} \citet{Magalhaes2021}. However, they may have difficulty detecting some objects, which can be resolved by bigger and more capable \ac{cnn} architecture. Transformers are also an upcoming \ac{dl} architecture for object detection with successful results \cite{Olenskyj2022}. Despite this analysis, most authors benchmark their works against powerful and high-consuming hardware not suitable for embedded or robotics applications \cite{Magalhaes2021,Sozzi2022}.  }

For overcoming the restrictions of real-time classification and power consumption, many researchers have studying small-size and effective \ac{dl} architectures, like Tiny-\ac{YOLO} \cite{Redmon2016,Redmon2018}, YOLACT \cite{Bolya2019}, and many other architectures \cite{Howard2017,Sandler2018,Liu2016}, that can be implemented in more cost-effective \acp{gpu} or even in \acp{cpu}. Alternatively, other researchers are studying low-power and efficient devices that may run parallelisable deep neural networks \cite{Puchtler2020}. These devices are generically characterised as embedded devices and are from many types and architectures: \acp{gpu}, \acp{fpga}, and \acp{asic}, more commonly, Coral \acp{tpu} and Intel \acp{ncs} (see more details at section~\ref{sec:heterogeneous-platforms}).  Another common technique used by some researchers is quantisation \cite{Yang2019}. By default, \acfp{ann} are trained in \ac{fp32}, but the optimisation algorithms are iterative and often converge to high-resolution precise values that are time-consuming to compute and meaningless for the classification process. The quantisation technique allows reducing the \ac{ann} resolution to \ac{int8} by rescaling the \ac{fp32} weights, improving the time of inference and, sometimes, the accuracy. The merge of different strategies to optimise the execution of \acp{ann} can create highly efficient \ac{dl} models that can process images at thousands of \ac{fps}.

Researchers have essentially focused on embedded \acp{gpu} from the NVIDIA Jetson family, using NVIDIA Jetson Nano, NVIDIA Jetson TX2 and NVIDIA Jetson AGX Xavier. \citet{Zhao2019} benchmarks two \ac{dl} models, Tiny-\ac{YOLO} and DNET, under NVIDIA Jetson TX2 and NVIDIA GTX Titan X. The authors could have a low accuracy drop (about \SI{1}{\percent}) in the quantisation process for the NVIDIA Jetson TX2. The inference speed was about ten times slower in the NVIDIA Jetson TX2 (running at \SI{18}{FPS}), as expected, but consumed 20 times less power, consuming only about \SI{8}{\watt}. \citet{Suzen2020,Chiu2020,Rahmaniar2021,Martinez2021} also benchmark \ac{dl} models efficiency between NVIDIA Jetson embedded boards. The NVIDIA Jetson AGX Xavier was the fastest board in the family but also the most power-expensive. On the other side, the NVIDIA Jetson Nano is less power-consuming but slower. The most benchmarked \ac{dl} models are \ac{ssd} MobileNet networks family and \ac{YOLO} family. Both are small-size networks that have fewer convolution layers and retain fewer images' features. \citet{Martinez2021} run a YOLACT at \SI{66}{FPS} in an NVIDIA Jetson AGX Xavier and at \SI{16}{FPS} in an NVIDIA Jetson TX2, revealing the substantial hardware improvement of the most recent NVIDIA Jetson board. \citet{Chiu2020,Rahmaniar2021} benchmark \ac{SSD} MobileNet v2 in the three boards and NVIDIA Jetson TX2 was the fastest with \SI{26}{FPS}. \citet{Suzen2020} also benchmarked the Raspberry Pi4, but it was slow and inefficient.

Despite the reasonable power-consumption improvement, Jetson \acp{gpu} have a similar architecture to traditional NVIDIA \acp{gpu}, sharing some of their limitations. So, some researchers started exploring the highly efficient \acp{fpga}. The most commonly explored \acp{fpga} in the literature now belongs to AMD-Xilinx, particularly to the AMD-Xilinx Zynq family. \citet{Venieris2017,Chen2019} compared a Zynq \ac{fpga} against a \ac{GPU}. \citet{Venieris2017} benchmark multiple \acp{cnn} between Xilinx Zynq-7045 and a NVIDIA Tegra X1. In all the cases, the \ac{fpga} was at least twice faster. \citet{Chen2019} benchmarked a Xilinx ZedBoard against a NVIDIA GTX 1080Ti in the ImageNet dataset \cite{Russakovsky2015}, using a ResNet-18 classifier. During the quantisation process, \citet{Chen2019} could improve the network's accuracy and efficiency, running it at \SI{20}{FPS} and saving \num{100} times less power (consumes about \SI{2.58}{\watt}). \citet{Lin2021} compared a quantised \ac{int8} MobileNet classifier running at the \ac{fpga}'s \ac{dpu} (the \ac{fpga} main core for processing \ac{dl} models, section \ref{sec:heterogeneous-platforms}) against multiple Xilinx \acp{fpga} in the literature. Their main study focused on the AMD-Xilinx ZCU104, which executed the algorithm at \SI{376}{FPS} while consuming only \SI{5}{\watt}. \citet{Zhao2021} benchmarked an AMD-Xilinx ZCU104 against an Amazon Cloud \ac{fpga} EC2, using an \ac{yolo} \ac{int8}. Both devices reached similar results, with up to \SI{13}{FPS} in the Penn Treebank dataset. Also \citet{Jain2021} benchmarked multiple \acp{fpga} using a Tiny-\ac{yolo} \ac{int8} and reached an inference speed between \SIrange{12}{23}{FPS} at the AMD-Xilinx XC7Z035.

Researchers are also looking for some \acp{asic} to execute the neural networks because they can become cheaper, smaller, and easier to integrate with other systems. The most common \acp{asic} are Google Coral \acp{tpu} and Intel \acp{ncs}. \citet{Puchtler2020} benchmarked Coral Edge \ac{tpu} USB Accelerator and Intel \ac{ncs} 2 using a \ac{ssd} MobileNet v2 \ac{int8} against an NVIDIA Jetson Nano and Raspberry Pi 4 with a \ac{ssd} MobileNet v2 with weights in \ac{fp16}. The \acp{asic} were the fastest devices, reaching inference framerates of \SI{55}{FPS} in the Coral Edge \ac{tpu} USB Accelerator and \SI{23}{FPS} at the Intel \ac{ncs} 2. Raspberry Pi 4 was the slowest device, inferring at \SI{4}{FPS}, and the Jetson Nano inferred at \SI{15.90}{FPS}. The authors did not do any power consumption analysis. Also \citet{Aguiar2021,Kovacs2021} evaluated the performance and efficiency of Coral Edge \ac{tpu} USB Accelerator.

{ As illustrated in the revised literature, researchers are constantly looking to improve the \ac{dl} models' speed and accuracy to meet real-time constraints, but most of the work focuses essentially on improving the models' architecture and not their intrinsic properties such as their high parallelisation ratio \cite{Redmon2016,Redmon2018,Bolya2019,Howard2017,Sandler2018,Liu2016}. Moreover, many works essay their algorithms in high-performance devices never used in robotics and mobile applications \cite{Magalhaes2021,Moreira2022}. Some authors argue that some models in embedded devices \cite{Martinez2021,Chiu2020,Rahmaniar2021,Venieris2017,Lin2021,Zhao2021}, but it is not clear which kind of device is more suitable for the target application. }
{Therefore,} our work aims to perform a wide benchmark between heterogeneous platforms for evaluating the performance in the evaluation metrics and time and power efficiency of these {edge computing} devices in robotics applications for running \ac{dl} models, giving continuity to \citet{Aguiar2021}'s work. The authors will focus only on using the RetinaNet ResNet-50 \cite{Lin2020,Humbarwadi2020} fine-tuned in the VineSet dataset \cite{Aguiar2021,Aguiar2021a} and compare them using multiple pointwise models (\ac{fp32}, \ac{fp16}, \ac{int8}) and heterogeneous platforms. The used embedded devices were two \acp{gpu} with \SI{1000}{TFLOPS} (NVIDIA Jetson Nano \SI{2}{GB} and \SI{4}{GB} -- Jetson Nano), one \ac{GPU} with \SI{2000}{TFLOPS} (NVIDIA Jetson TX2 -- TX2), one \ac{tpu} (Coral Dev Board \ac{tpu} --- \acs{tpu}), and \acp{dpu} (AMD-Xilinx ZCU104 Development Board -- ZCU104 -- and AMD-Xilinx Kria KV260 Starter Kit -- KV260). To the author's knowledge, this is the first study involving a big object detection model like RetinaNet ResNet-50 and benchmarking the AMD-Xilinx Kria KV260.

{ The authors aim to assess the RetinaNet using ResNet-50 to near-real-time applications. Although the proposed method is suitable for farming application, this might not be the case for other use-cases. Because farming robots typically run at speeds of \SI{0.5}{\metre\per\second} when operating in vineyards. Using a camera vision sensor with a field of view of \SI{45}{\degree} that operates at \SI{0.5}{\metre} from the grapevine, this sensor can see \SI{0.5}{\metre} of the grapevine. Thus, if the \ac{ann} could infer the images at \SI{5}{FPS}, then the processed images will have an overlap between frames of about \SI{0.4}{\metre} (i.e. \SI{80}{\percent}), which should be sufficient for object detection and tracking. }

{Therefore, the current work aims to innovate in the following aspects:}

\begin{itemize}
    \item {in the authors' knowledge, this is the first research to apply and study a big and complex object detection model like RetinaNet ResNet-50 in heterogeneous platforms;}
    \item {a larger benchmark towards object detection using many different heterogeneous platforms, when compared with the reviewed literature, containing embedded \acp{gpu}, \acp{asic} (i.e. \ac{tpu}) and embedded \acp{fpga} (including the new AMD-Xilinx KV260, designed for robotics applications).   }
\end{itemize}


The next sections of this manuscript are structured as follows. In section \ref{sec:materials and methods}, the author will explore the different used heterogeneous platforms, stating their features and limitations, as well as the required software to deploy the \acp{ann} for the different devices. In the same section, the authors also state the assumptions made and the methodology. In section \ref{sec:results}, the time and power efficiency and performance results in the evaluation metrics are presented. These results are deeply discussed in the section \ref{sec:discussion}, comparing between them and with the revised literature. Section \ref{sec:conclusion} summarises the experiences and the main conclusions, framing them with future required work.

\section{Materials and Methods}\label{sec:materials and methods}

The current section details the methodology and the required material to reproduce this experience. Once this is a \ac{dl} study, it requires a dataset and a \ac{dl} model. The deep \ac{dl} was built and trained in TensorFlow 2.8 Keras\footnote{See TensorFlow, 2022, TensorFlow, URL: \url{https://www.tensorflow.org/}. Last accessed on 05/08/2022 and Keras, 2022, Keras, URL: \url{https://keras.io/}. Last accessed on 05/08/2022}. Because the authors used heterogeneous platforms, additional libraries were required to optimise the models for the specific platforms architectures: Vitis-AI 1.4, Edge \ac{tpu} Compiler, and \ac{tf-trt}\footnote{See AMD-Xilinx, 2022, Vitis-AI, URL: \url{https://www.xilinx.com/products/design-tools/vitis/vitis-ai.html}. Last accessed on 05/08/2022; Coral, 2022, Edge TPU Compiler, URL: \url{https://coral.ai/docs/edgetpu/compiler/}. Last accessed on 05/08/2022; and NVIDIA, 2022, Deep Learning Frameworks Documentation, URL: \url{https://docs.nvidia.com/deeplearning/frameworks/tf-trt-user-guide/index.html}. Last accessed on 05/08/2022, respectively.}.

\subsection{Heterogeneous Platforms}\label{sec:heterogeneous-platforms}

The current research topic aimed to benchmark heterogeneous platforms, looking for faster inference devices, minimising the accuracy drop. The authors compared three embedded \acp{gpu} with \SI{1000}{TFLOPS} and \SI{2000}{TFLOPS} (Jetson Nano \SI{2}{GB}, Jetson Nano \SI{4}{GB}, and TX2), \acp{dpu}, recurring to \acp{fpga} (ZCU104, KV260), and \ac{tpu} (Coral Dev Board \ac{tpu}). For optimisation purposes, each platform required its compiler to improve operations performance in the hardware and thus the inference speed. Additionally, the RTX3090 was used to train the \ac{dl} model and baseline the benchmark with a powerful and efficient \ac{gpu}.

Besides the dedicated hardware, all the used boards also have a \ac{ps} to coordinate the desired tasks and manage the operating system. The \ac{ps} can have multiple architectures, but  AMD64 and ARM64 are the most common in the current state-of-the-art.

\subsubsection{NVIDIA \acp{gpu} and \ac{tf-trt}}

Four NVIDIA \acp{gpu} were used for the current benchmark. NVIDIA RTX3090\footnote{See NVIDIA, 2022, GeForce RTX3090 Family, URL: \url{https://www.nvidia.com/en-eu/geforce/graphics-cards/30-series/rtx-3090-3090ti/}. Last accessed on 05/08/2022} is a powerful GPU designed with Ampere Architecture and \SI{24}{GB} of \ac{vram}. Its powerful features allow the \ac{gpu} to train deep neural networks quickly and with big training batches. Because the NVIDIA RTX3090 is very powerful and efficient, any straight benchmark of speed inference cannot be made, but it could work as a reference \ac{gpu} for the evaluation. Besides, it is unsuitable for embedded applications because of its high power-consumption ratios, until \SI{350}{\watt}.

The NVIDIA Jetson \acp{gpu} were designed as embedded devices to assemble in low-power systems like robots. The two Jetson Nano\footnote{See NVIDIA, 2022, Jetson Nano 2GB Developer Kit
, URL: \url{https://developer.nvidia.com/embedded/jetson-nano-2gb-developer-kit}. Last accessed on 05/08/2022; and 
NVIDIA, 2022, Jetson Nano Developer Kit, URL: \url{https://developer.nvidia.com/embedded/jetson-nano-developer-kit}. Last accessed on 05/08/2022} have similar architecture but differ in the amount of available RAM (\SI{2}{GB} and \SI{4}{GB}). TX2\footnote{See NVIDIA, 2022, Harness AI at the Edge with the Jetson TX2 Developer Kit, URL: \url{https://developer.nvidia.com/embedded/jetson-tx2-developer-kit}. Last accessed on 05/08/2022} is the second generation of Jetson Nano with a TX2 \ac{gpu} against TX1 \ac{gpu}. In all of these boards, the available RAM is shared between the \ac{gpu} and \ac{cpu}. 

Although all the \acp{gpu} are compatible with TensorFlow 2 Keras models, they only reach their maximum performance and efficiency when the \ac{dl} models are optimised for their architecture and specialised CUDA and Tensor cores. NVIDIA deployed CUDA cores and Tensor cores that aim to optimise parallel and matrices operations for maximum performance with \acp{cnn}. \ac{tf-trt} is an NVIDIA library that operates with TensorFlow and \ac{trt} and is responsible for analysing the \ac{ann} graph and inferring the best transformations for speed efficiency using the dedicated cores. Besides these operations, \ac{tf-trt} also allows to change the network's graph resolution between \ac{fp32}, \ac{fp16}, and \ac{int8} (the last one through quantisation). The advantage of \ac{tf-trt} against \ac{trt} is that the first one is compatible with TensorFlow and allows to have a hybrid solution when some operations cannot be converted to a \ac{trt} graph. Therefore, the main graph can have some operations executed in TensorFlow, and others executed in\ac{trt}.

\subsubsection{AMD-Xilinx \acp{fpga} and Vitis-AI}

\acfp{fpga} are integrated circuits that can be reconfigured to meet the designer's needs. Due to its high-reconfiguration capability, \acp{fpga} can be useful for executing parallelizable algorithms while keeping the power consumption low. These boards always have two main components \acf{ps} and \acf{pl}. The \ac{ps} is responsible for managing the operations and memory in the \ac{fpga}, while \ac{pl} concerns to the reconfigurable integrated circuits. AMD-Xilinx deployed the \ac{dpu} cores \cite{AMDXilinx2022}, a proprietary programable engine dedicated for \ac{cnn}. This unit has a register configure module, a data controller module, and a convolution computing module. The \ac{dpu} \ac{ip} can be integrated as a block in the \ac{pl} with direct access to \ac{ps}.

For the current benchmark, the authors chose two \acp{fpga}, ZCU104 and KV260. Both boards have similar architecture and compatibility, but KV260 is newer, more compact and designed thinking in robotics applications. ZCU104 has two \ac{DPU} cores, while KV260 has only one. These two \acp{dpu} allow the ZCU104 to simultaneously process two neural network graphs.

For executing the models in the \ac{dpu}, the graph should be quantised in \ac{int8} weights and converted to a readable \ac{dpu} format. Vitis-AI is a fully integrated system in a Docker\footnote{See Docker, 2022, Docker, URL: \url{https://www.docker.com/}. Last accessed on 05/08/2022} environment created by AMD-Xilinx to manage this process. Vitis-AI is characterised as a comprehensive AI inference development platform for AMD-Xilinx devices. Among other features, Vitis-AI processes TensorFlow, Pytorch, and Caffe models using specific quantisers for the \ac{fpga}'s design. Vitis-AI compiles and optimises the quantised models for the \ac{dpu} architecture. This environment also has additional tools to optimise and debug the compiled neural network, such as pruning and profiling tools.

\subsubsection{Coral \ac{tpu} and Edge \ac{tpu} compiler}

\ac{tpu} is an AI accelerator \ac{asic} designed by Google to optimise the execution of \ac{ann}. This \ac{asic} was made compatible with TensorFlow and accepts \ac{dl} models build with the lite version of TensorFlow (\acs{tflite}). Similarly to \ac{fpga}, the \acp{ann} running in edge computing \acp{tpu} should be quantised to make the models fully compatible with the \ac{asic} architecture.

The whole design and management of the model are made with TensorFlow and \ac{tflite}. The compatible model to the \ac{tpu} is got in \ac{tflite} by the Edge \ac{tpu} Compiler.

The authors used the Coral Dev Board \ac{tpu} which is an embedded board with a \ac{ps} and a \ac{tpu} \ac{som} attached.

\subsection{Dataset}\label{sec:dataset}

\begin{figure*}[!hbt]
    \centering
    \subfloat[][]{\includegraphics[width=0.19\textwidth]{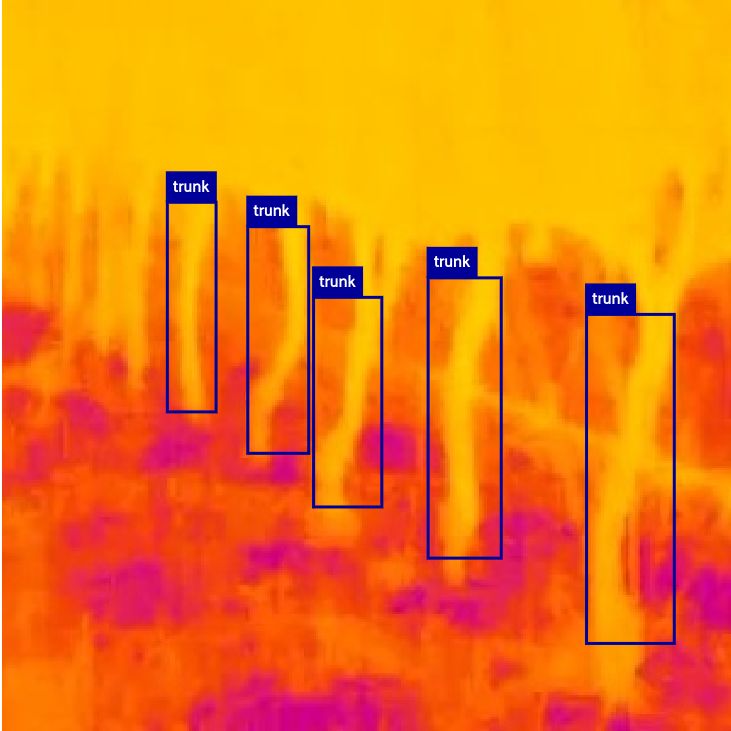}\label{fig:s1}} \hfill
    \subfloat[][]{\includegraphics[width=0.19\textwidth]{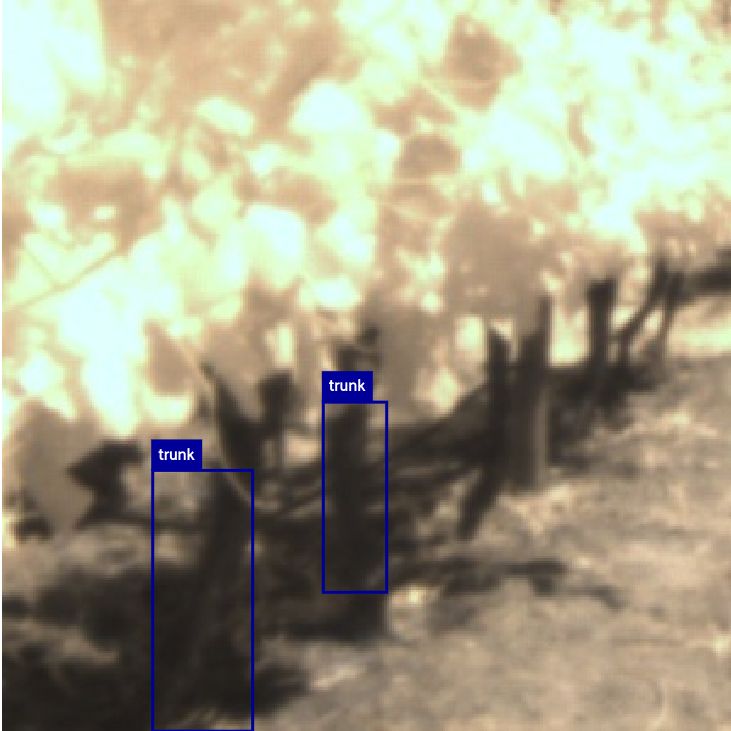}\label{fig:s2}} \hfill
    \subfloat[][]{\includegraphics[width=0.19\textwidth]{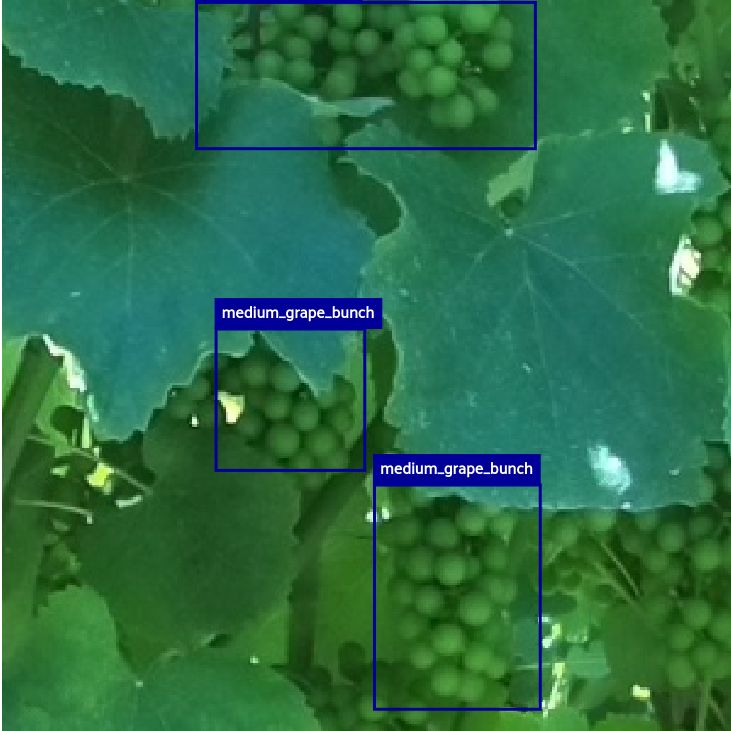}\label{fig:s3}} \hfill
    \subfloat[][]{\includegraphics[width=0.19\textwidth]{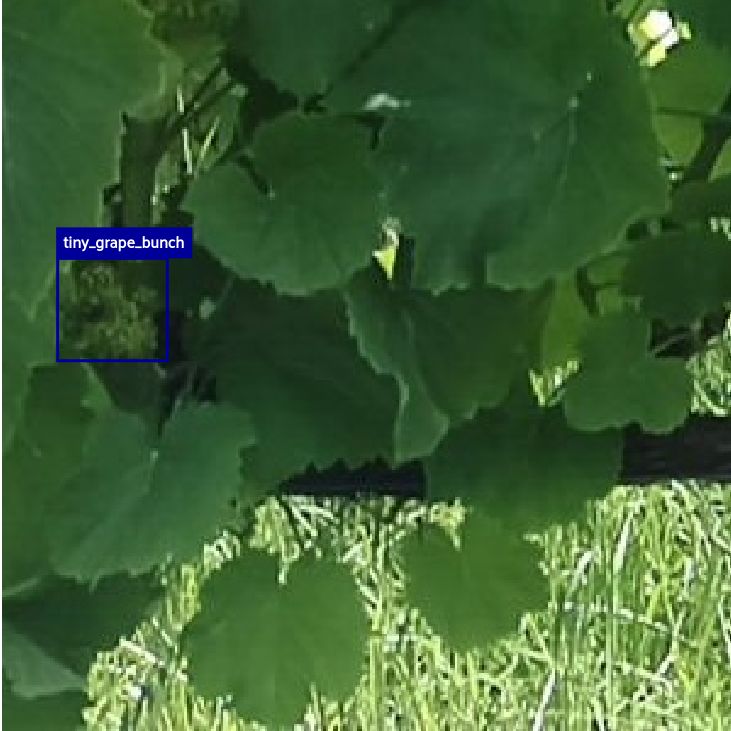}\label{fig:s4}} \hfill
    \subfloat[][]{\includegraphics[width=0.19\textwidth]{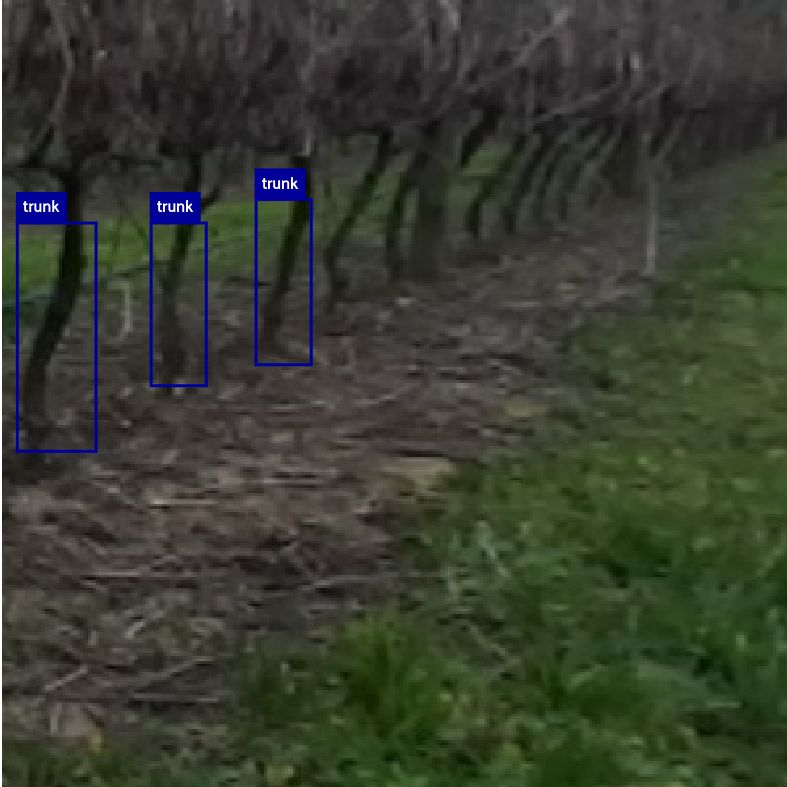}\label{fig:s5}}
    \caption{Sample of images in the dataset \cite{Aguiar2021a} {with the respective ground truth bounding boxes in blue squares}. (a) Thermal image of vines' trunks; (b) image of vines' trunks without infra-red filter; (c) image of bunches of medium-size grapes; (d) image of bunches of corn-size grapes; (e) image of vines' trunks.}
    \label{fig:dataset}
\end{figure*}

The different classification models were benchmarked using the VineSet \cite{Aguiar2021a} dataset composed of \num{428498} images of $300\times300$~\si{px}, manually labelled and gathered from multiple sources (stereo cameras, high-quality cameras, and thermal cameras). Furthermore, the VineSet is composed of natural vineyards images split into the following three classes: vines' trunks, bunches of berry-corn size grapes, and bunches of berry-closed grapes. Figure \ref{fig:dataset} illustrates some images inside the dataset. 


The dataset was split into three batches: train set (\num{411360} images), validation set (\num{8569} images), and test set (\num{8569} images). For consistency in the results with real-world data, the augmentation images in the test set were removed, resuming in \num{1125} images.

\subsection{RetinaNet}

\begin{figure}[!htb]
    \centering
    \includegraphics[width=\linewidth]{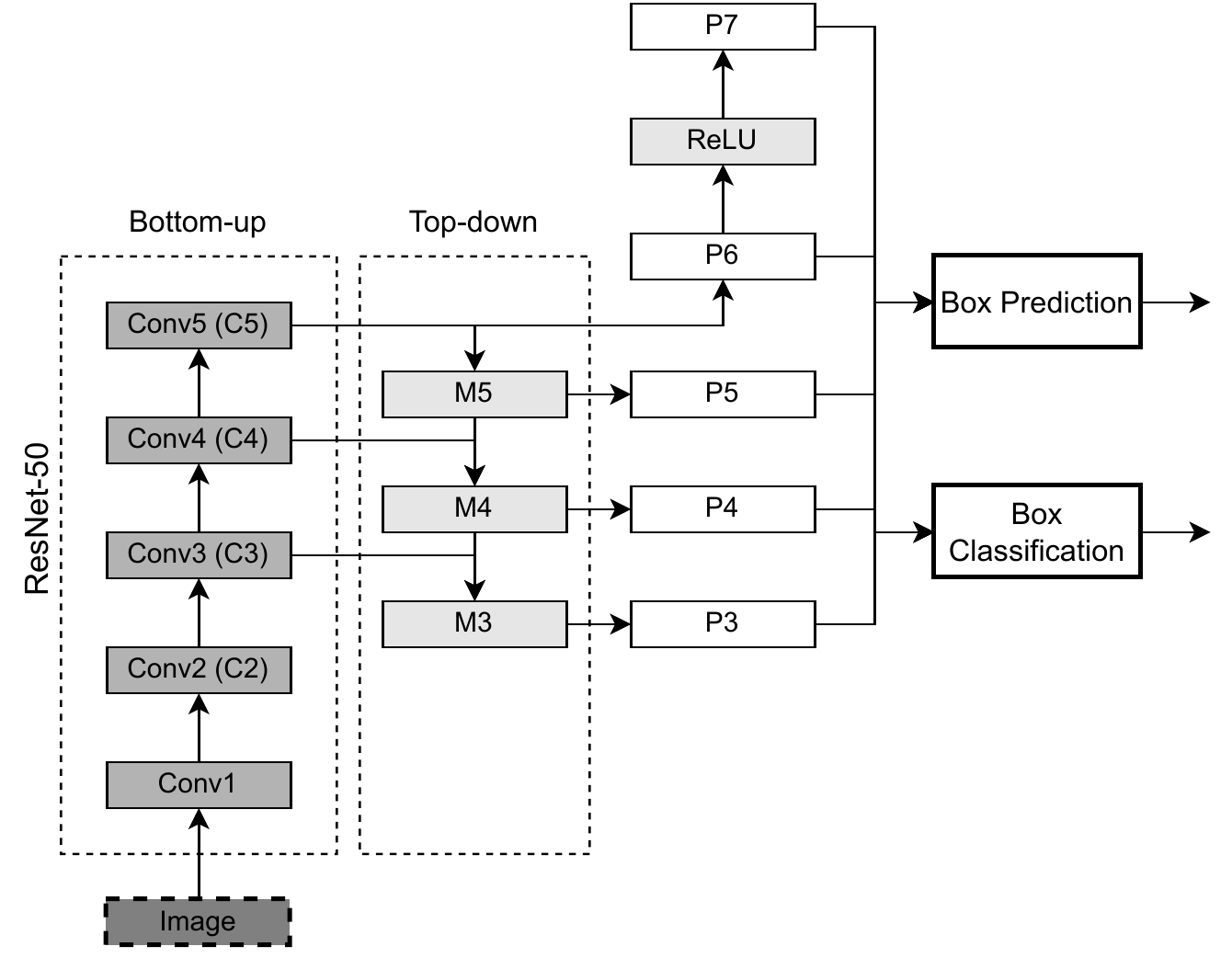}
    \caption{Overview of a simplified diagram of the RetinaNet ResNet-50. Conv$_i$ are convolutional layers; M$_i$ are intermediate layers composed by upsampling, additions and convolutions to generate \ac{fpn} output layers P$_i$; P$_i$ are convolution layers for the output of \ac{fpn}.}
    \label{fig:retinanet-original}
\end{figure}

RetinaNet \cite{Lin2020,Humbarwadi2020} is a state-of-the-art \ac{dl} model for object detection in the class of one-stage detectors. This \ac{dl} model is very similar to an \ac{ssd} \ac{ann} \cite{Liu2016}. After the input layer, a backbone will process the different feature maps and extract the image's features. The backbone is some \ac{cnn} but ResNet-50 is the implemented backbone in the presentation article \cite{Lin2020}. Following the backbone, a \ac{fpn} \cite{Lin2017} is used. These layers follow a top-down architecture (Figure \ref{fig:retinanet-original}) and recover the information processed by the \ac{cnn}, aiming to improve the box classification and regression performance \cite{Lin2017}. The main improvement of RetinaNet against \ac{ssd} \ac{dl} models is the implementation of a new custom loss function, focal loss \cite{Lin2020}, that aims to prioritize the correct detection and classification of the objects, \ac{tp}, against the correct not detection of objects, \ac{tn}.  

Given the improvements in the state-of-the-art provided by RetinaNet ResNet-50 against \ac{ssd} networks, and because these \acp{ann} usually provide better results than \ac{yolo} \cite{Magalhaes2021,Tan2021,Morera2020}, the authors of this benchmark chose to use RetinaNet ResNet-50 as initially stated by \citet{Lin2020}. The authors used a previous model already implemented in TensorFlow 2 Keras by \citet{Humbarwadi2020}, making the necessary changes to the architecture to make it compatible with all the heterogeneous platforms. The model had to be implemented using a functional strategy\footnote{See Tensorflow, 2022, The Functional API, \url{https://www.tensorflow.org/guide/keras/functional}. Last accessed on 05/08/2022}, but pre-processing and post-processing layers were kept in the sub-modelling format because they were not converted or recompiled for any heterogeneous platform. Instead, these layers were reimplemented. The ResNet-50 \cite{He2016} was configured with the same pre-trained weights used by the ImageNet dataset \cite{Russakovsky2015} to ensure consistency and avoid deterioration of processing speed.

\begin{figure}[!htb]
    \centering
    \includegraphics[width=\linewidth]{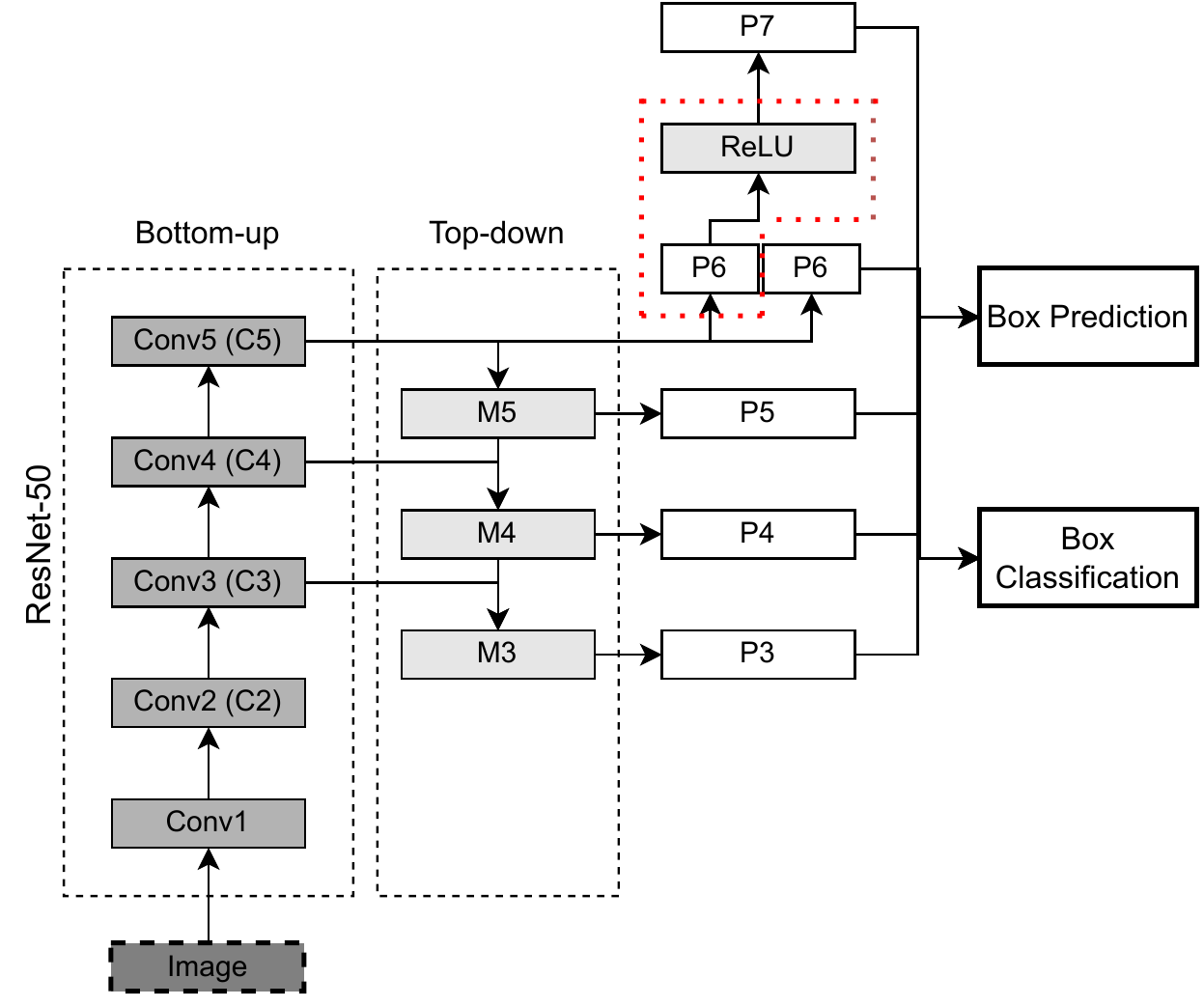}
    \caption{Overview of a simplified diagram of a changed version of RetinaNet ResNet-50 for \ac{fpga} compatibility. Conv$_i$ are convolutional layers; M$_i$ are intermediate layers composed by upsampling, additions and convolutions to generate \ac{fpn} output layers P$_i$; P$_i$ are convolution layers for the output of \ac{fpn}.}
    \label{fig:retinanet-changed}
\end{figure}

Vitis-AI has some operations constraints for compiling the \ac{dl} model to the \acp{fpga}. These constraints were found at \ac{relu} operations that should be immediately preceded by another operation, like a convolution or a mathematical operation. This compromises the compilation of the network, mainly between P6 and P7 of the \ac{fpn} (Fig. \ref{fig:retinanet-original}), because an output for the regression and classification layers are required at the convolution 2D P6 and the convolution 2D P7. Therefore, an additional convolutional 2D layer was added at P6, cloning the initial P6 convolution 2D layer (Figure \ref{fig:retinanet-changed}). In this way, the Vitis-AI compiler can further compile all layers of the model's core at the \ac{dpu}; otherwise, a split of the architecture could happen, and some operations could be executed at the \ac{cpu}. 

The changed version of RetinaNet ResNet-50 (Figure \ref{fig:retinanet-changed}) was trained by fine-tuning until the convergence of the train loss function. The training algorithm used the focal loss function and the \ac{sgd} optimiser. For better adjustment of the learning rate and momentum values, the authors used the Keras Tuner library \cite{omalley2019kerastuner} with the Hyperband algorithm \cite{JMLR:v18:16-558} to search for the best values that optimise the validation loss. During this stage, only two batches of the dataset are used, the train set for training the model and the validation set for evaluating the model performance in the evaluation metrics and tracking the model's overfitting. The model was trained in the \ac{gpu} RTX3090.

\subsection{Deploying RetinaNet ResNet-50 for heterogeneous platforms}\label{sec:edge devices}

The main aim of this study is to assess the performance reliability in the evaluation metrics of \ac{dl} models in heterogeneous platforms and assess their effectiveness for real-time object detection. 

Deploying the RetinaNet for each heterogeneous platform is very similar but requires the use of proprietary libraries. Therefore, the steps to deploy a model for each device are:

\begin{enumerate}
    \item RetinaNet ResNet-50 fine-tuning in the VineSet train set;
    \item Quantise the model to INT8 (optional, depends on the platform);
    \item Deploy the model to a platform's compatible format
\end{enumerate} 

The first step implies the train of the \ac{ann}, which is the same for all platforms and happens in TensorFlow 2 in the RTX3090, as stated in section \ref{sec:dataset}. Because pre-processing and post-processing layers cannot be compiled in some devices, only the core of the \ac{ann} is used in the following steps. Whenever required, these layers are implemented. 

After training, the model is manipulated using the proprietary specific libraries. \ac{tpu} and \acp{fpga} require the use of quantisation. The quantisation can be aware of training of be agnostic to it, happening when the model has already converged. For compatibility issues, only post-training quantisation is compatible with all devices. Therefore, a dataset calibration was derived from the train set to quantise the \ac{ann} weights and calibrate them to the input calibration data. Because any train is being performed, the calibration set did not require the ground truth labels. However, compatible with quantised networks, RTX3090 and Jetson \acp{gpu} did not require them. Besides, as we could conclude in section \ref{sec:results} Jetson devices could not generate quantised models of RetinaNet ResNet-50. 

The last step is to optimise the \ac{ann} nodes to the hardware where they run. That is made with proprietary compilers, namely \ac{tf-trt} for \acp{gpu}, Edge \ac{tpu} Compiler for \ac{tpu}, and Vitis-AI for \acp{fpga}. A full comprehensive tutorial for deploying the RetinaNet ResNet-50 at AMD-Xilinx \acp{fpga} is published in \cite{Magalhaes2022}. 

The deploying of \acp{ann} is heterogeneous devices also require the implementation of pre-processing and post-processing layers whenever required. Because these layers were removed after training, these layers were reimplemented for each device in Python using OpenCV library.

It is important to realise that this work only focuses on the core of the \ac{dl} model. Pre-processing and post-processing tasks are not being optimised and are being executed in the devices \acp{cpu} because of some limitations of some operations with the compilers. 

\subsection{Evaluation}

The network's performance and efficiency in the different devices were evaluated at two levels: results in accuracy and inference speed. 

The authors only considered the model's core to assess the inference. Because the pre-processing and post-processing layers are running at the devices' \ac{cpu}, the authors could had made some unfair comparisons if these layers were used. Besides, in some platforms, these layers could be optimised to increase the level of parallelism, using \ac{gpu} or \ac{pl}. 

The platforms' speed of inference was only assessed in the permanent stage of the platforms. For reaching the permanent stage, the platforms were required to infer \num{50} random images. During this stabilisation stage, the hardware prepares for inference, and the inference times may oscillate. The inference time is counted as the average inference value ($t_{avg}$) between all the images ($N$ images) in the dataset (eq.~\ref{eq:inference_time}).


\begin{equation}
    t_{avg} = \dfrac{\sum_i^N t_i}{N}
    \label{eq:inference_time}
\end{equation}

Additionally, the model is also assessed in terms of results accuracy. Because the authors used post-training quantisation and different quantisation approaches, it was expected differences between results and some degradation relative to the \ac{fp32} model. For assessing this performance, the authors used the Precision (eq.~\ref{eq:precision}), Recall (eq.~\ref{eq:recall}), F1 (eq.~\ref{eq:f1}), and \ac{map}. The \ac{map} is computed through the Precision $\times$ Recall curves and corresponds to the area under the curve. 

\begin{eqnarray}
    \text{Precision} &=& \dfrac{\text{TP}}{\text{TP} + \text{FP}}     \label{eq:precision}\\
    \text{Recall}    &=& \dfrac{\text{TP}}{\text{TP} + \text{FN}}     \label{eq:recall}\\
    \text{F1}        &=& 2 \cdot \dfrac{\text{Precision} \cdot \text{Recall}}{\text{Precision} + \text{Recall}}     \label{eq:f1}
\end{eqnarray}

Because we are considering an object detection problem, the matching between the detection and the ground-truth is made using the \ac{iou} ratio. In the current work, if the \ac{iou} between two labels is higher than \SI{50}{\percent}, than the detection is a \acf{tp}, otherwise is a \ac{fp}. The ground truths that do not have any matching detection are reported as \acp{fn}.

Despite the inference efficiency and framerate, in heterogeneous systems is also relevant to assess the devices' power consumption on standby and while inferring. Heterogeneous platforms are usually applied to mobile systems powered by batteries and should perform for long periods. Therefore, the good selection of a power-effective device may be critical. The devices' power consumption was measured at the power input of the board using a Fluke 175 True RMS multimeter\footnote{See Fluke, 2022, Fluke 175 True-RMS Digital Multimeter, URL: \url{https://www.fluke.com/en-gb/product/electrical-testing/digital-multimeters/fluke-175}. Last accessed on 05/08/2022}. Because this multimeter cannot compute the power directly, that was made in two steps mathematically. A devices power consumption is given $\text{P} = \text{V} \cdot \text{I}$ (\si{\watt}), where V is the powering voltage in Volt and I is the consumed current in Ampere. In the first stage, the authors measured the powering voltage, in parallel, during standby and while inferring. After, they measured the current, assembling the multimeter in series and under the same conditions.

\section{Results} \label{sec:results}

\begin{figure}[!htb]
    \centering
    \includegraphics[width=\linewidth]{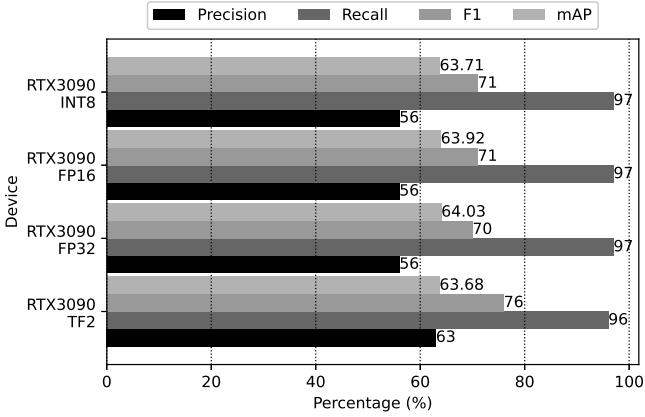}
    \caption{Inference performance in the evaluation metrics in the reference \ac{gpu} considering RAW TensorFlow 2 and the optimised models for NVIDIA Tensor cores. \ac{int8} report to the model's weights quantised into 8-bit integers, \ac{fp16} to weights into 16-bit floating-point, and \ac{fp32} to weights into 32-bit floating-point.}
    \label{fig:RTX3090_results}
\end{figure}

As stated before, the authors are using the RTX3090 as the reference platform to benchmark the RetinaNet ResNet-50 model with the other heterogeneous platforms. The RTX3090 is a high-performing and power-consuming device, therefore, the presented values are only reference values, and no straight comparison should be made, mainly in terms and speed of inference. Figure \ref{fig:RTX3090_results} illustrates the model's accuracy in the test set. The model was compiled to optimise the hardware usage, mainly using Tensor cores. The RetinaNet got similar results in all its compiled versions but slightly better results in the default TensorFlow 2 model's version. This fact can be due to some detection's confidence drop after compilation (some detections were removed due to being inferior to the confidence threshold).

\begin{figure}[!htb]
    \centering
    \includegraphics[width=\linewidth]{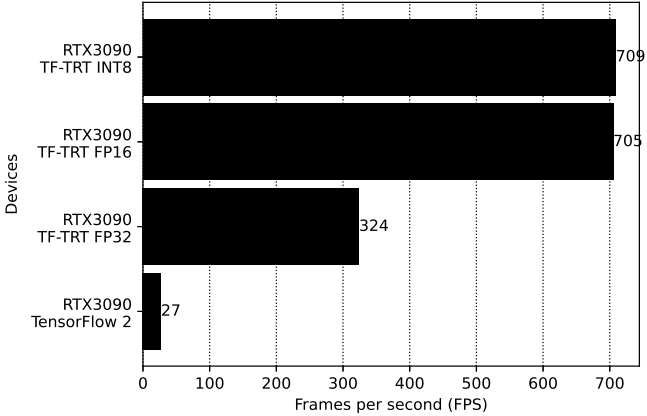}
    \caption{Processing frame rate in the reference \ac{gpu} NVIDIA RTX3090 considering RAW TensorFlow 2 and the optimised models for NVIDIA Tensor cores. \ac{int8} report to the model's weights quantised into 8-bit integers, \ac{fp16} to weights into 16-bit floating-point, and \ac{fp32} to weights into 32-bit floating-point.}
    \label{fig:RTX3090_time}
\end{figure}

In Figure \ref{fig:RTX3090_time} is clear the advantage of compiling the \ac{dl} models for NVIDIA specifics hardware. Without modelling the weights' variables type, i.e., keeping the weights in \ac{fp32}, \ac{tf-trt} could increase the inferences speed 10 times to TensorFlow 2. Reducing the weights resolution from \ac{fp32} to \ac{fp16}, the models got 2.2 times faster than \ac{tf-trt} \ac{fp32} and 26 times faster than TensorFlow 2. Because RTX3090 is not optimised to operate with integers, the conversion to \ac{int8} is meaningless.

\begin{figure}[!htb]
    \centering
    \includegraphics[width=\linewidth]{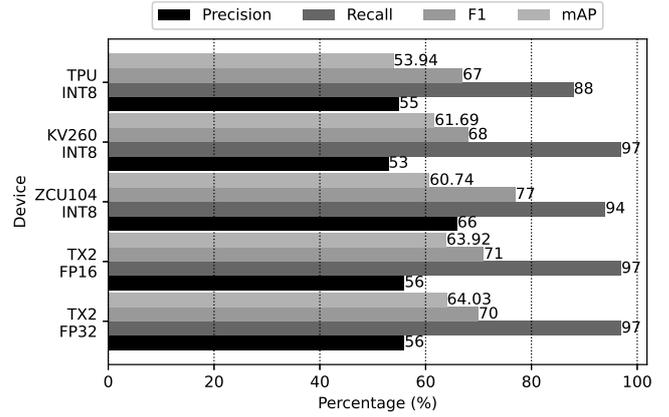}
    \caption{Inference performance in the evaluation metrics in the edge computing devices. \ac{int8} report to the model's weights quantised into 8-bit integers, \ac{fp16} to weights into 16-bit floating-point, and \ac{fp32} to weights into 32-bit floating-point.}
    \label{fig:EgeDevices_results}
\end{figure}

Despite similarities, the performance in the evaluation metrics between the embedded platforms is different (Figure \ref{fig:EgeDevices_results}). The model could not be assessed in any Jetson Nano due to memory and devices' limitations. The best performing device in the evaluation metrics was the TX2. This device could only compile \ac{fp32} and \ac{fp16} models because the device did not get enough memory to convert and quantise the model to \ac{int8}. TX2 got a good balance between precision and recall, which allowed for keeping F1. Conversely, the TPU was the worst performing device in the stated evaluation metrics. The quantisation process caused significant changes in the model's weights and loss of resolution, which reduced both precision and recall and, consequentially, F1. In the mid-term, the \acp{fpga} compensates for the metrics' performance because when they reduce the recall, they increase the precision; or otherwise. The phenomena aid in keeping F1 stable between each other. The \ac{map} follows the analysis made until now. 

\begin{figure}[!htb]
    \centering
    \includegraphics[width=\linewidth]{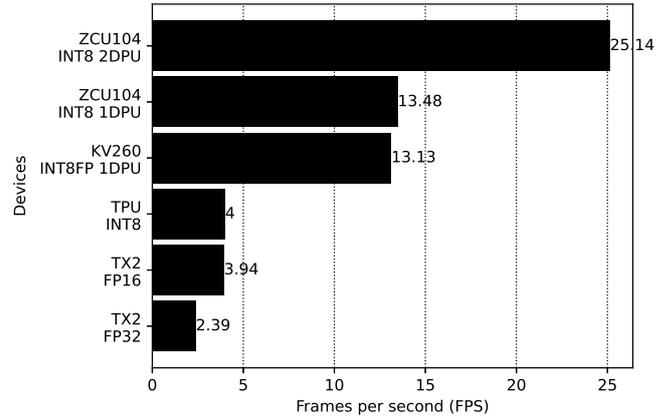}
    \caption{Processing frame rate in the edge computing devices. \ac{int8} report to the model's weights quantised into 8-bit integers, \ac{fp16} to weights into 16-bit floating-point, and \ac{fp32} to weights into 32-bit floating-point. \acp{fpga} can have multiple \ac{dpu} cores: 1\ac{dpu} remains to the use of a single \ac{dpu} and 2\ac{dpu} is the simultaneous use of 2 \ac{dpu} cores.}
    \label{fig:EdgeDevices_time}
\end{figure}

Figure \ref{fig:EdgeDevices_time} illustrates the inference speed of the different devices in the study. The \ac{gpu} was the slowest device between the heterogeneous platforms. The improvement of using \ac{fp32} against \ac{fp16} is in 1.6 times. The model could not be compiled and quantised to \ac{int8}. Conversely, \acp{fpga} prove to be the fastest devices. While using one \ac{dpu} these devices are 5.6 times faster than TX2 \ac{fp32} and 3.4 times faster than TX2 \ac{fp16} and \ac{tpu}. Using the two \acp{dpu} from ZCU104, the inference reaches \SI{25}{FPS}. 

\begin{figure}[!htb]
    \centering
    \includegraphics[width=\linewidth]{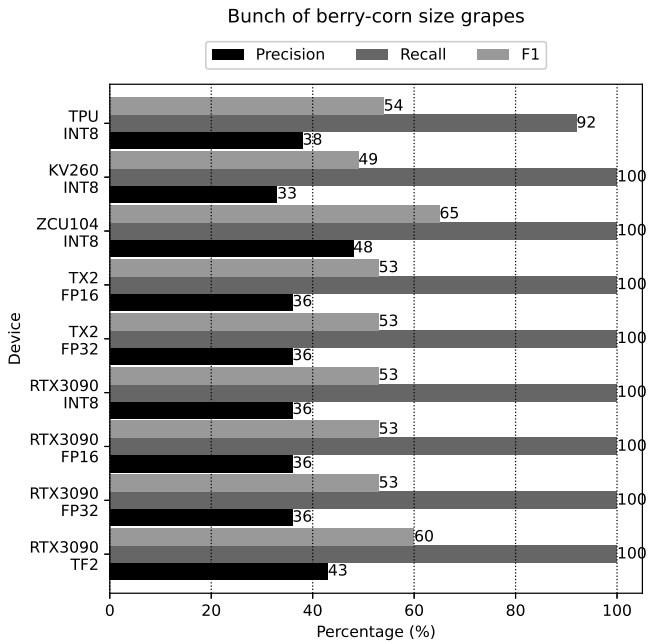}
    \caption{Inference performance for the evaluation metrics in the different heterogeneous devices for the class of bunches of berry-corn size grapes. \ac{int8} report to the model's weights quantised into 8-bit integers, \ac{fp16} to weights into 16-bit floating-point, and \ac{fp32} to weights into 32-bit floating-point. \acp{fpga} can have multiple \ac{dpu} cores: 1\ac{dpu} remains to the use of a single \ac{dpu} and 2\ac{dpu} is the simultaneous use of 2 \ac{dpu} cores.}
    \label{fig:EdgeDevices_corn_grape}
\end{figure}

\begin{figure}[!htb]
    \centering
    \includegraphics[width=\linewidth]{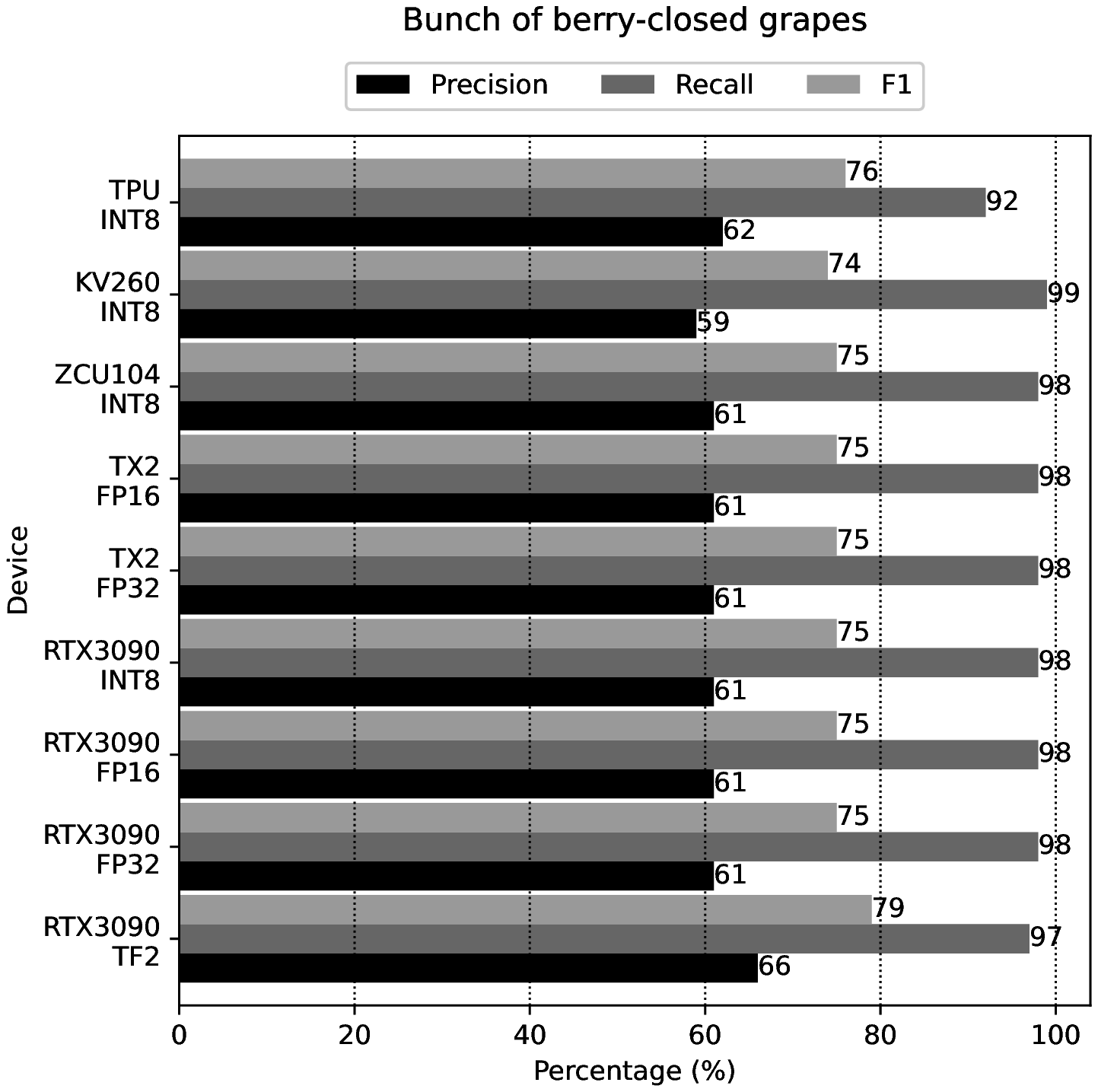}
    \caption{Inference performance for the evaluation metrics in the different heterogeneous devices for the class bunches of berry-closed grapes class. \ac{int8} report to the model's weights quantised into 8-bit integers, \ac{fp16} to weights into 16-bit floating-point, and \ac{fp32} to weights into 32-bit floating-point. \acp{fpga} can have multiple \ac{dpu} cores: 1\ac{dpu} remains to the use of a single \ac{dpu} and 2 \ac{dpu} is the simultaneous use of 2 \ac{dpu} cores.}
    \label{fig:EdgeDevices_medium_grape}
\end{figure}

\begin{figure}[!htb]
    \centering
    \includegraphics[width=\linewidth]{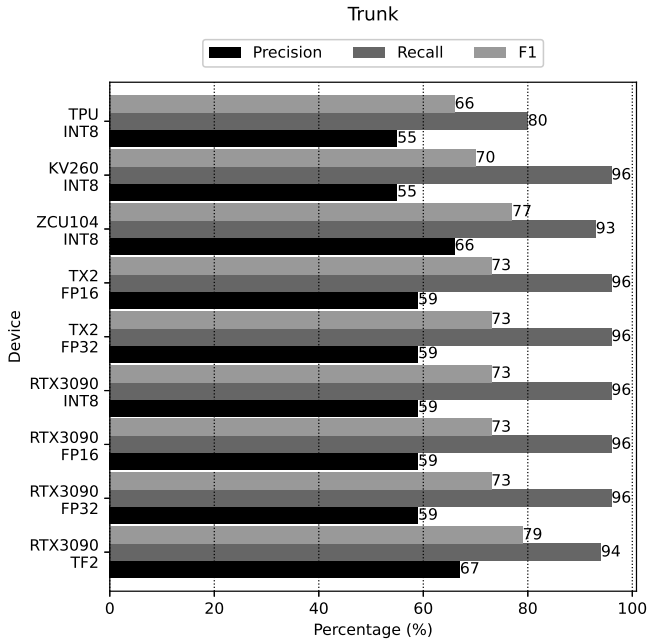}
    \caption{Inference performance for the evaluation metrics in the different heterogeneous devices for the class of trunks. \ac{int8} report to the model's weights quantised into 8-bit integers, \ac{fp16} to weights into 16-bit floating-point, and \ac{fp32} to weights into 32-bit floating-point. \acp{fpga} can have multiple \ac{dpu} cores: 1\ac{dpu} remains to the use of a single \ac{dpu} and 2\ac{dpu} is the simultaneous use of 2 \ac{dpu} cores.}
    \label{fig:EdgeDevices_trunk}
\end{figure}

\begin{figure*}[!htb]
    \centering
    \subfloat[][]{\includegraphics[width=0.19\textwidth]{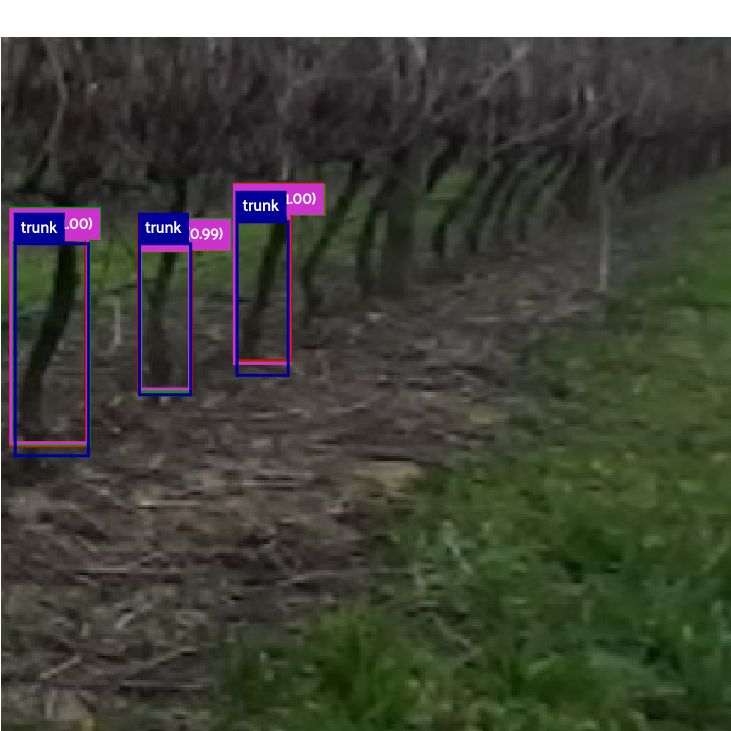}\label{fig:r1}} \hfill
    \subfloat[][]{\includegraphics[width=0.19\textwidth]{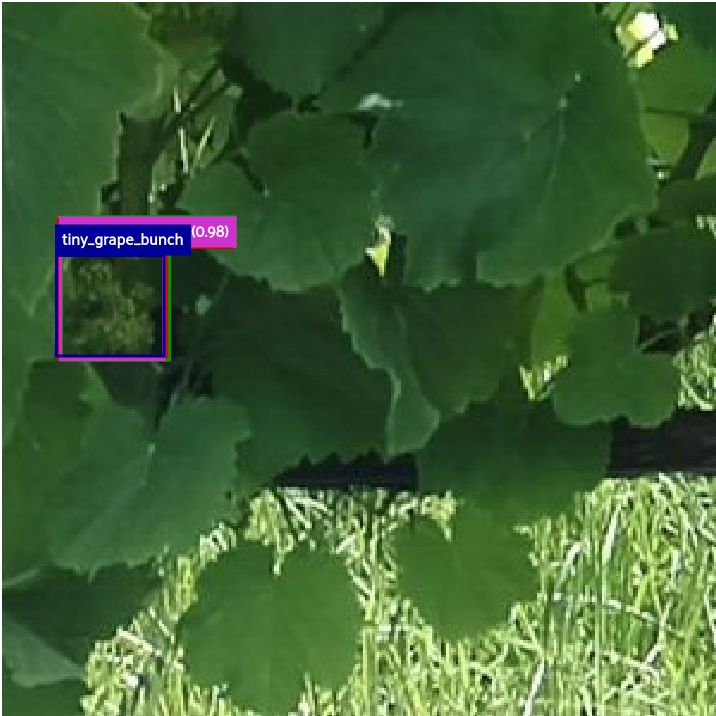}\label{fig:r2}} \hfill
    \subfloat[][]{\includegraphics[width=0.19\textwidth]{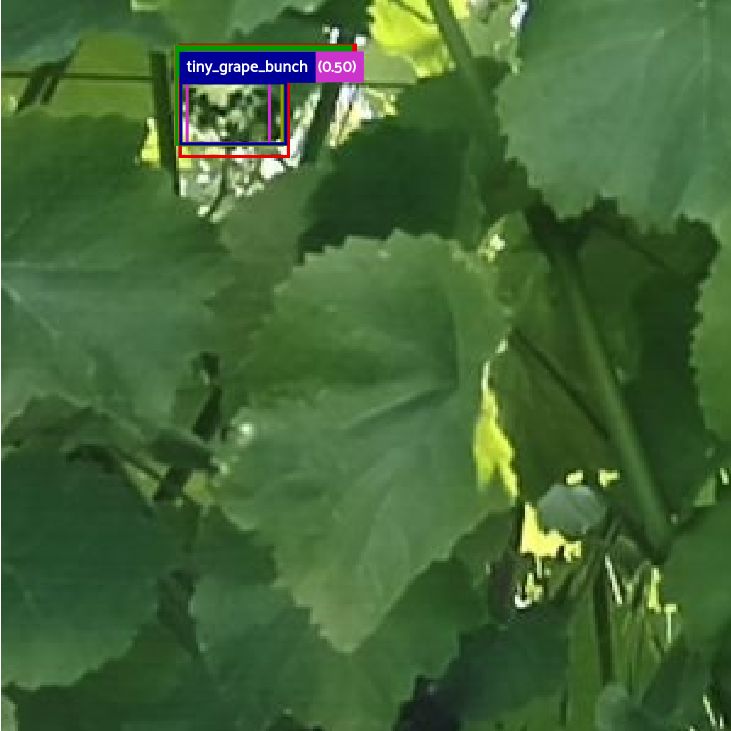}\label{fig:r3}} \hfill
    \subfloat[][]{\includegraphics[width=0.19\textwidth]{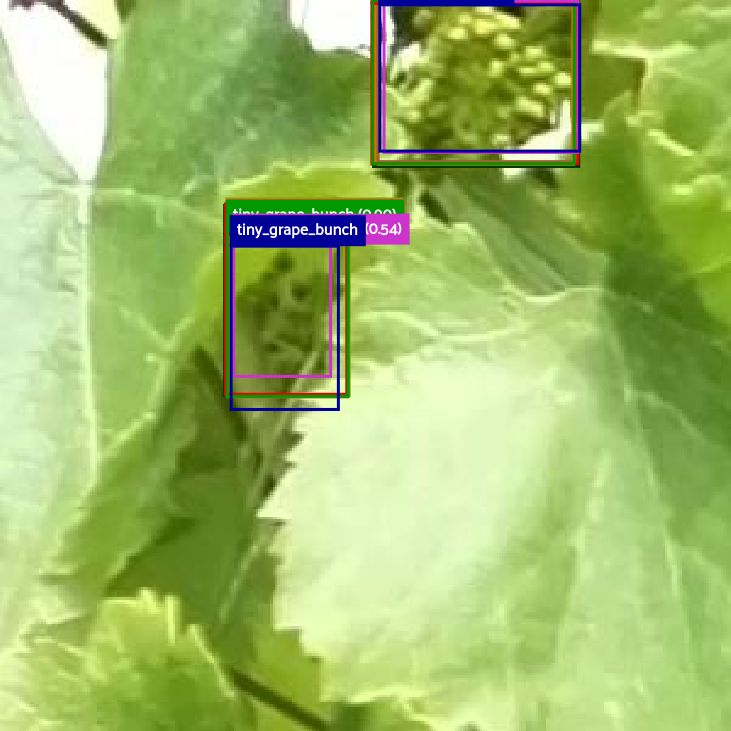}\label{fig:r4}} \hfill
    \subfloat[][]{\includegraphics[width=0.19\textwidth]{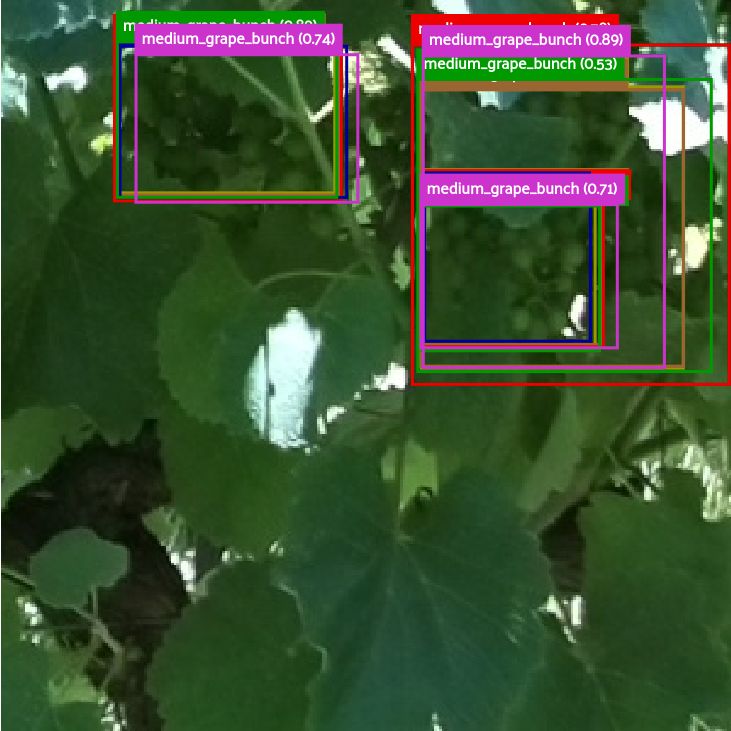}\label{fig:r5}} \hfill
    \subfloat[][]{\includegraphics[width=0.19\textwidth]{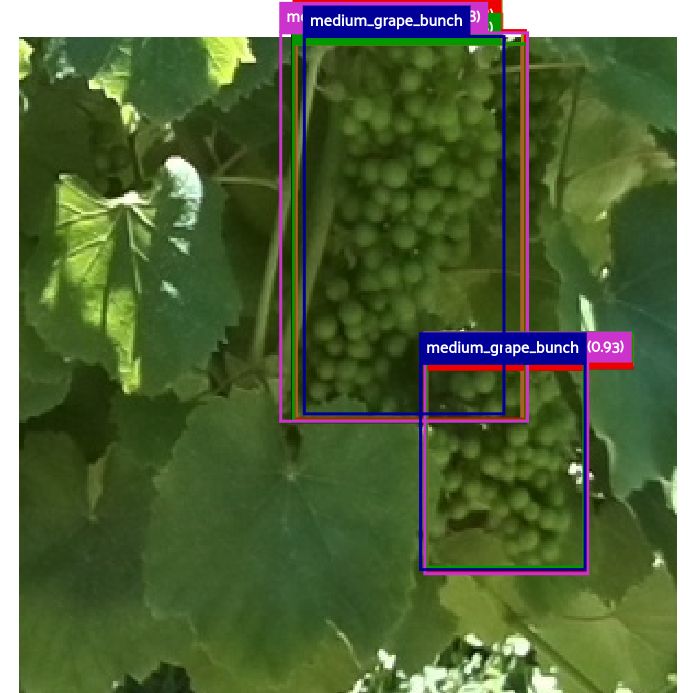}\label{fig:r6}} \hfill
    \subfloat[][]{\includegraphics[width=0.19\textwidth]{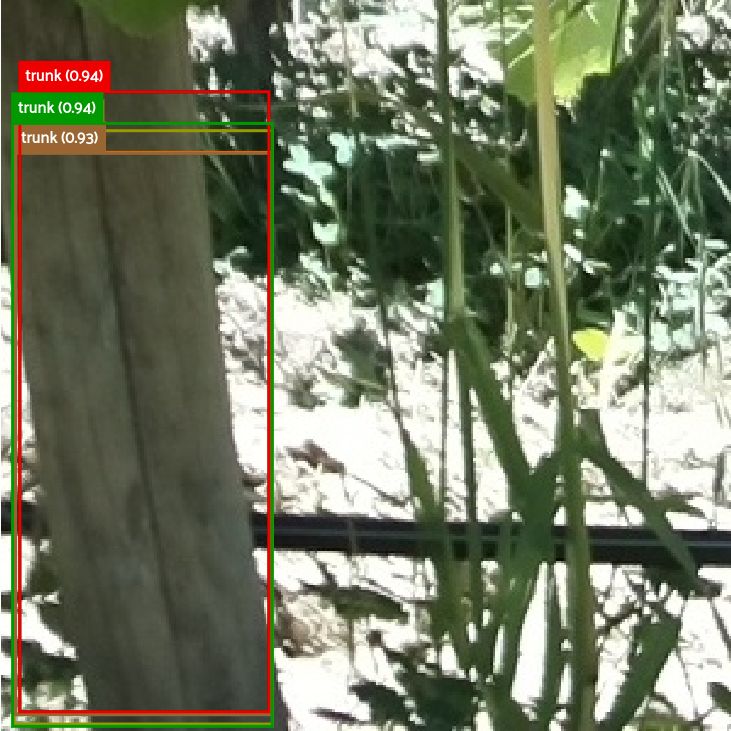}\label{fig:r7}} \hfill
    \subfloat[][]{\includegraphics[width=0.19\textwidth]{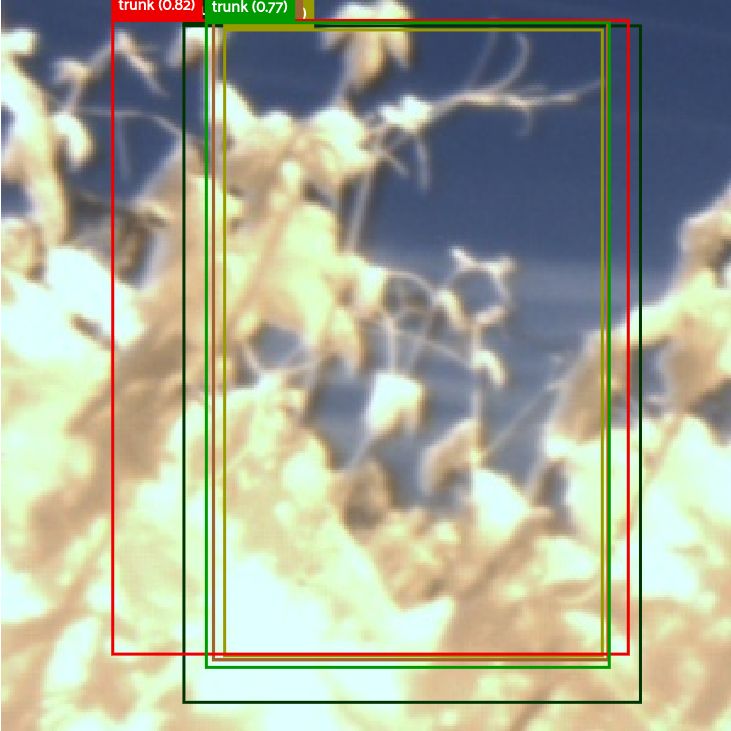}\label{fig:r8}} \hfill
    \subfloat[][]{\includegraphics[width=0.19\textwidth]{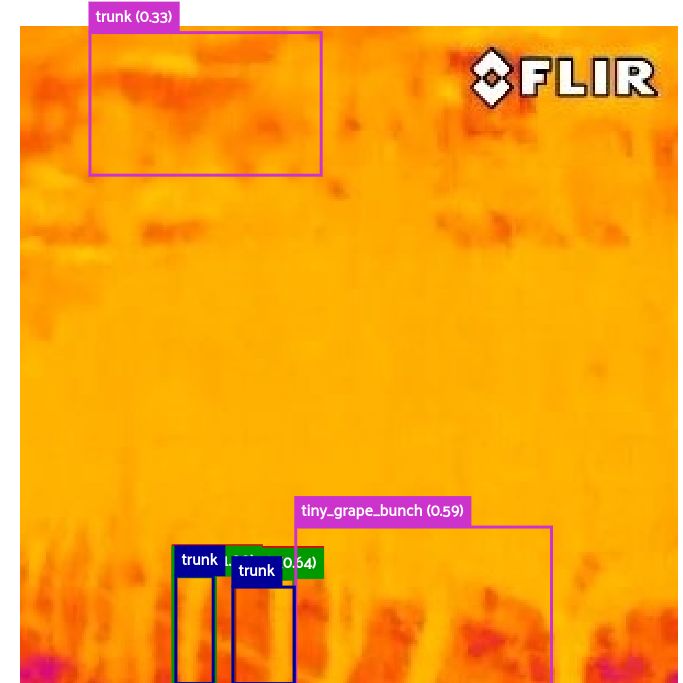}\label{fig:r9}} \hfill
    \subfloat[][]{\includegraphics[width=0.19\textwidth]{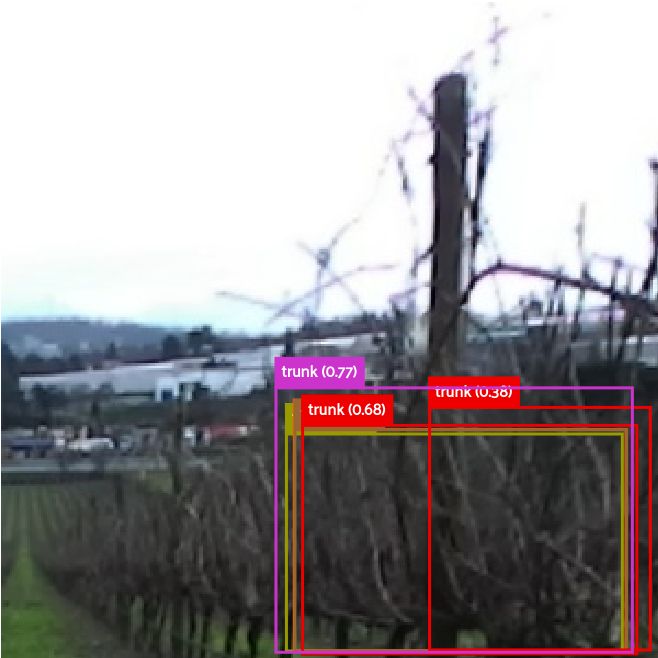}\label{fig:r10}} 
    \caption{Some sample images with the inference results. {Details of this figure were added to \ref{ap:inf_img} in figures \ref{fig:11a} to \ref{fig:11j}.} Blue -- ground-truth; light green -- NVIDIA RTX3080 TF2; orange -- NVIDIA RTX3090 \ac{tf-trt} \ac{fp32}; brown -- NVIDIA RTX3090 \ac{tf-trt} \ac{fp16}; dark yellow -- NVIDIA RTX3090 \ac{tf-trt} \ac{int8}; red -- AMD-Xilinx Kria KV260; dark green -- AMD-Xilinx ZCU104; pink -- Coral Dev Board \ac{tpu}}
    \label{fig:image_results}
\end{figure*}

For better understanding of the effects of quantisation or type of variable changing, figures \ref{fig:EdgeDevices_corn_grape}, \ref{fig:EdgeDevices_medium_grape} and \ref{fig:EdgeDevices_trunk} illustrates the networks' performance in the evaluation metrics for each class. Bunch of berry-closed grapes (figure \ref{fig:EdgeDevices_medium_grape}) is the most stable and predictable class. Changes in the network's weights do not make big changes in evaluation metrics' performance detection. Bunches of berry-corn size grapes and trunks have more difficult features (figure \ref{fig:image_results} {and appendix \ref{ap:inf_img}}). Bunches of berry-corn size grapes are very small (these bunches appear just after inflorescence and are very similar) and have a colour similar to the background. Trunks are highly-variable in shape and size. The images also have many sources. Besides, the network confuses many masts in the vineyards as vines' trunks. The quantisation process in limited resources of \ac{tpu} reduces the number of detections (figures \ref{fig:EdgeDevices_corn_grape} and \ref{fig:EdgeDevices_trunk}), which reduces the \ac{tpu}'s recall (eq. \ref{eq:recall}). The reduction of the number of detections also reduces the number of \ac{fp} and, consequently, the \ac{tpu}'s precision (eq. \ref{eq:precision}). ZCU104 also reveals the marginal case where quantisation reduces the network's noise and improves the detection performance of the evaluation metrics \citet{Gong2014}. 
 
\begin{figure}[!htb]
    \centering
    \includegraphics[width=\linewidth]{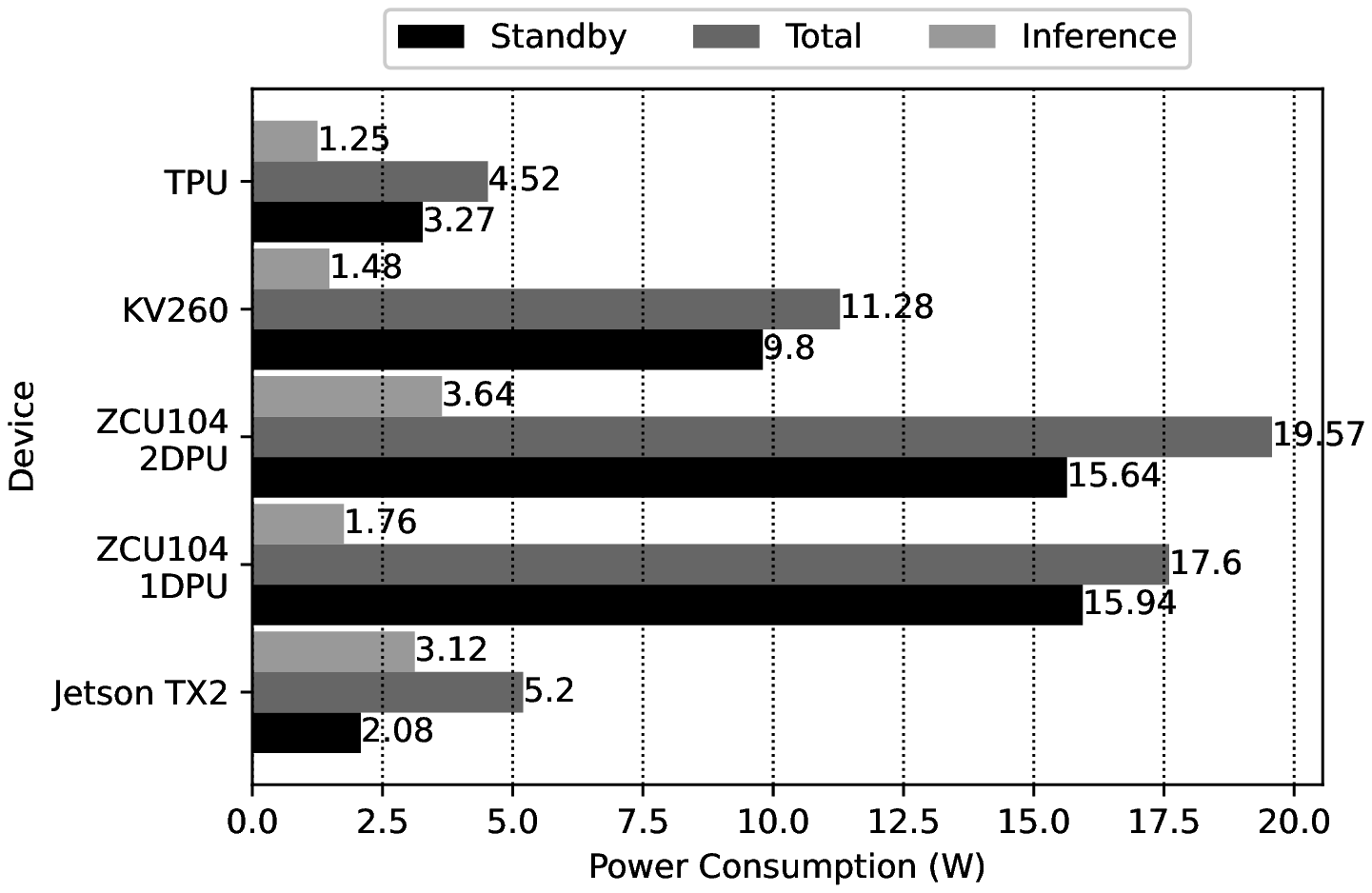}
    \caption{Power consumption.}
    \label{fig:EdgeDevices_power}
\end{figure}

Using heterogeneous platforms in mobile systems, mainly powered by batteries, requires careful power consumption control. In the literature, these are the most common devices for mobile applications. Therefore, figure \ref{fig:EdgeDevices_power} provides power consumption for all devices. Only for inferring, all the devices consume a similar amount of energy, but they vary extremely for their operating system (standby) operations.

\section{Discussion} \label{sec:discussion}

Comparing all the benchmarked devices, it is still clear that when maximum performance in the evaluation metrics and time efficiency are required, using high-performance \acp{gpu} is the best option. However, it is important to mind that the current study does not benchmark other high-performing devices, like server-side \acp{fpga} (like AMD-Xilinx Alveo family), but low-power heterogeneous devices that can be assembled to mobile systems like robots. The compilation of the network to the different devices did not severely change the model's performance in the evaluation metrics, despite some resolution reduction.

Inside edge computing devices, despite \acp{gpu} having the best performing results in the evaluation metrics, \acp{fpga} were much faster. Realise that during this study, only the model's core is being benchmarked, i.e., the authors are excluding pre-processing and post-processing layers. Therefore, due to its features, \acp{fpga} could be more capable of parallelising these layers. Besides \ac{dpu}, they also have the \ac{pl} and an on-board \ac{gpu} that can be used to optimise both blocks of layers. 

The authors also tried to benchmark NVIDIA Jetson Nano \SI{2}{GB} and \SI{4}{GB}, but their limited features impeded converting and compiling the model into \ac{tf-trt}. Because of that, these boards had to be excluded from this research analysis.

Figure \ref{fig:image_results} illustrates some images of the test set with the respective detections registered for each device and ground truth. {Details and extended versions of these images can be found in \ref{ap:inf_img} in figures \ref{fig:11a} to \ref{fig:11j}. } Generically, all the devices could perform well in detecting the target objects (most of the detections are clearly overlapped in the different samples). Figure \ref{fig:r5} shows one of the grapes being detected twice, which was a consequence of its size and because of being overlapped by a leaf. From the images \ref{fig:r2} and \ref{fig:r3} is possible to verify that berry-corn size grapes are the hardest object to detect. {
The reported issue is evident in figure \ref{fig:EdgeDevices_corn_grape}, where the F1 score is generally lower than \SI{60}{\percent}. However, this could not be an impact issue in practical applications once other landmarks can be used for robot localisation, for instance. Nevertheless, trunks' and berry-closed size grapes' detection is more important. The trunks' class is very important for obstacles and the robot's localisation, while the berry-closed size grapes are usually targeted for performing tasks like monitoring or harvesting. These two classes have detection ratios between \SIrange{70}{80}{\percent} (figures \ref{fig:EdgeDevices_medium_grape} and \ref{fig:EdgeDevices_trunk}), which are feasible for practical applications. Therefore, the low \ac{map} of about \SI{60}{\percent} illustrated in figures \ref{fig:RTX3090_results} and \ref{fig:EgeDevices_results} can be induced by the low detection ratio of berry-corn size grapes. 
}
Figures \ref{fig:r6} to \ref{fig:r10} depict some detection errors introduced by the different model's versions. {
The current detection ratios of the different \acp{ann}' versions should conduct further improvements at two levels: optimise the neural network's structure and parameters and deeply review the dataset. Hyper-parameters such as the confidence threshold can be optimised \cite{Magalhaes2021}. The metric results also reveal a possible misannotation of some objects that are being correctly identified, i.e., some objects like trunks could be successfully detected by the model, but they were not labelled in the ground truth.
}

In the revised literature, no publication researched the application of RetinaNet ResNet-50 or \ac{ssd} ResNet-50 \ac{fpn} in heterogeneous devices. So, it is not possible directly compare our results with state-of-the-art results. Although the results show that our experience was slightly slower than state-of-the-art results, RetinaNet ResNet-50 is more complex than \ac{yolo} and \ac{ssd} MobileNet. Given the fast inference times with high-performing rates, which sometimes are similar to \ac{yolo}  results from the revised literature, the authors can conclude that the research from this work is suitable for near real-time applications. 

{
\citet{Aguiar2021} also essayed the VineSet dataset using an \ac{ssd} object detection model with two backbone feature extractors, MobileNet v1 and Inception v2, in a USB Coral Accelerator \ac{tpu}. They reached a \ac{map} of \SI{66.96}{\percent} and \SI{55.78}{\percent}, respectively, without the trunk's class. Given the conditions of not using the trunks' class, we can assume that these are similar results to ours, and the inclusion of trunks in the dataset may lead to the metrics' degradation. Therefore, we can induce that we are working near the limits of the dataset, requiring a deeper labelling review to identify possible misannotations. Besides, \citet{Aguiar2021} performed an inference threshold analysis to identify the best confidence score that optimises the metrics, while we are using a standard confidence score of \SI{30}{\percent}.
}

{
As expected, MobileNet networks aim to be faster and designed for mobile applications. Similarly, Inception networks are also less complex than ResNet networks and, because of that, faster. Inside a \ac{tpu}, the networks reached \SI{158.98}{FPS} and \SI{38.36}{FPS}, respectively. Undoubtedly, this previous work reached faster performances than ours with similar performances. However, it is unclear if there is any difference in the networks' performance between a USB Accelerator Edge \ac{tpu} and the Dev Board \ac{tpu}. The authors did not make a formal power consumption analysis but could infer an average power consumption of \SI{2.5}{\watt} for the USB stick, ignoring all the power consumption for the computer maintenance and processing. 
}

Considering our results, the \ac{tpu} is the best solution when reducing the power is a demand, despite the small variations in the networks' performance in the evaluation metrics and the reduced inference speed. However, when applications are looking for a balance between power consumption and inference speed, KV260 has high potential. In all the cases, keep in mind that ZCU104 and KV260 have installed a standard PetaLinux\footnote{See AMD-Xilinx, 2022, PetaLinux Tools, URL: \url{https://www.xilinx.com/products/design-tools/embedded-software/petalinux-sdk.html}, Last accessed on 05/08/2022} image provided by AMD-Xilinx. These images have all the \acp{fpga}' resources active. Most of the resources are not necessary. Therefore, a deeper analysis with a better configured PetaLinux image can better assess the power consumption of \acp{fpga}.


\section{Conclusions} \label{sec:conclusion}

In this work, multiple heterogeneous platforms (i.e., \ac{gpu}, \ac{tpu}, and \ac{fpga}) were benchmarked using RetinaNet ResNet-50. The code used in this work is publicly available at GitLab INESC TEC, URL: \url{https://gitlab.inesctec.pt/agrob/xilinx-acc2021}. AMD-Xilinx ZCU104 performed better than the other benchmarked platforms because of its fast inference speed. Besides, ZCU104 also has the possibility to execute two models simultaneously. Furthermore, \acp{fpga} offer more flexibility to implement and parallelise algorithms because of their onboard \ac{cpu}, \ac{gpu} and \ac{pl}. \ac{tpu} are better optimised and specified for running \ac{ann} (but more task restrictive), offering a lower power consumption. These devices may be the recommended option when saving power is crucial and the application is not time-restrictive. 

Concerning the frameworks for \acp{ann}' deploying, all of them have similar steps. Vitis-AI is the most complete but complex framework, becoming the hardest to use. Conversely, Edge \ac{tpu} Compiler and \ac{tf-trt} are similar and easier to use, but they depend strongly on TensorFlow. Edge \ac{tpu} Compiler is the easiest framework because it has cross-compiling capabilities, allowing to use of more powerful devices to deploy the model for the \ac{tpu}. \ac{tf-trt} requires the model to be compiled on-device, highlighting the devices' limitations.

Future work intends to optimise the researched \ac{dl} model by applying some optimisation strategies like pruning and exploring the use of binary neural networks. {Besides, \acp{gpu}, \acp{tpu} and \acp{fpga} have computational resources that could be considered to redesign and optimise RetinaNet ResNet-50 nodes to reach lower inference times.} The authors will also evaluate other pre-processing and post-processing techniques for reducing the inference time. 
{Besides, the current work allows the authors to identify possible issues in the dataset labelling, therefore, a deep review of the dataset labels must be an important future step.}







\section*{Funding}

Sandro Costa Magalhães was granted by the Portuguese funding agency, Fundação para a Ciência e Tecnologia (FCT), and the European Social Found (EFS) under scholarship SFRH/BD/147117/2019. 

This work was supported by the European Union’s Horizon 2020 Research and Innovation Program under Grant 101004085.



\bibliographystyle{elsarticle-num-names}
\bibliography{sn-bibliography}

\appendix
\section{Sample images of the dataset} \label{ap:inf_img}

\begin{figure*}[!htb]
    \centering
    \subfloat[][Ground truth]{\includegraphics[width=0.24\textwidth]{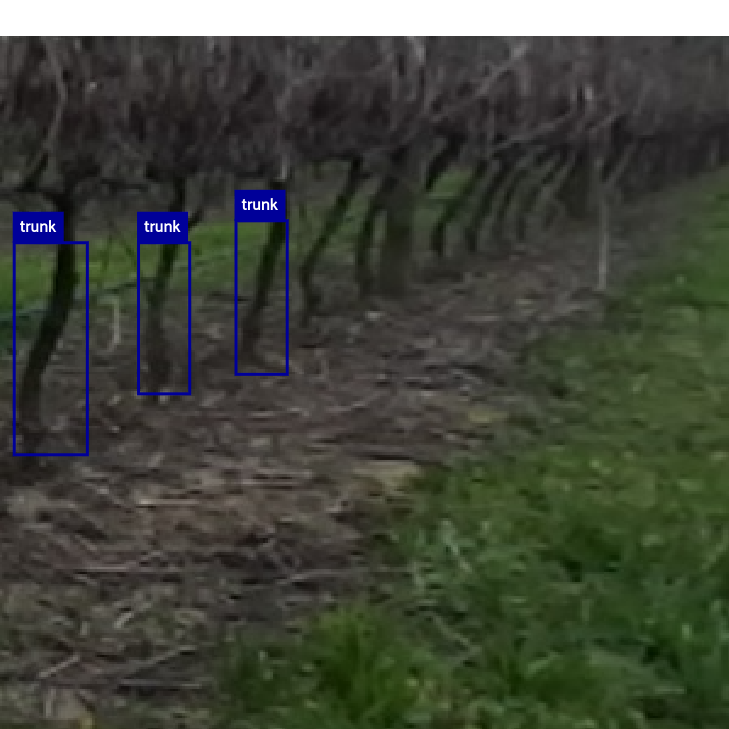}} \hfill
    \subfloat[][RTX3090 TF2]{\includegraphics[width=0.24\textwidth]{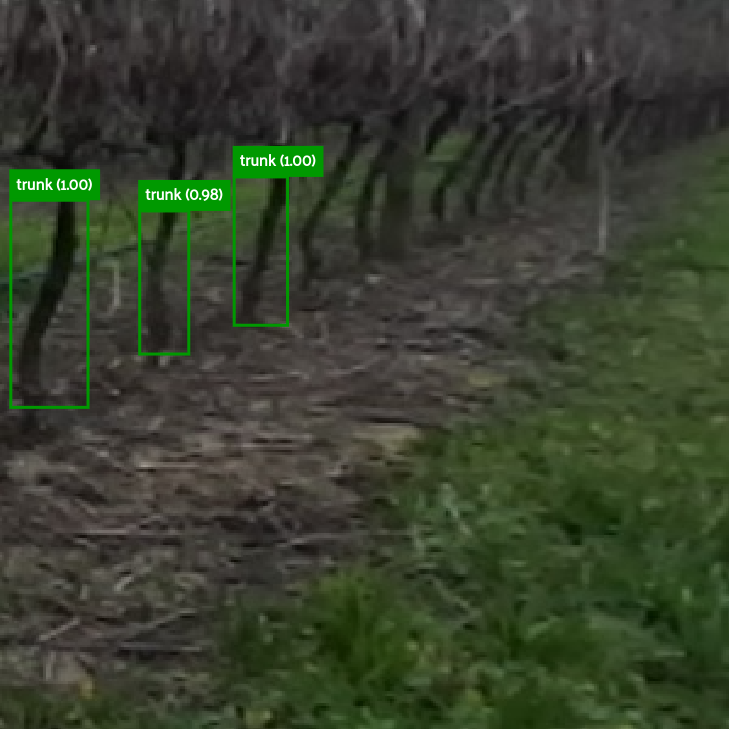}} \hfill
    \subfloat[][TF-TRT FP32]{\includegraphics[width=0.24\textwidth]{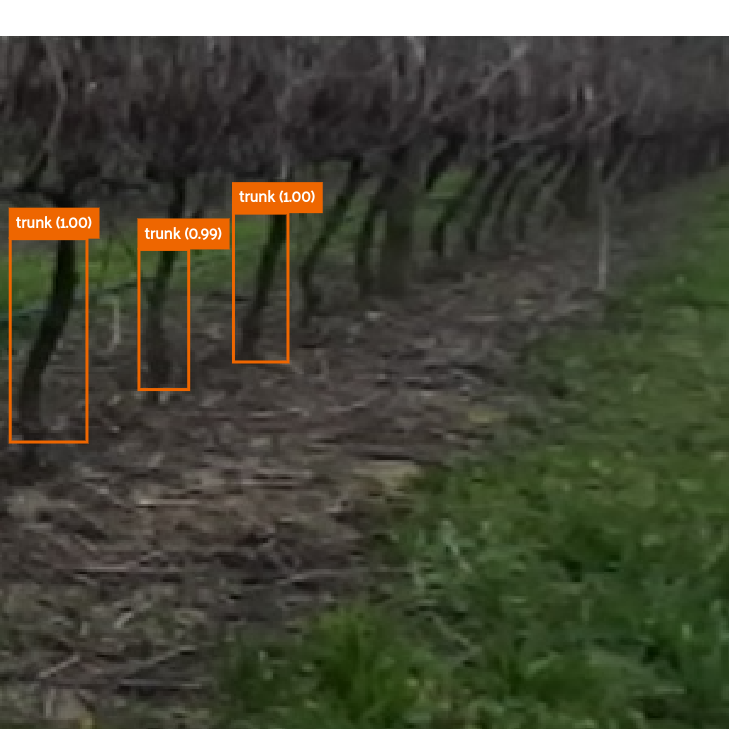}} \hfill
    \subfloat[][TF-TRT FP16]{\includegraphics[width=0.24\textwidth]{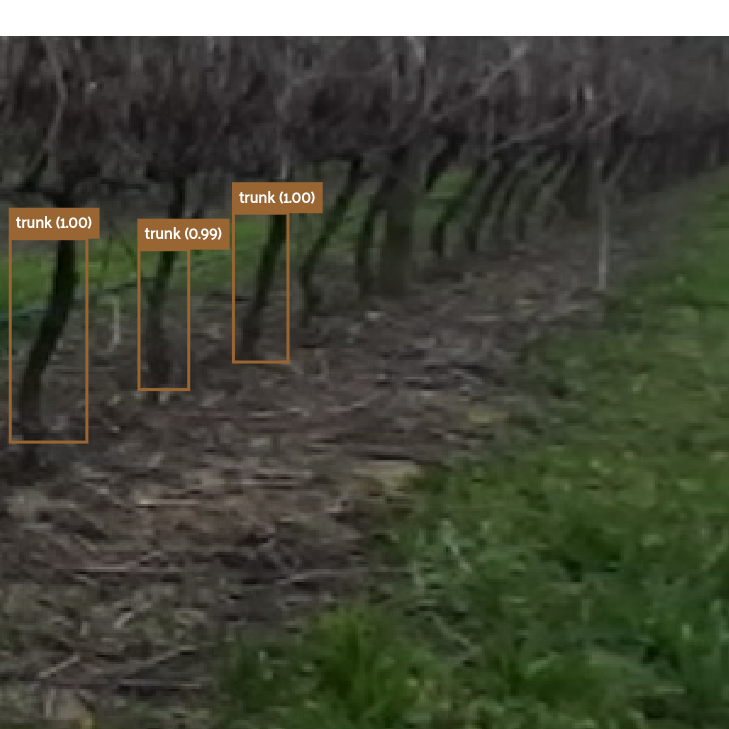}} \hfill
    \subfloat[][TF-TRT INT8]{\includegraphics[width=0.24\textwidth]{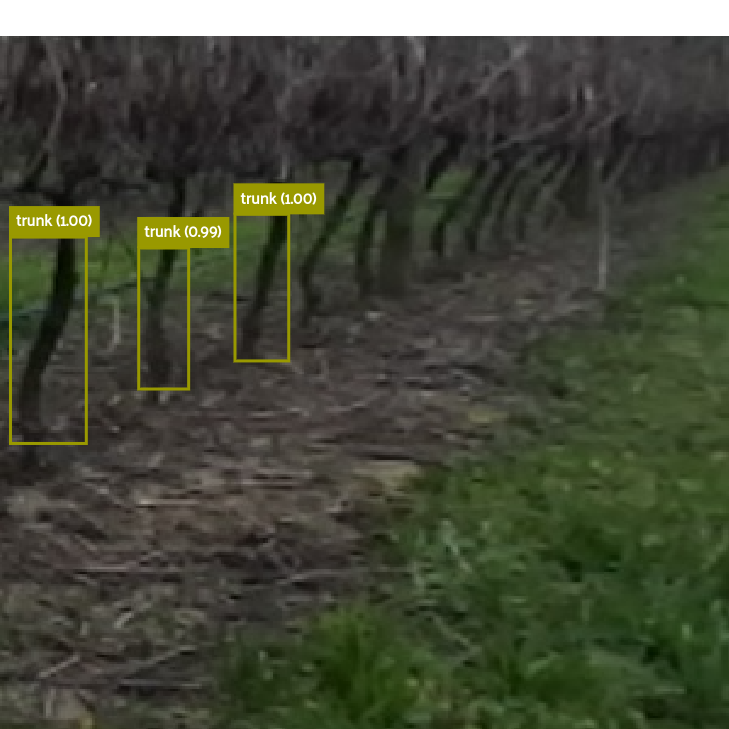}} \hfill
    \subfloat[][KV260]{\includegraphics[width=0.24\textwidth]{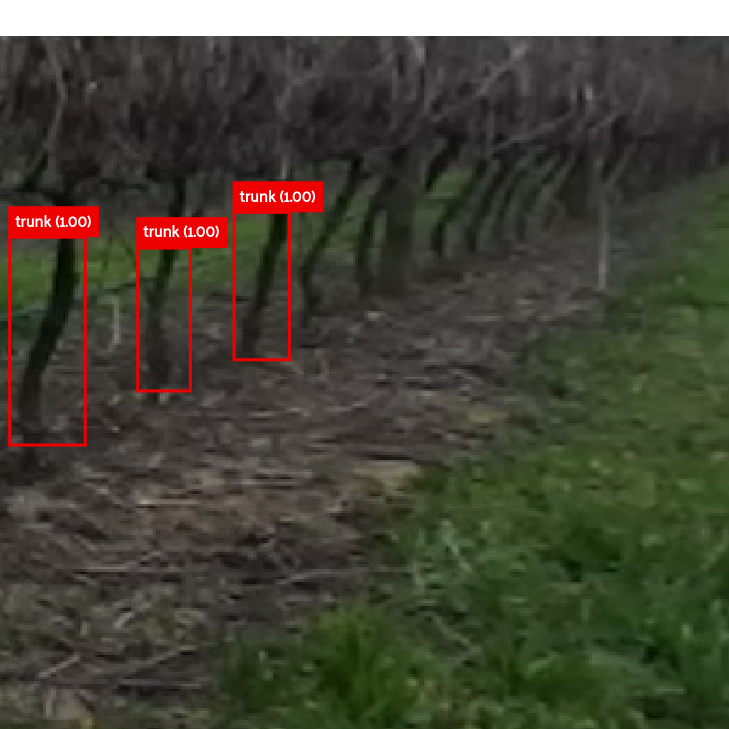}} \hfill
    \subfloat[][ZCU104]{\includegraphics[width=0.24\textwidth]{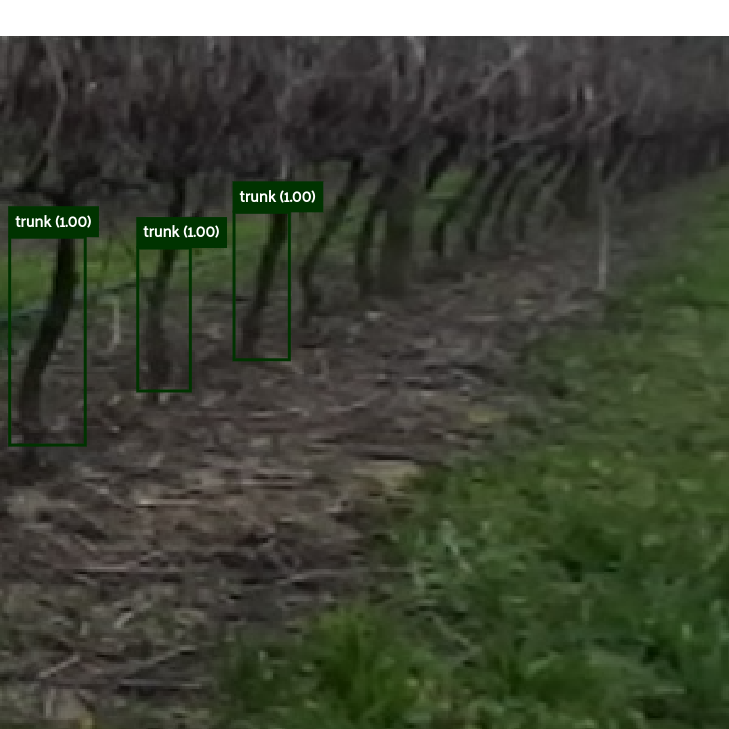}} \hfill
    \subfloat[][TPU]{\includegraphics[width=0.24\textwidth]{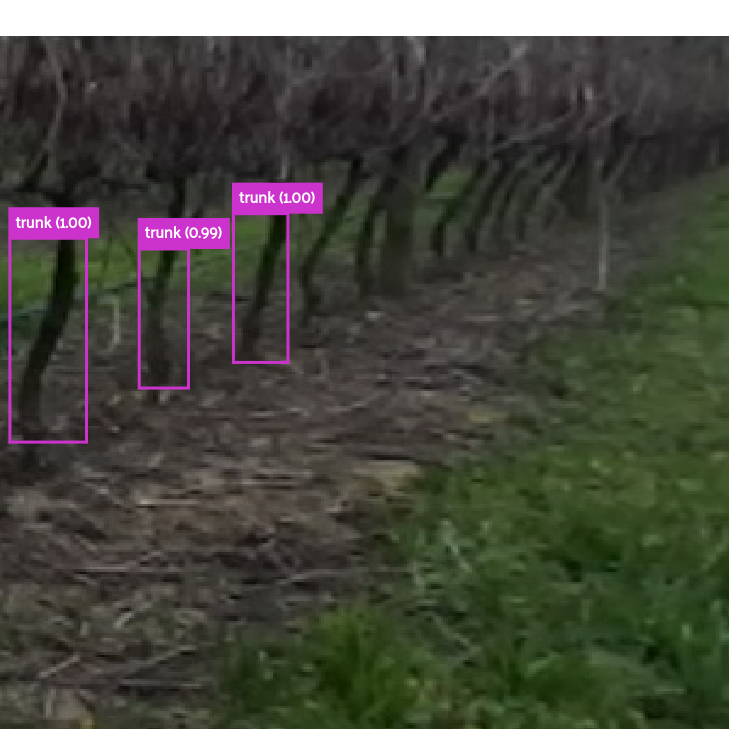}} \hfill
    \caption{Detailed sample image \ref{fig:r1} from figure \ref{fig:image_results} from  Blue -- ground-truth; light green -- NVIDIA RTX3080 TF2; orange -- NVIDIA RTX3090 \ac{tf-trt} \ac{fp32}; brown -- NVIDIA RTX3090 \ac{tf-trt} \ac{fp16}; dark yellow -- NVIDIA RTX3090 \ac{tf-trt} \ac{int8}; red -- AMD-Xilinx Kria KV260; dark green -- AMD-Xilinx ZCU104; pink -- Coral Dev Board \ac{tpu}}
    \label{fig:11a}
\end{figure*}

\begin{figure*}[!htb]
    \centering
    \subfloat[][Ground truth]{\includegraphics[width=0.24\textwidth]{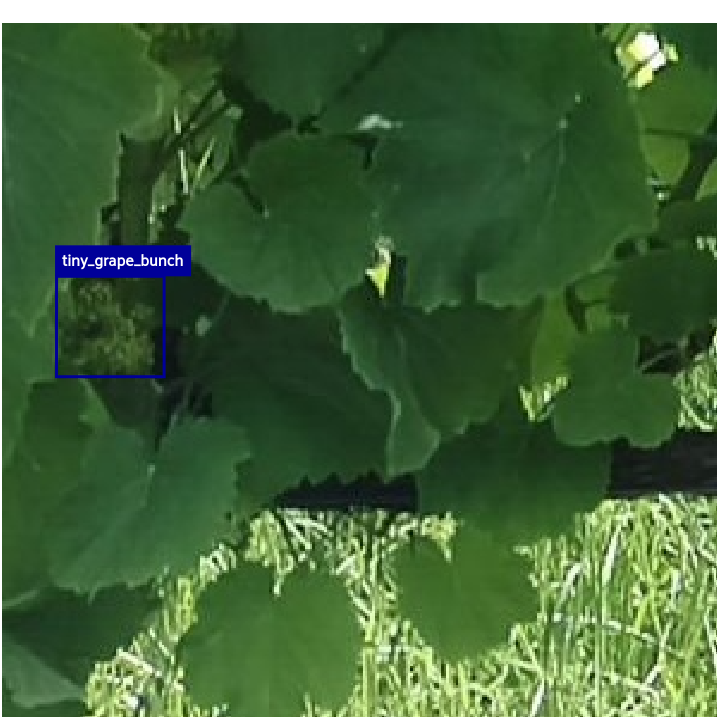}} \hfill
    \subfloat[][RTX3090 TF2]{\includegraphics[width=0.24\textwidth]{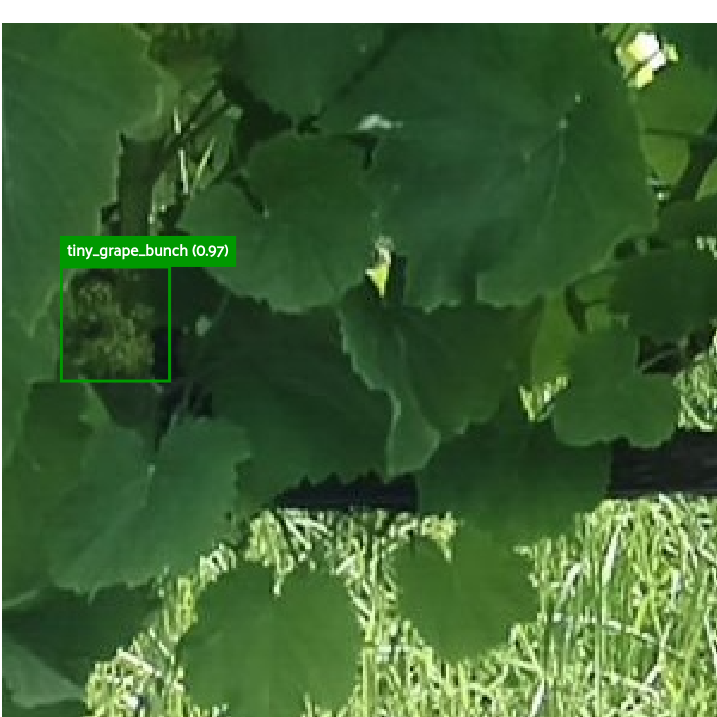}} \hfill
    \subfloat[][TF-TRT FP32]{\includegraphics[width=0.24\textwidth]{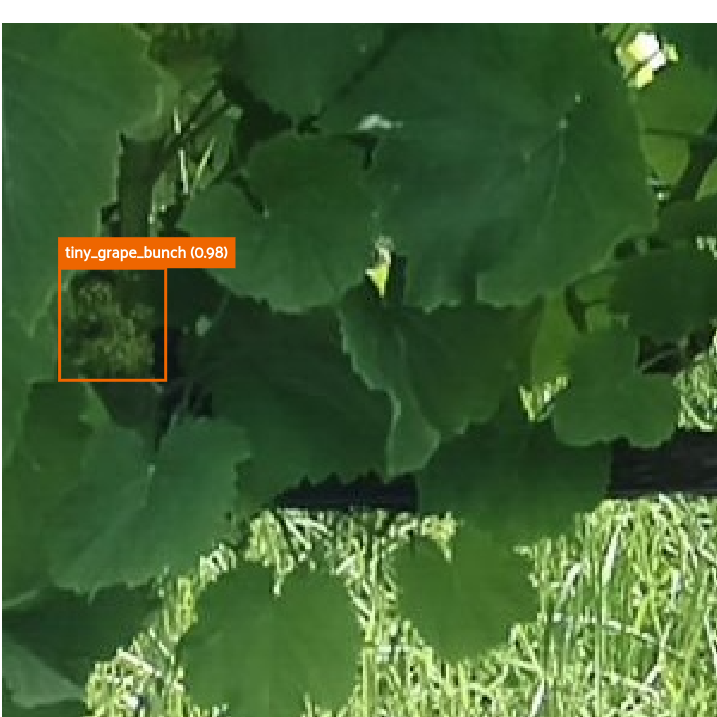}} \hfill
    \subfloat[][TF-TRT FP16]{\includegraphics[width=0.24\textwidth]{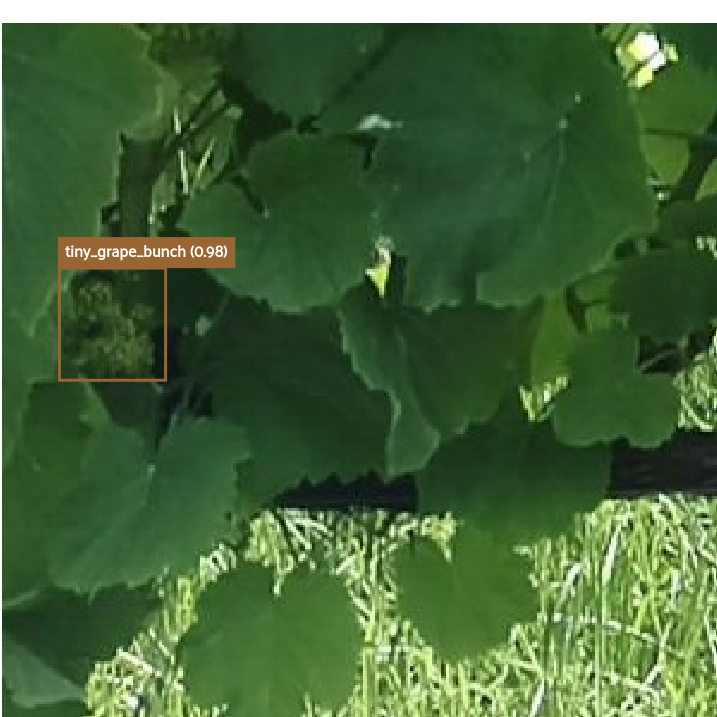}} \hfill
    \subfloat[][TF-TRT INT8]{\includegraphics[width=0.24\textwidth]{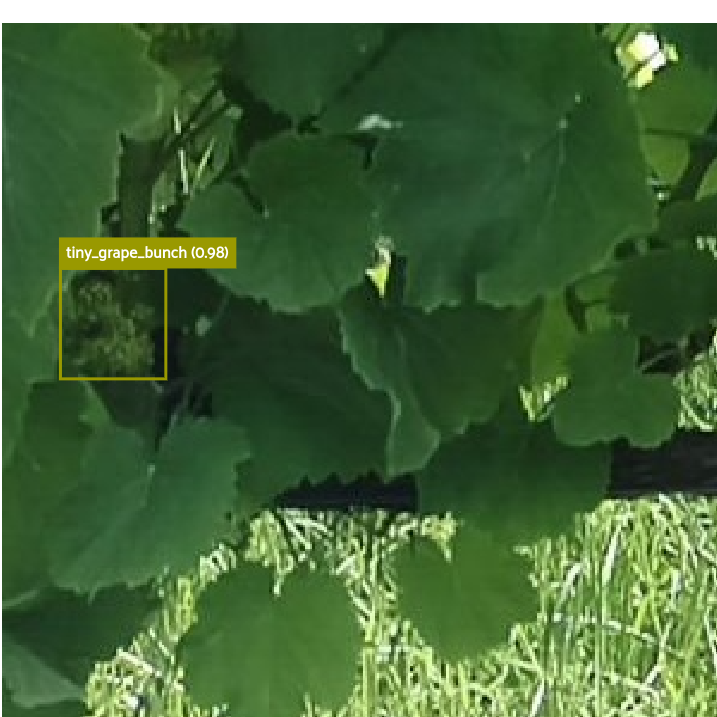}} \hfill
    \subfloat[][KV260]{\includegraphics[width=0.24\textwidth]{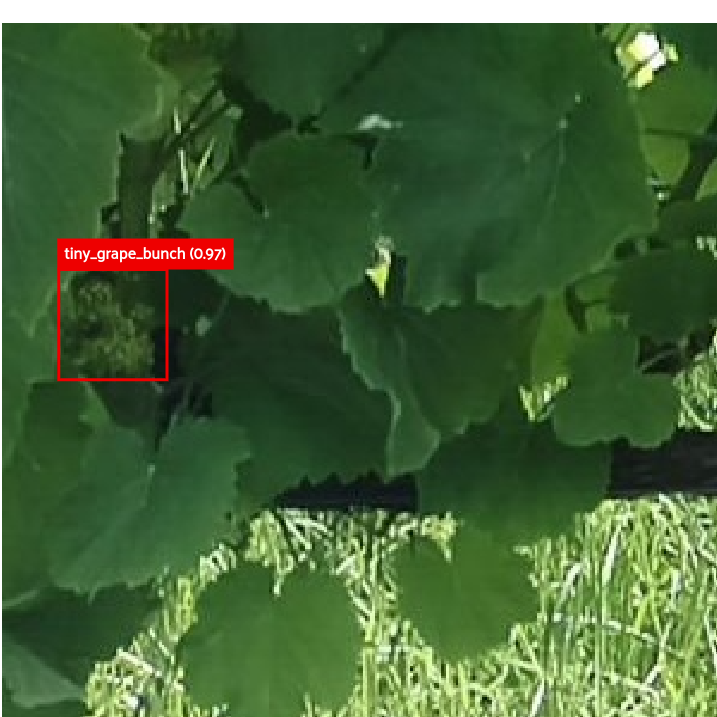}} \hfill
    \subfloat[][ZCU104]{\includegraphics[width=0.24\textwidth]{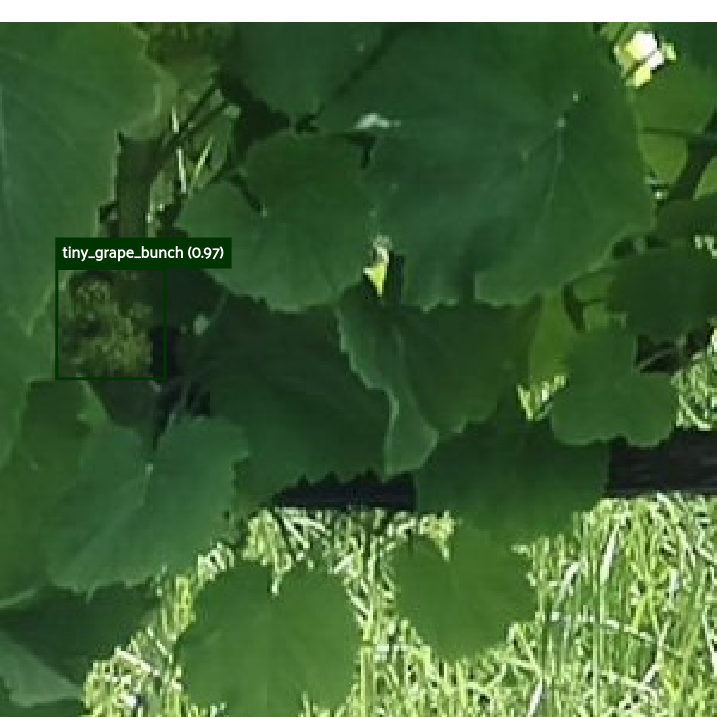}} \hfill
    \subfloat[][TPU]{\includegraphics[width=0.24\textwidth]{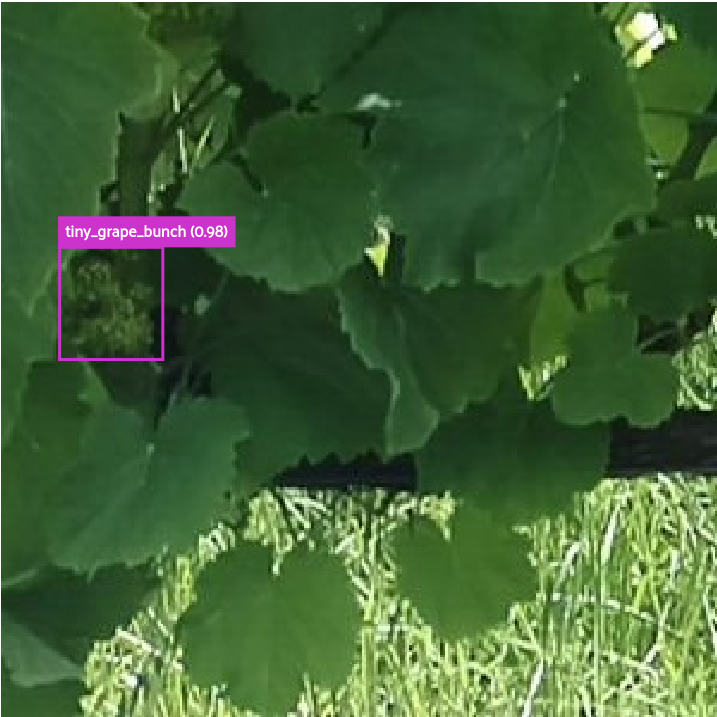}} \hfill
    \caption{Detailed sample image \ref{fig:r2} from figure \ref{fig:image_results} from  Blue -- ground-truth; light green -- NVIDIA RTX3080 TF2; orange -- NVIDIA RTX3090 \ac{tf-trt} \ac{fp32}; brown -- NVIDIA RTX3090 \ac{tf-trt} \ac{fp16}; dark yellow -- NVIDIA RTX3090 \ac{tf-trt} \ac{int8}; red -- AMD-Xilinx Kria KV260; dark green -- AMD-Xilinx ZCU104; pink -- Coral Dev Board \ac{tpu}}
    \label{fig:11b}
\end{figure*}

\begin{figure*}[!htb]
    \centering
    \subfloat[][Ground truth]{\includegraphics[width=0.24\textwidth]{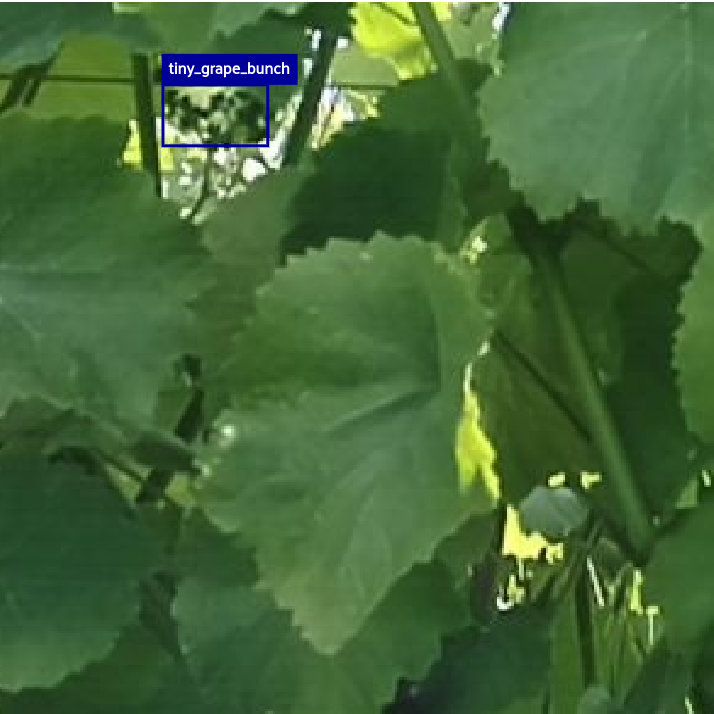}} \hfill
    \subfloat[][RTX3090 TF2]{\includegraphics[width=0.24\textwidth]{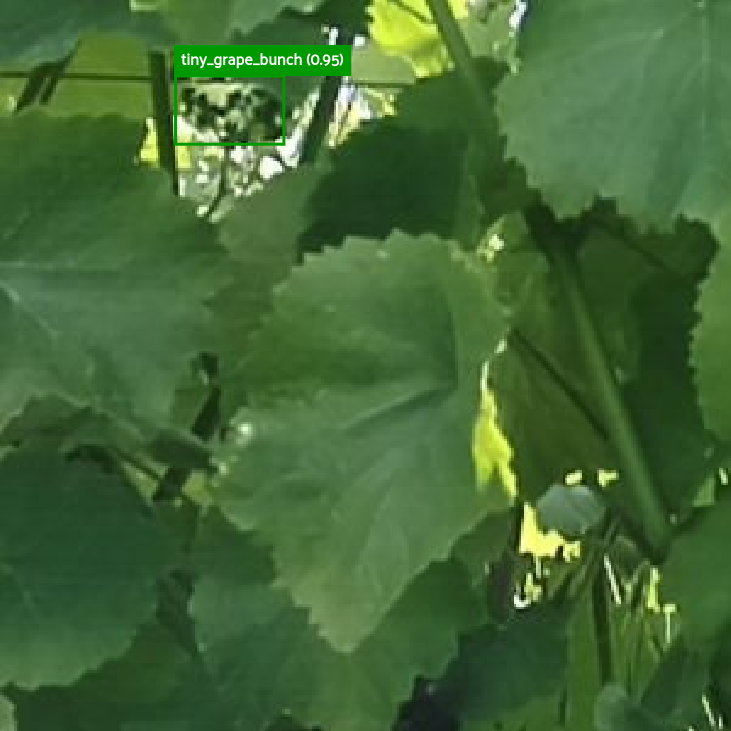}} \hfill
    \subfloat[][TF-TRT FP32]{\includegraphics[width=0.24\textwidth]{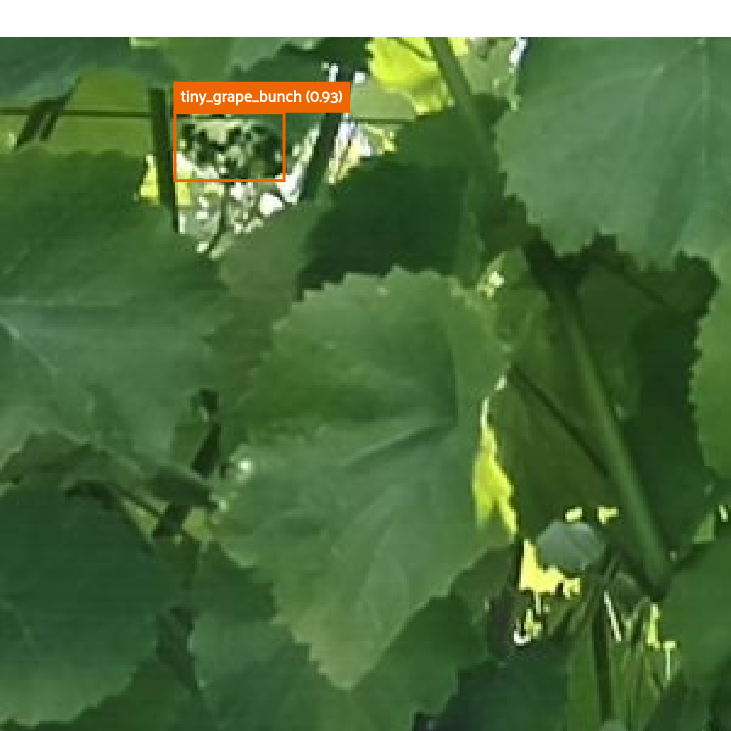}} \hfill
    \subfloat[][TF-TRT FP16]{\includegraphics[width=0.24\textwidth]{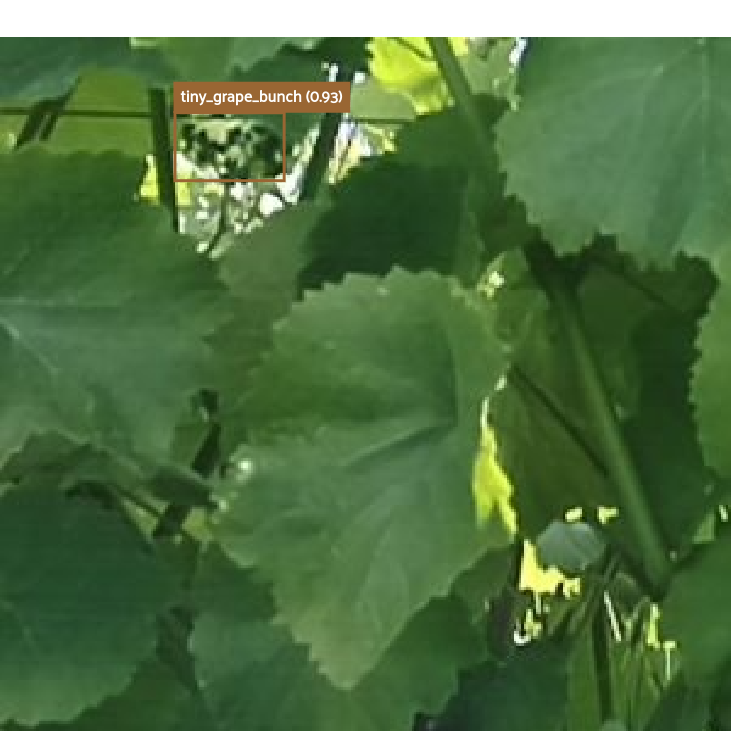}} \hfill
    \subfloat[][TF-TRT INT8]{\includegraphics[width=0.24\textwidth]{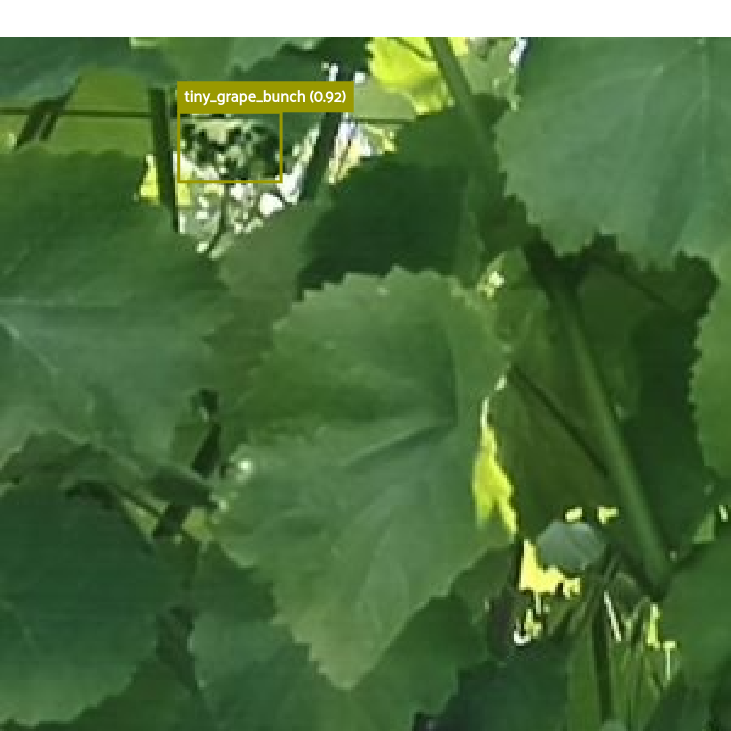}} \hfill
    \subfloat[][KV260]{\includegraphics[width=0.24\textwidth]{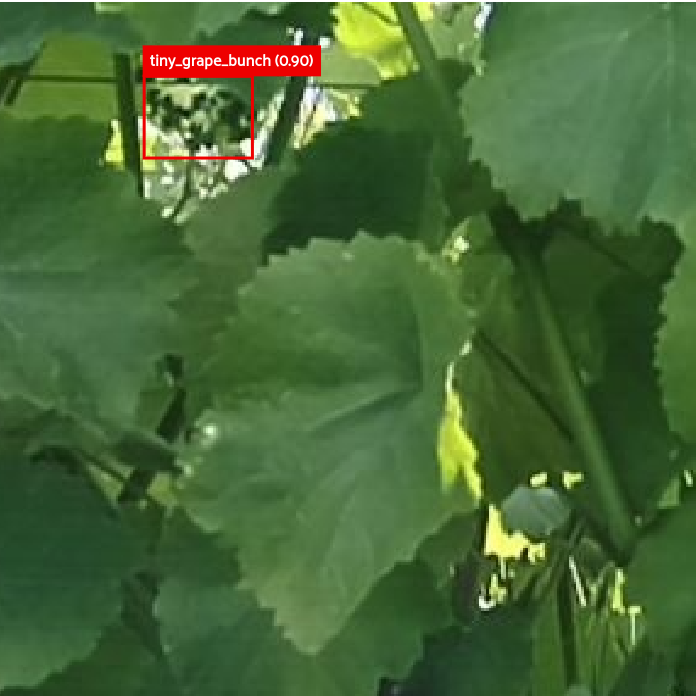}} \hfill
    \subfloat[][ZCU104]{\includegraphics[width=0.24\textwidth]{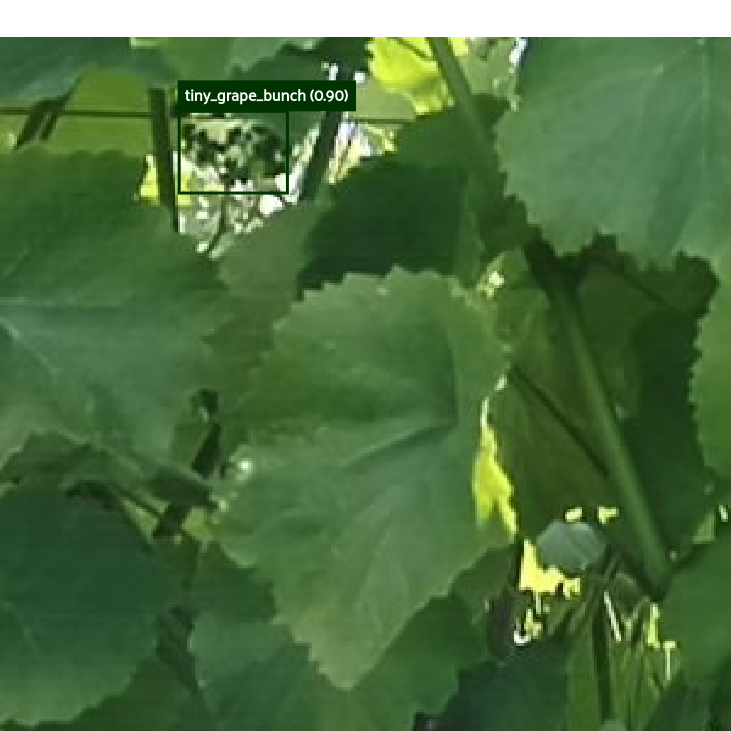}} \hfill
    \subfloat[][TPU]{\includegraphics[width=0.24\textwidth]{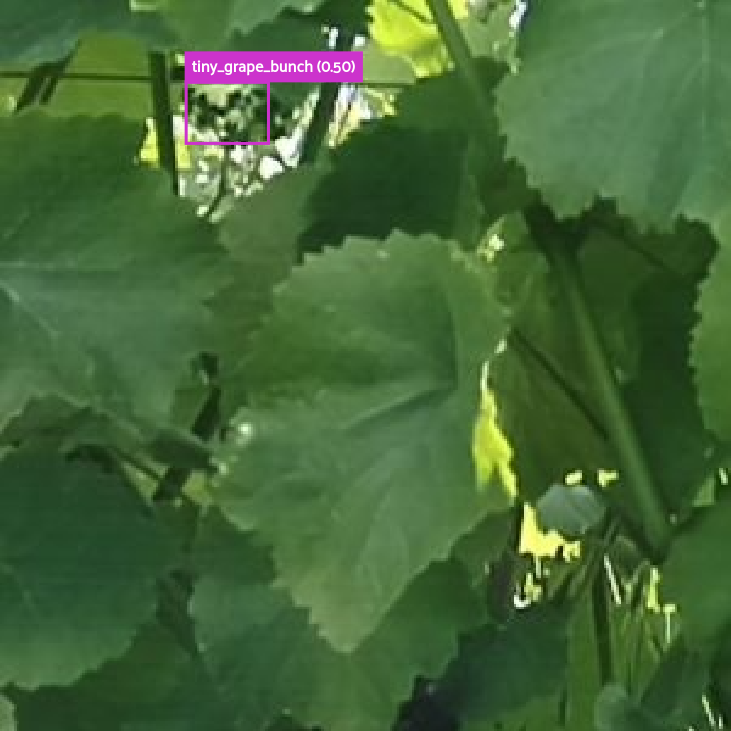}} \hfill
    \caption{Detailed sample image \ref{fig:r3} from figure \ref{fig:image_results} from  Blue -- ground-truth; light green -- NVIDIA RTX3080 TF2; orange -- NVIDIA RTX3090 \ac{tf-trt} \ac{fp32}; brown -- NVIDIA RTX3090 \ac{tf-trt} \ac{fp16}; dark yellow -- NVIDIA RTX3090 \ac{tf-trt} \ac{int8}; red -- AMD-Xilinx Kria KV260; dark green -- AMD-Xilinx ZCU104; pink -- Coral Dev Board \ac{tpu}}
    \label{fig:11c}
\end{figure*}

\begin{figure*}[!htb]
    \centering
    \subfloat[][Ground truth]{\includegraphics[width=0.24\textwidth]{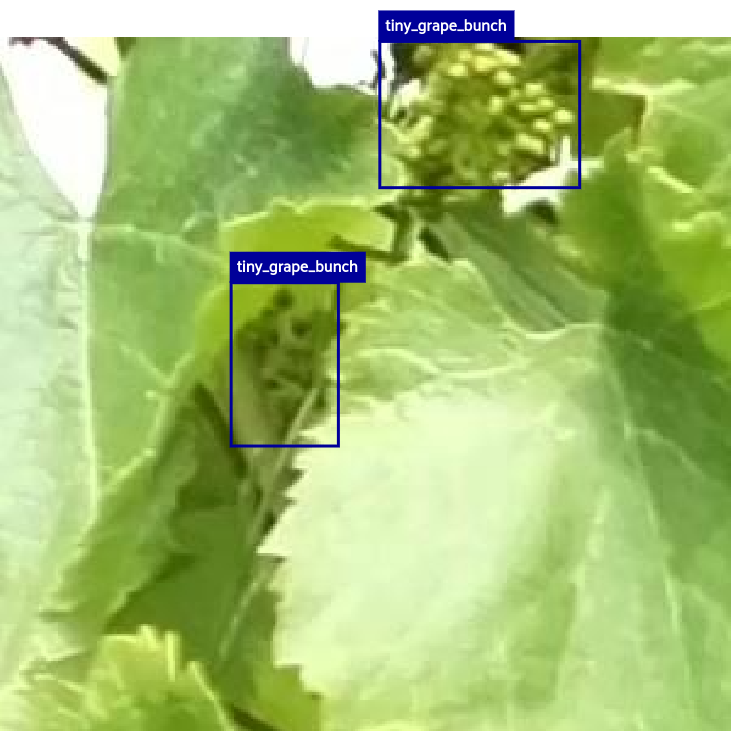}} \hfill
    \subfloat[][RTX3090 TF2]{\includegraphics[width=0.24\textwidth]{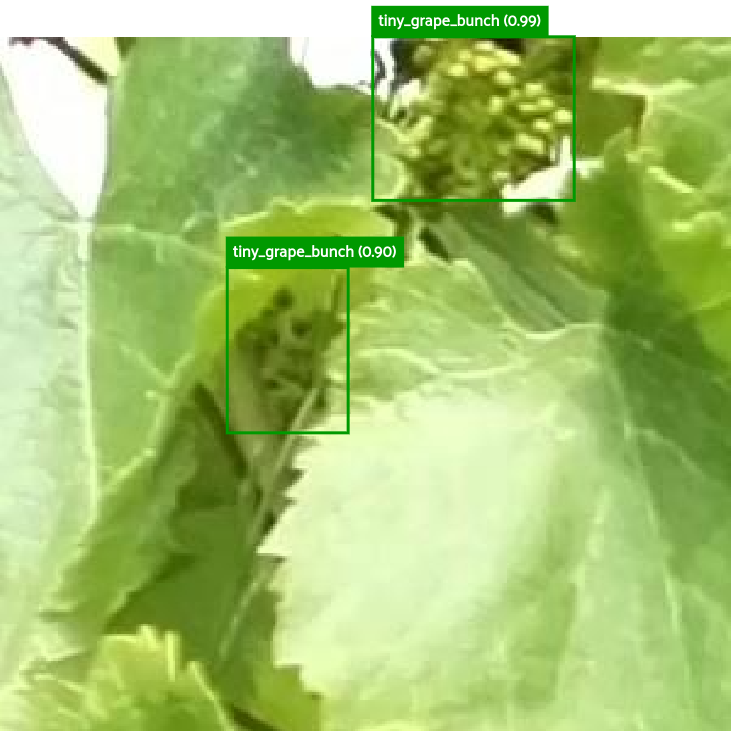}} \hfill
    \subfloat[][TF-TRT FP32]{\includegraphics[width=0.24\textwidth]{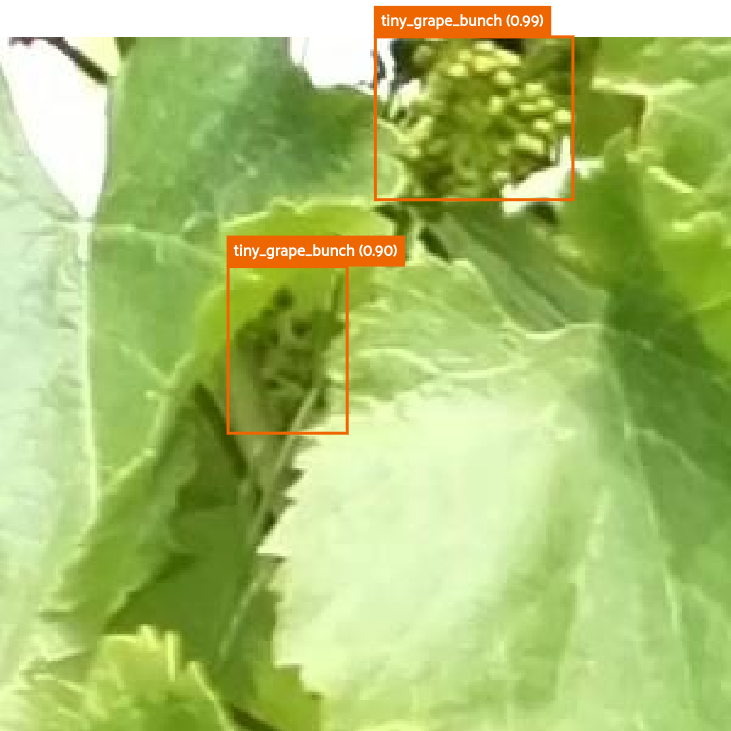}} \hfill
    \subfloat[][TF-TRT FP16]{\includegraphics[width=0.24\textwidth]{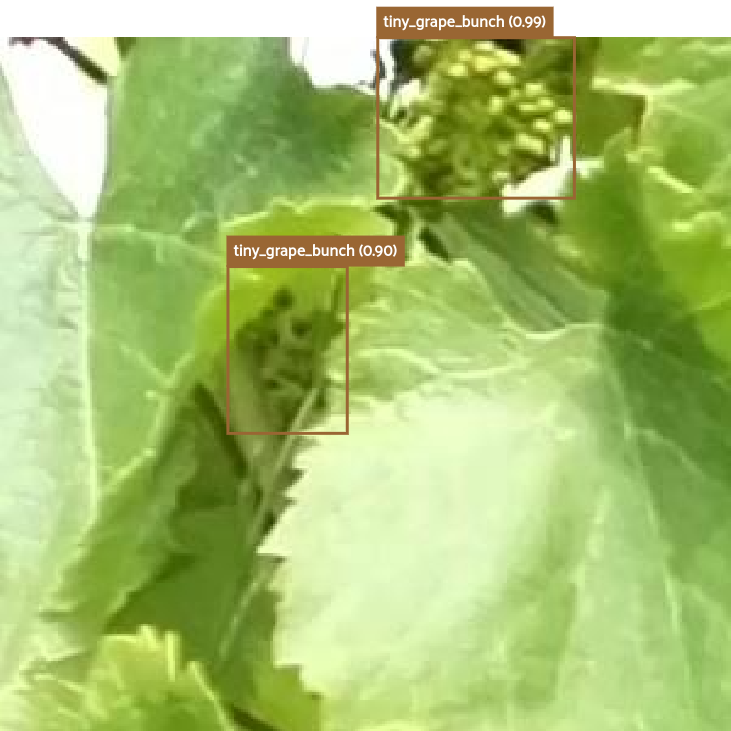}} \hfill
    \subfloat[][TF-TRT INT8]{\includegraphics[width=0.24\textwidth]{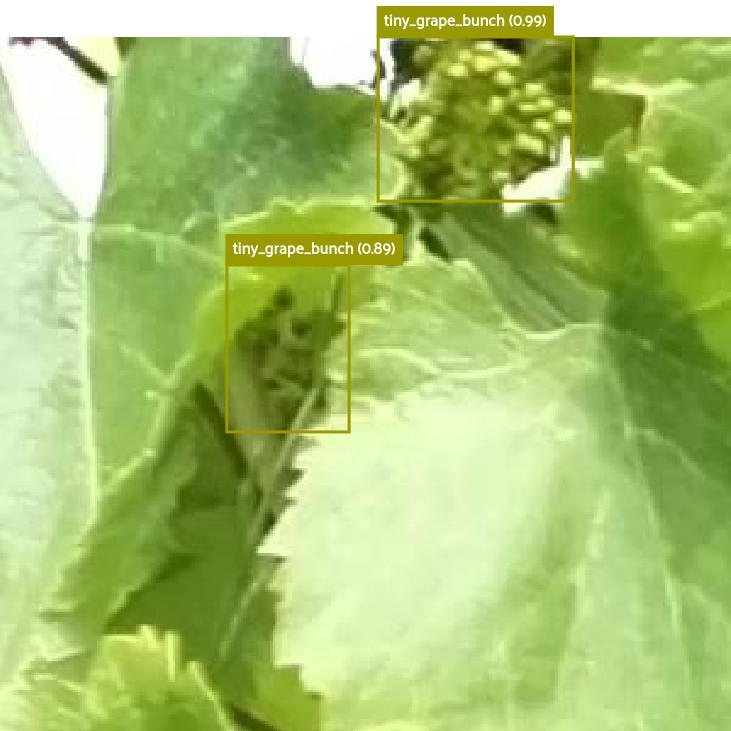}} \hfill
    \subfloat[][KV260]{\includegraphics[width=0.24\textwidth]{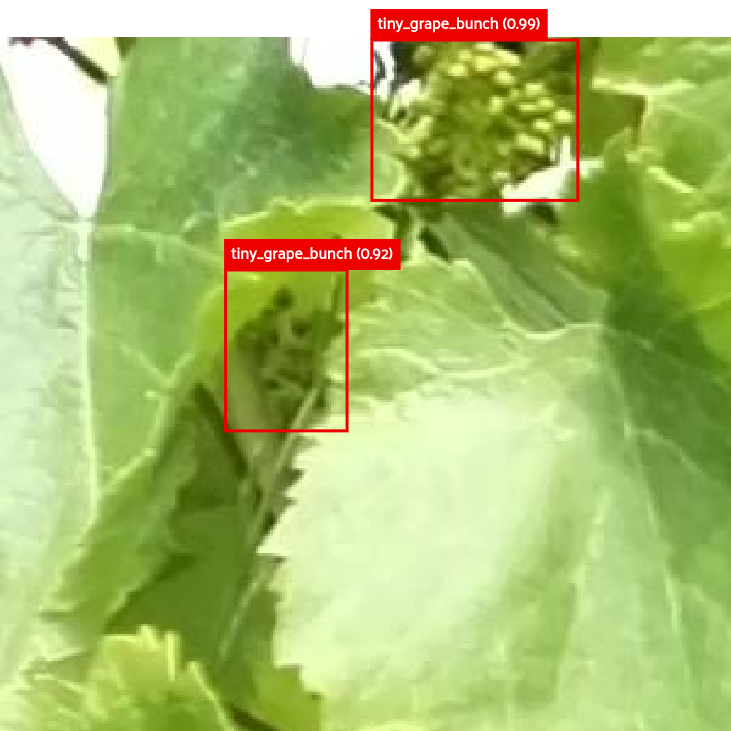}} \hfill
    \subfloat[][ZCU104]{\includegraphics[width=0.24\textwidth]{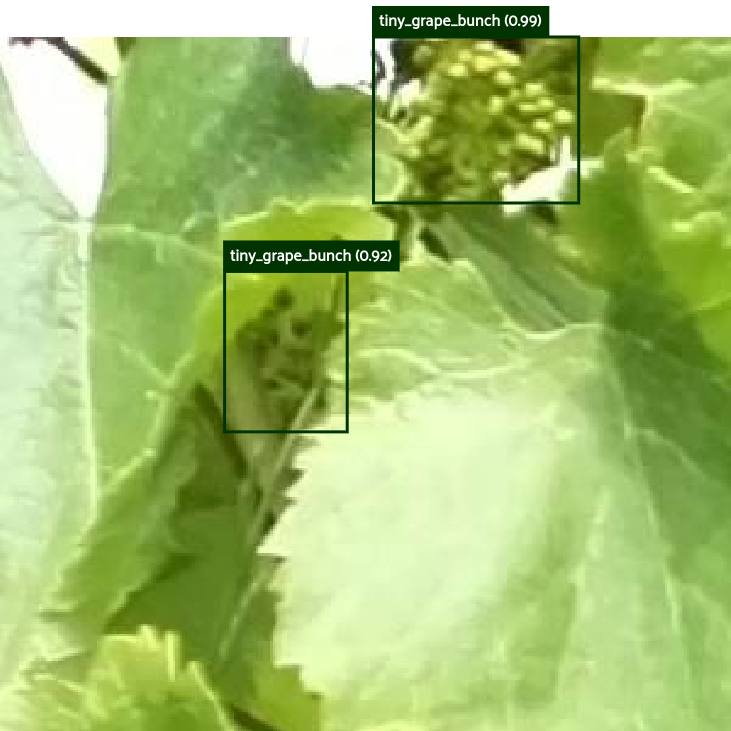}} \hfill
    \subfloat[][TPU]{\includegraphics[width=0.24\textwidth]{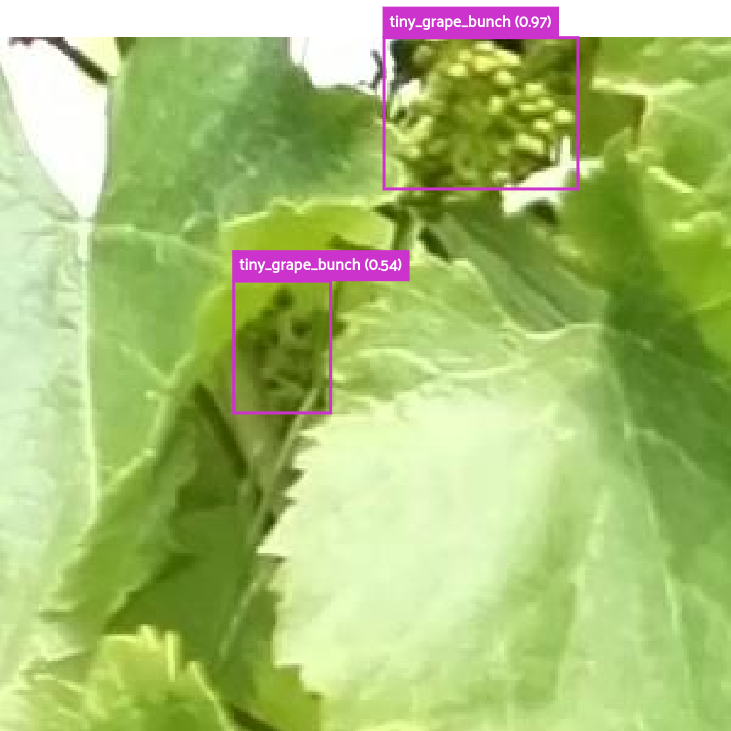}} \hfill
    \caption{Detailed sample image \ref{fig:r4} from figure \ref{fig:image_results} from  Blue -- ground-truth; light green -- NVIDIA RTX3080 TF2; orange -- NVIDIA RTX3090 \ac{tf-trt} \ac{fp32}; brown -- NVIDIA RTX3090 \ac{tf-trt} \ac{fp16}; dark yellow -- NVIDIA RTX3090 \ac{tf-trt} \ac{int8}; red -- AMD-Xilinx Kria KV260; dark green -- AMD-Xilinx ZCU104; pink -- Coral Dev Board \ac{tpu}}
    \label{fig:11d}
\end{figure*}

\begin{figure*}[!htb]
    \centering
    \subfloat[][Ground truth]{\includegraphics[width=0.24\textwidth]{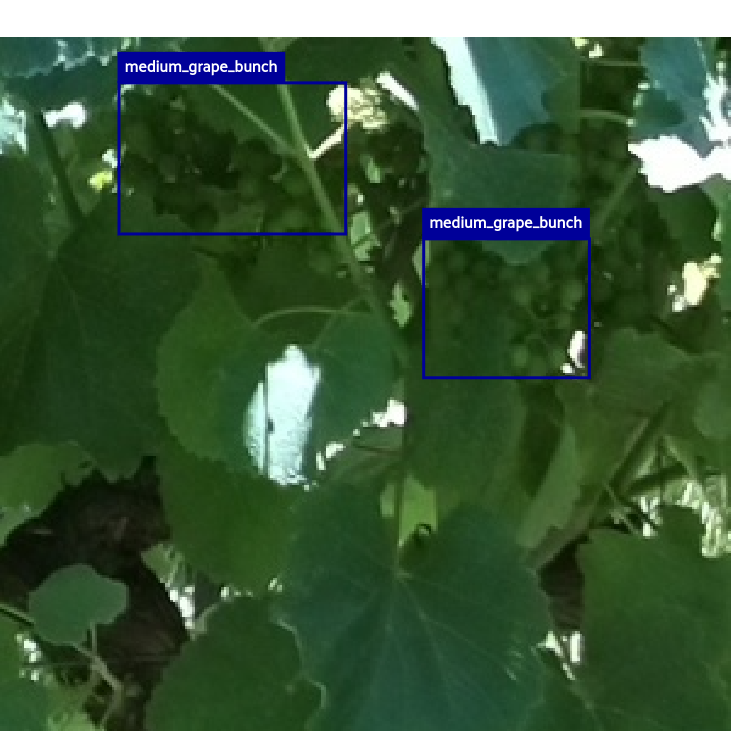}} \hfill
    \subfloat[][RTX3090 TF2]{\includegraphics[width=0.24\textwidth]{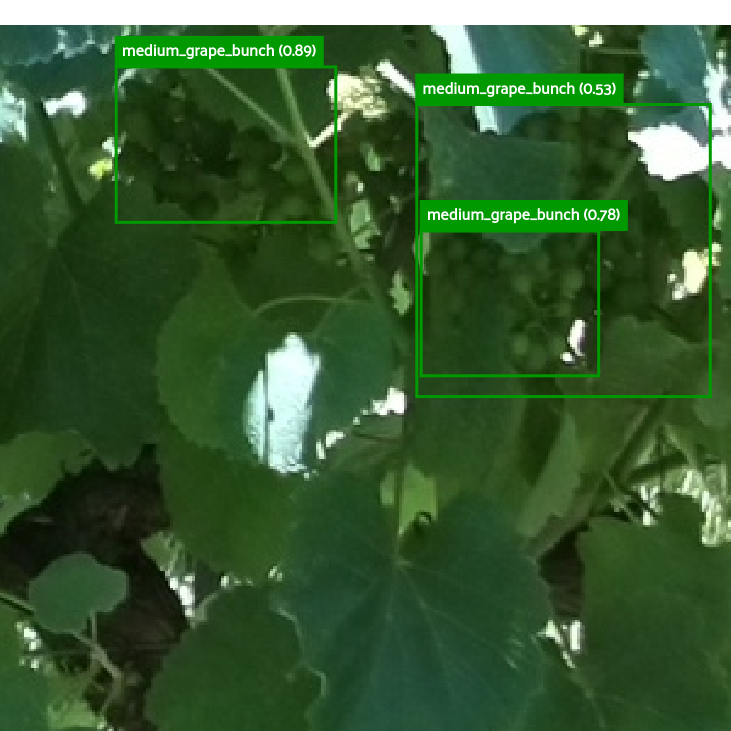}} \hfill
    \subfloat[][TF-TRT FP32]{\includegraphics[width=0.24\textwidth]{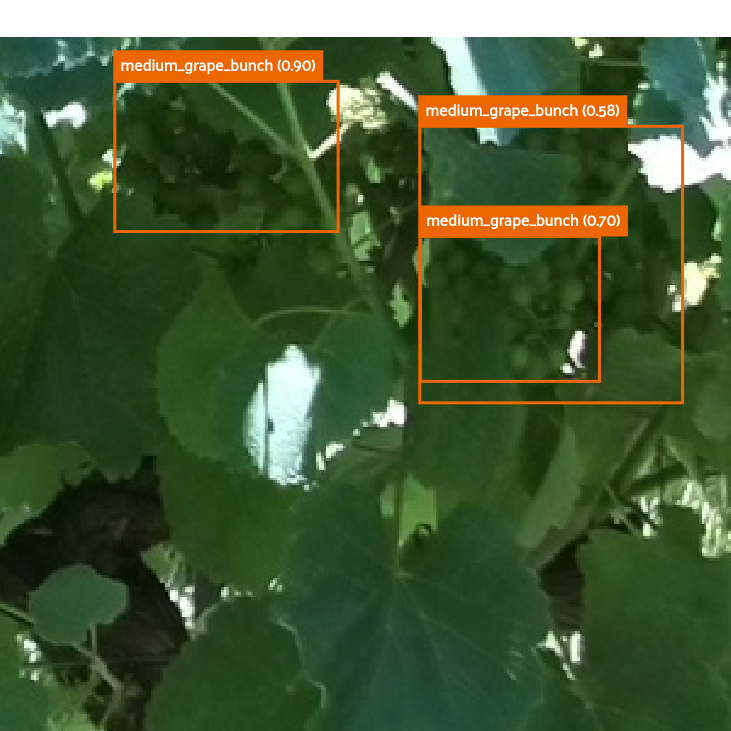}} \hfill
    \subfloat[][TF-TRT FP16]{\includegraphics[width=0.24\textwidth]{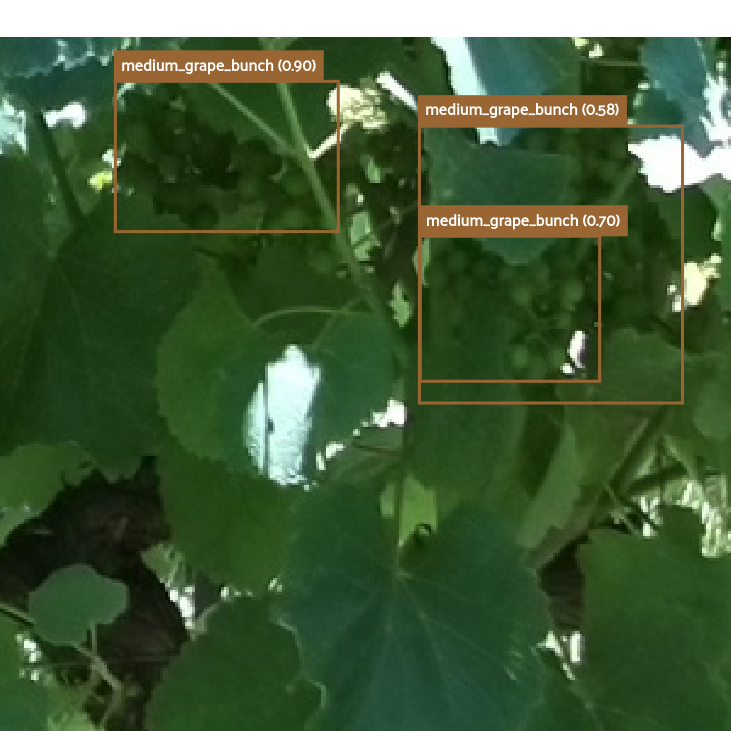}} \hfill
    \subfloat[][TF-TRT INT8]{\includegraphics[width=0.24\textwidth]{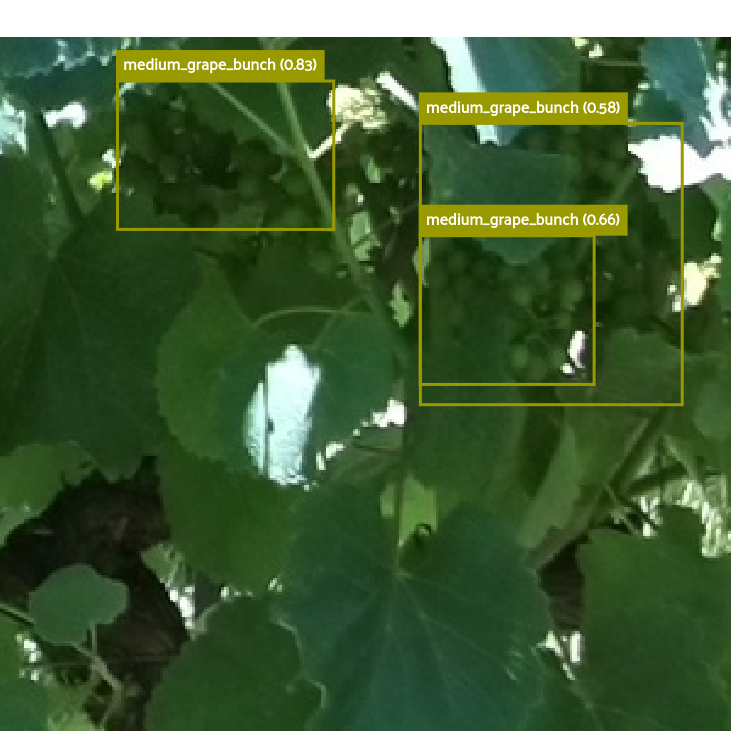}} \hfill
    \subfloat[][KV260]{\includegraphics[width=0.24\textwidth]{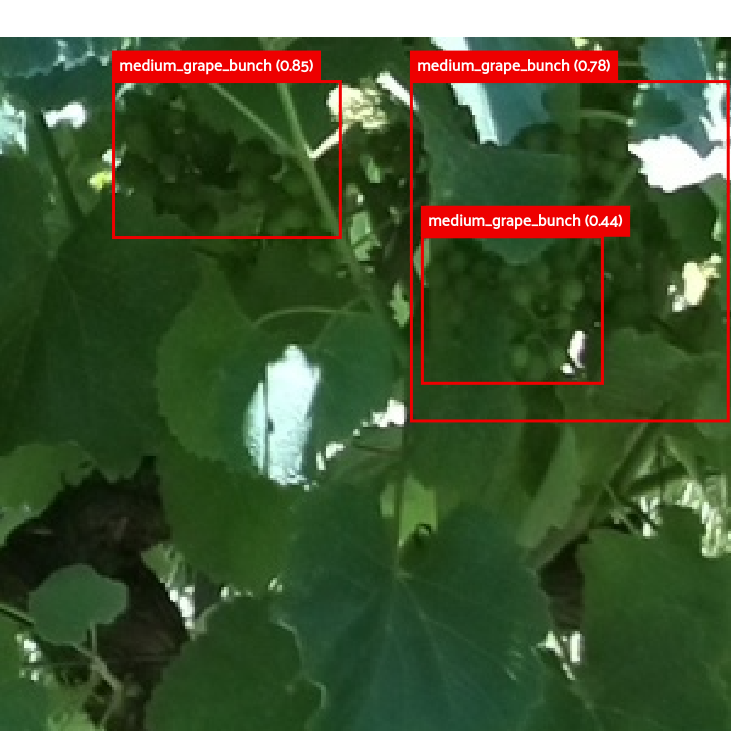}} \hfill
    \subfloat[][ZCU104]{\includegraphics[width=0.24\textwidth]{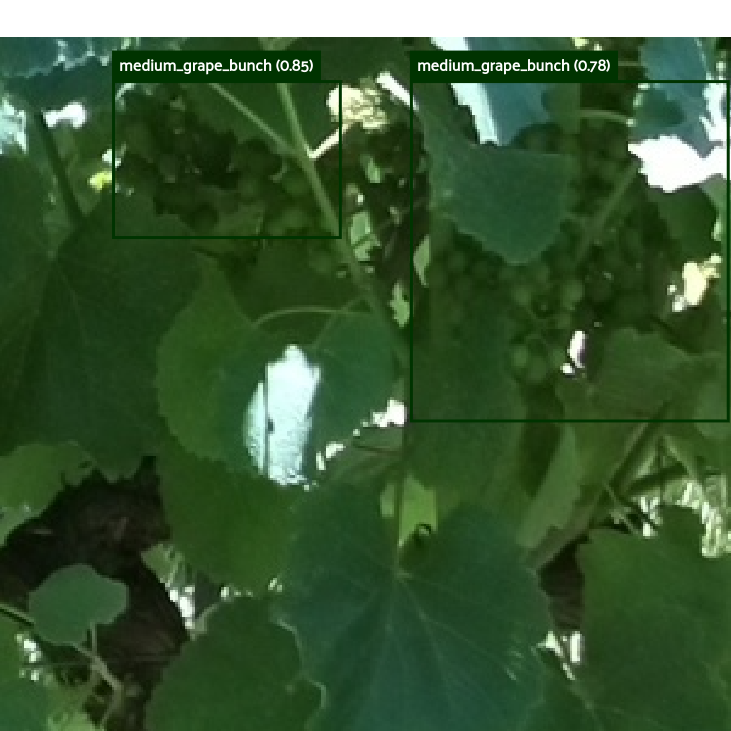}} \hfill
    \subfloat[][TPU]{\includegraphics[width=0.24\textwidth]{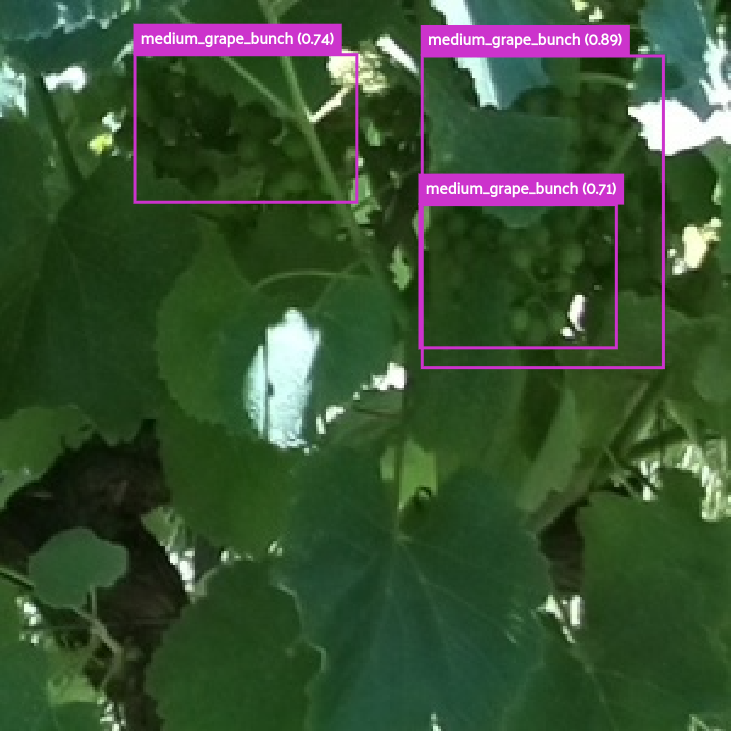}} \hfill
    \caption{Detailed sample image \ref{fig:r5} from figure \ref{fig:image_results} from  Blue -- ground-truth; light green -- NVIDIA RTX3080 TF2; orange -- NVIDIA RTX3090 \ac{tf-trt} \ac{fp32}; brown -- NVIDIA RTX3090 \ac{tf-trt} \ac{fp16}; dark yellow -- NVIDIA RTX3090 \ac{tf-trt} \ac{int8}; red -- AMD-Xilinx Kria KV260; dark green -- AMD-Xilinx ZCU104; pink -- Coral Dev Board \ac{tpu}}
    \label{fig:11e}
\end{figure*}

\begin{figure*}[!htb]
    \centering
    \subfloat[][Ground truth]{\includegraphics[width=0.24\textwidth]{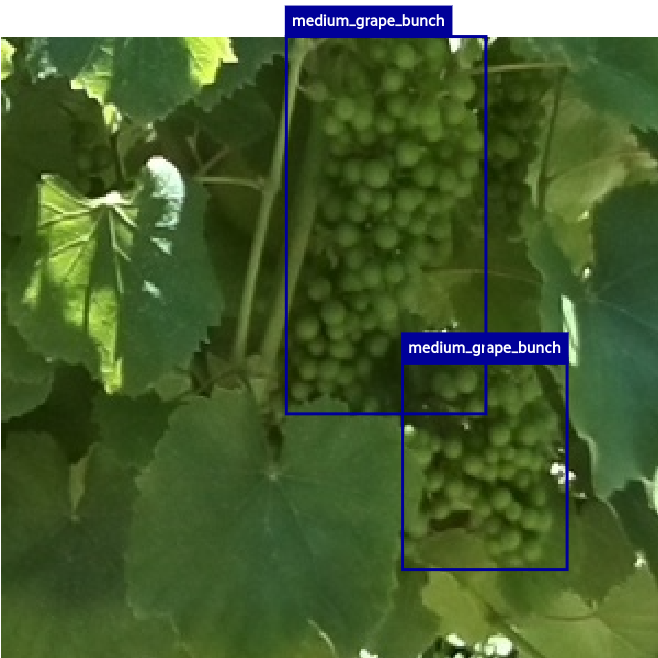}} \hfill
    \subfloat[][RTX3090 TF2]{\includegraphics[width=0.24\textwidth]{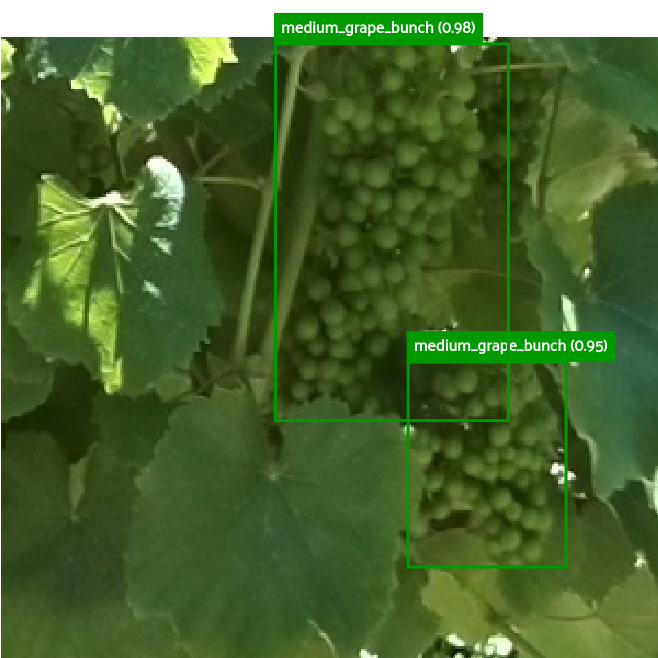}} \hfill
    \subfloat[][TF-TRT FP32]{\includegraphics[width=0.24\textwidth]{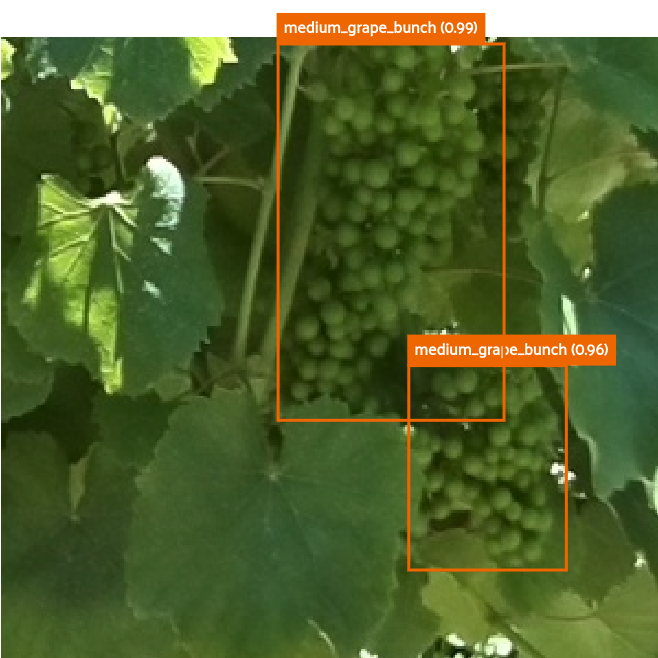}} \hfill
    \subfloat[][TF-TRT FP16]{\includegraphics[width=0.24\textwidth]{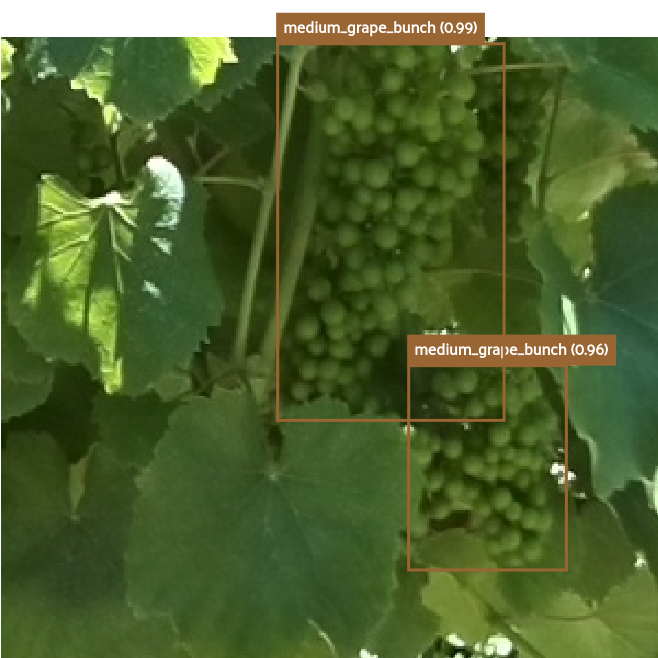}} \hfill
    \subfloat[][TF-TRT INT8]{\includegraphics[width=0.24\textwidth]{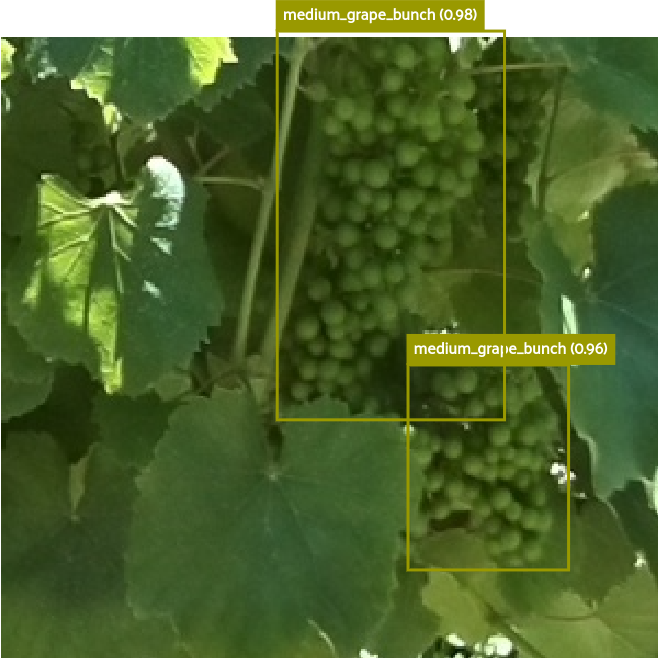}} \hfill
    \subfloat[][KV260]{\includegraphics[width=0.24\textwidth]{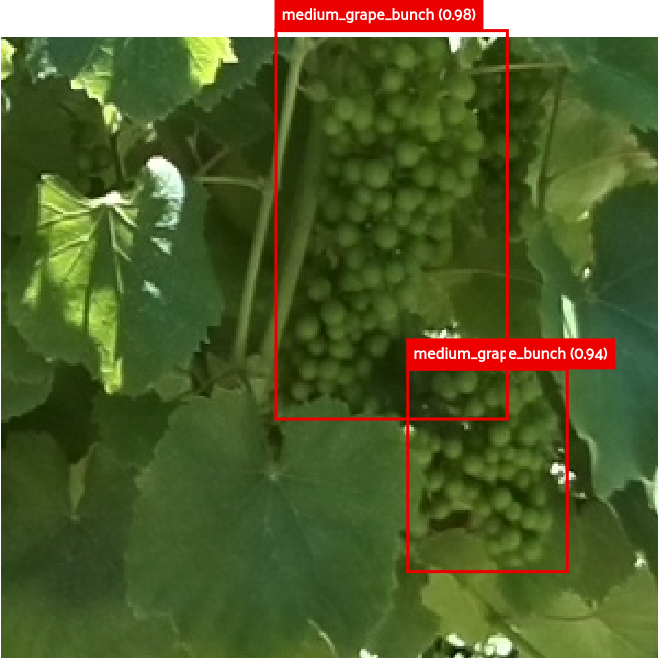}} \hfill
    \subfloat[][ZCU104]{\includegraphics[width=0.24\textwidth]{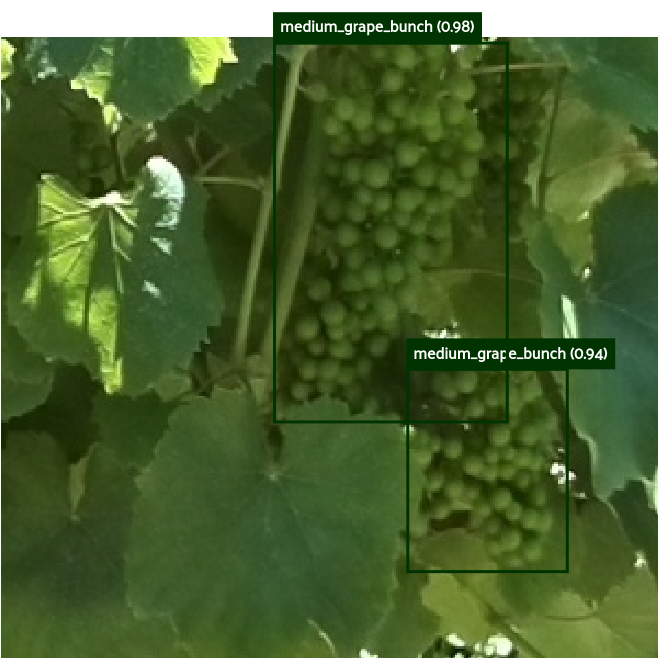}} \hfill
    \subfloat[][TPU]{\includegraphics[width=0.24\textwidth]{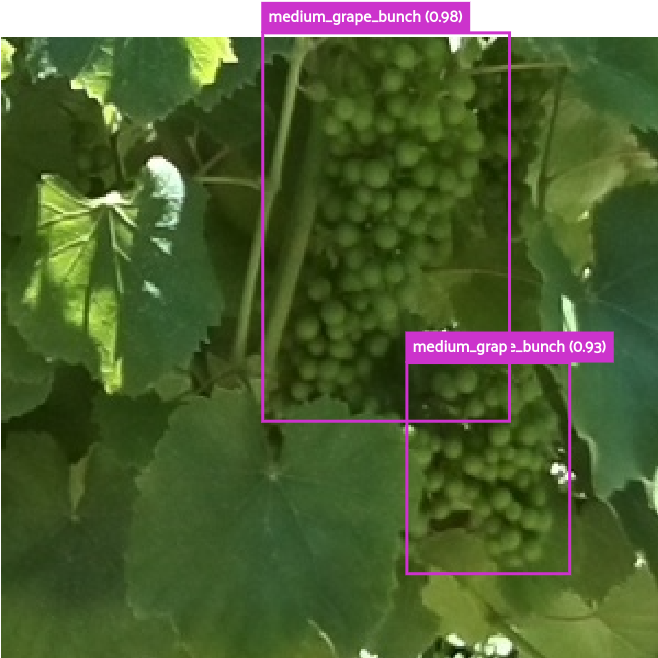}} \hfill
    \caption{Detailed sample image \ref{fig:r6} from figure \ref{fig:image_results} from  Blue -- ground-truth; light green -- NVIDIA RTX3080 TF2; orange -- NVIDIA RTX3090 \ac{tf-trt} \ac{fp32}; brown -- NVIDIA RTX3090 \ac{tf-trt} \ac{fp16}; dark yellow -- NVIDIA RTX3090 \ac{tf-trt} \ac{int8}; red -- AMD-Xilinx Kria KV260; dark green -- AMD-Xilinx ZCU104; pink -- Coral Dev Board \ac{tpu}}
    \label{fig:11f}
\end{figure*}

\begin{figure*}[!htb]
    \centering
    \subfloat[][Ground truth]{\includegraphics[width=0.24\textwidth]{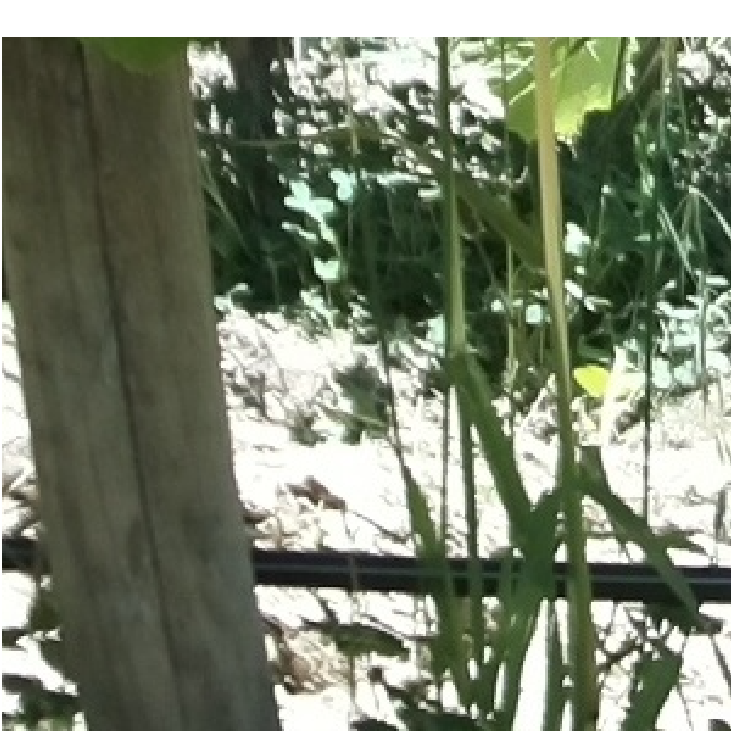}} \hfill
    \subfloat[][RTX3090 TF2]{\includegraphics[width=0.24\textwidth]{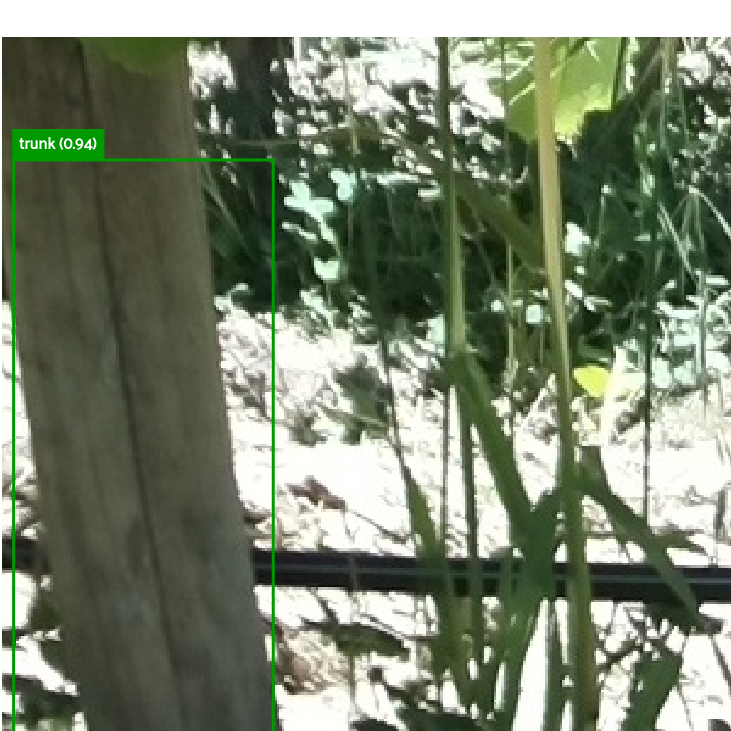}} \hfill
    \subfloat[][TF-TRT FP32]{\includegraphics[width=0.24\textwidth]{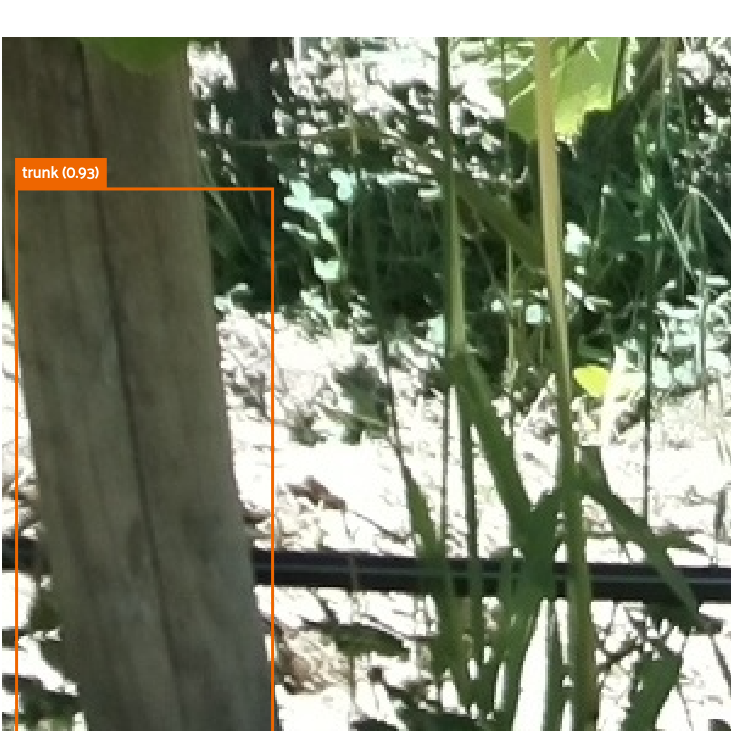}} \hfill
    \subfloat[][TF-TRT FP16]{\includegraphics[width=0.24\textwidth]{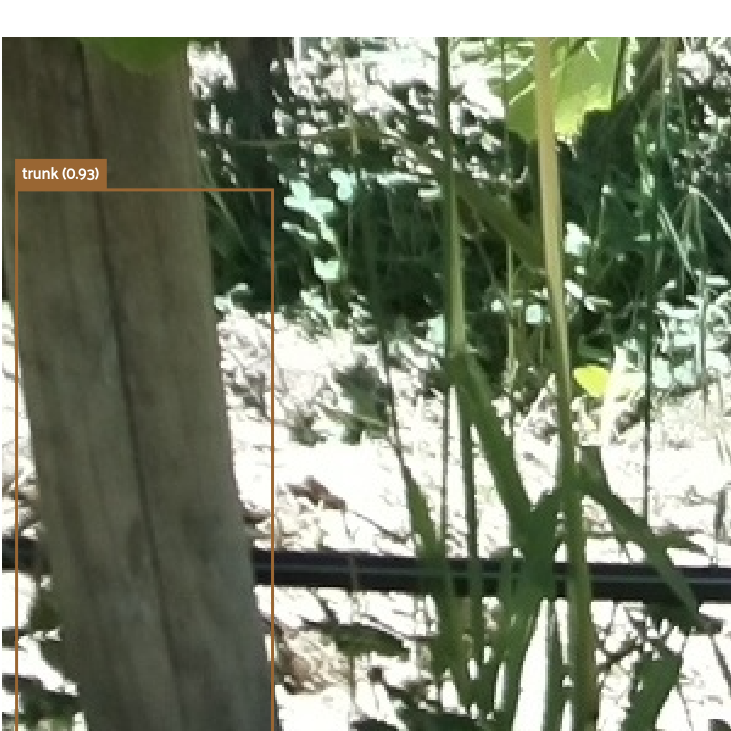}} \hfill
    \subfloat[][TF-TRT INT8]{\includegraphics[width=0.24\textwidth]{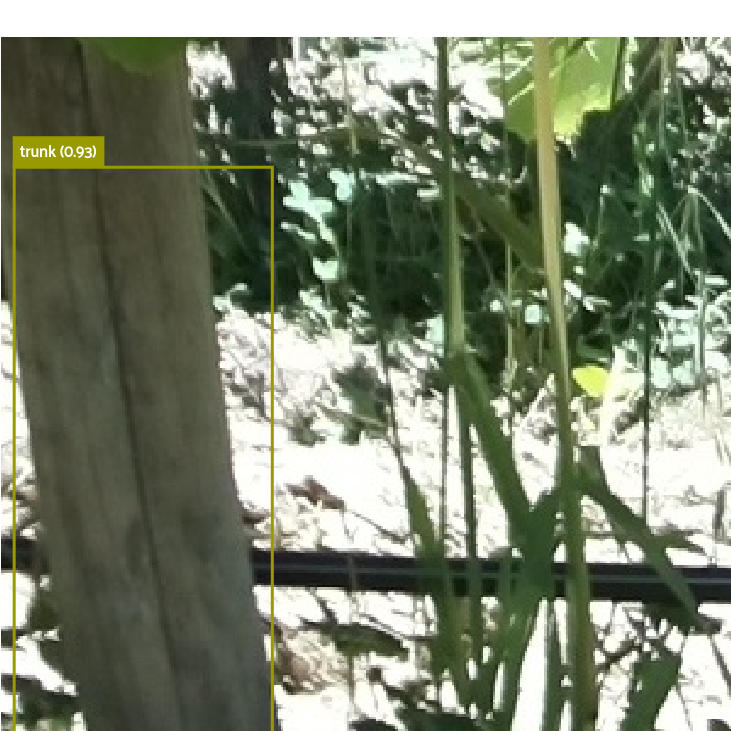}} \hfill
    \subfloat[][KV260]{\includegraphics[width=0.24\textwidth]{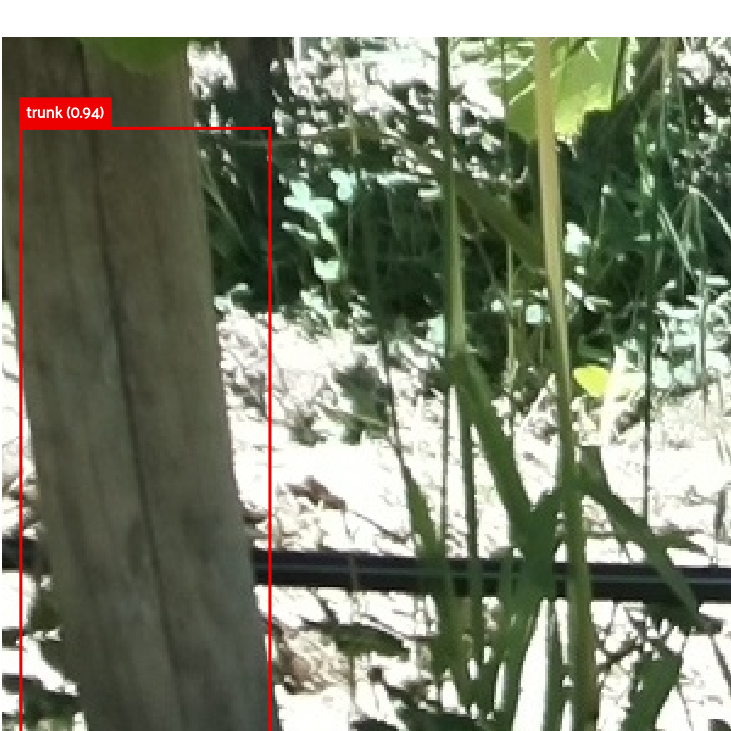}} \hfill
    \subfloat[][ZCU104]{\includegraphics[width=0.24\textwidth]{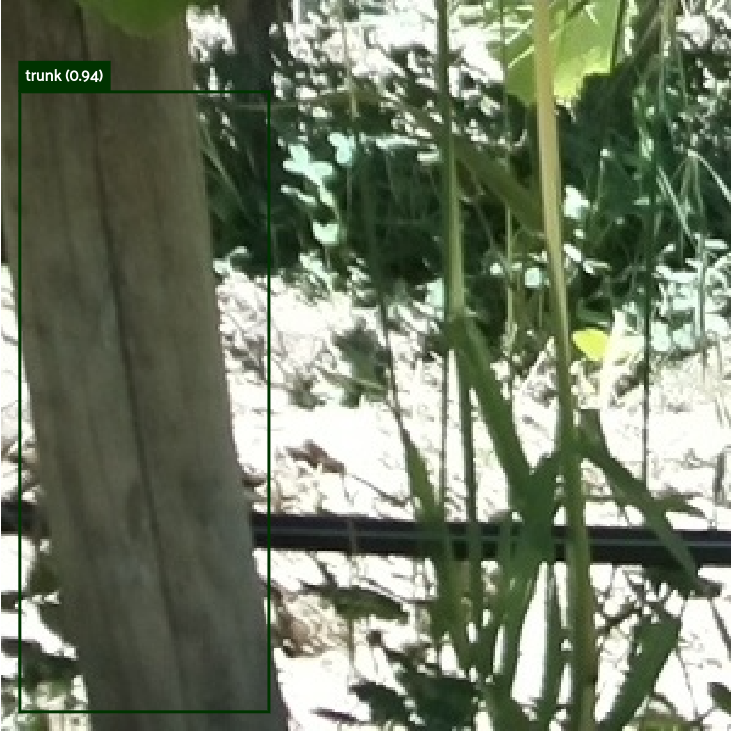}} \hfill
    \subfloat[][TPU]{\includegraphics[width=0.24\textwidth]{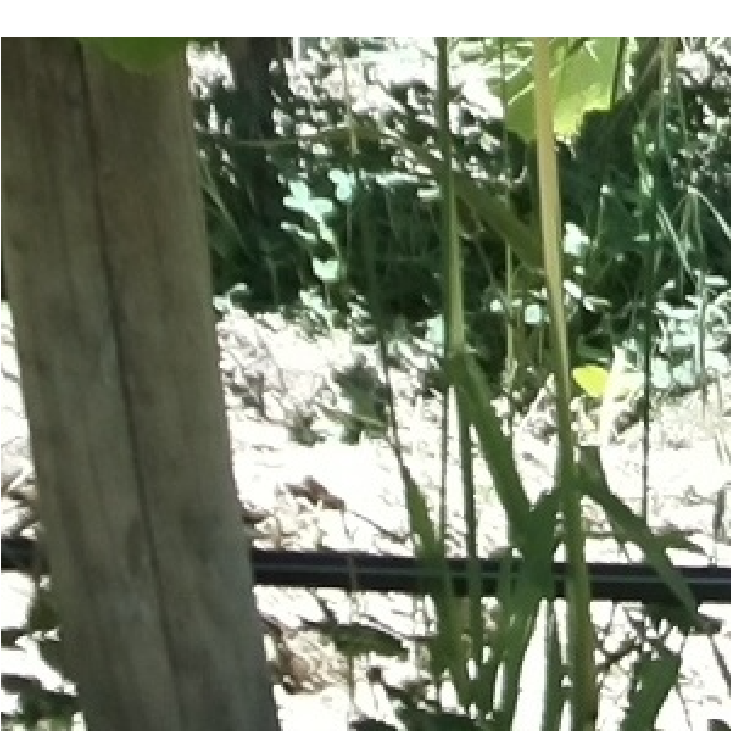}} \hfill
    \caption{Detailed sample image \ref{fig:r7} from figure \ref{fig:image_results} from  Blue -- ground-truth; light green -- NVIDIA RTX3080 TF2; orange -- NVIDIA RTX3090 \ac{tf-trt} \ac{fp32}; brown -- NVIDIA RTX3090 \ac{tf-trt} \ac{fp16}; dark yellow -- NVIDIA RTX3090 \ac{tf-trt} \ac{int8}; red -- AMD-Xilinx Kria KV260; dark green -- AMD-Xilinx ZCU104; pink -- Coral Dev Board \ac{tpu}}
    \label{fig:11g}
\end{figure*}

\begin{figure*}[!htb]
    \centering
    \subfloat[][Ground truth]{\includegraphics[width=0.24\textwidth]{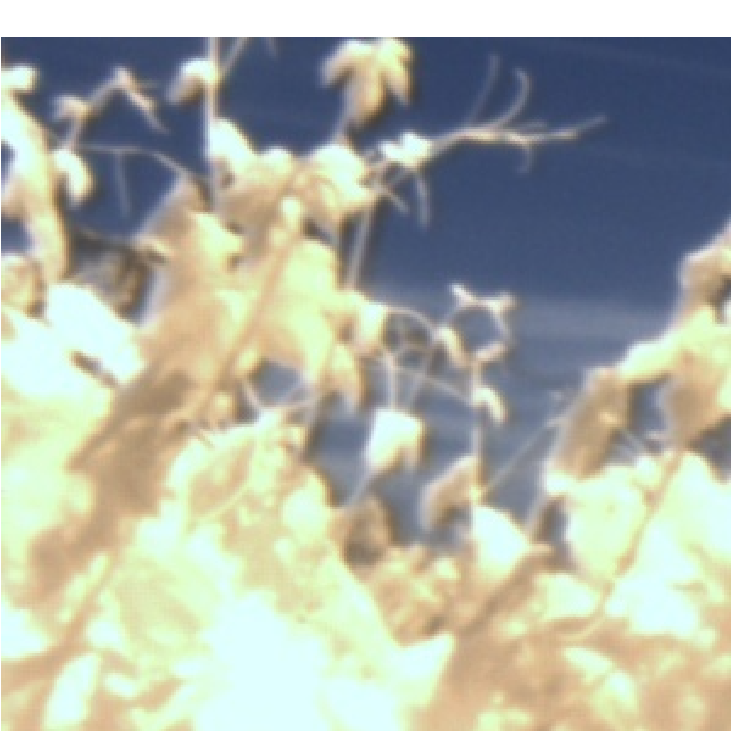}} \hfill
    \subfloat[][RTX3090 TF2]{\includegraphics[width=0.24\textwidth]{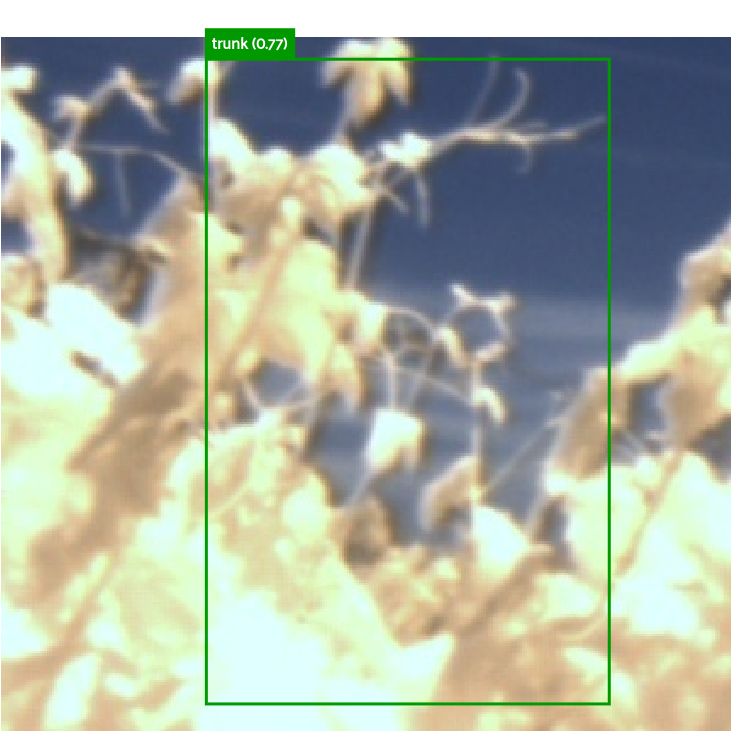}} \hfill
    \subfloat[][TF-TRT FP32]{\includegraphics[width=0.24\textwidth]{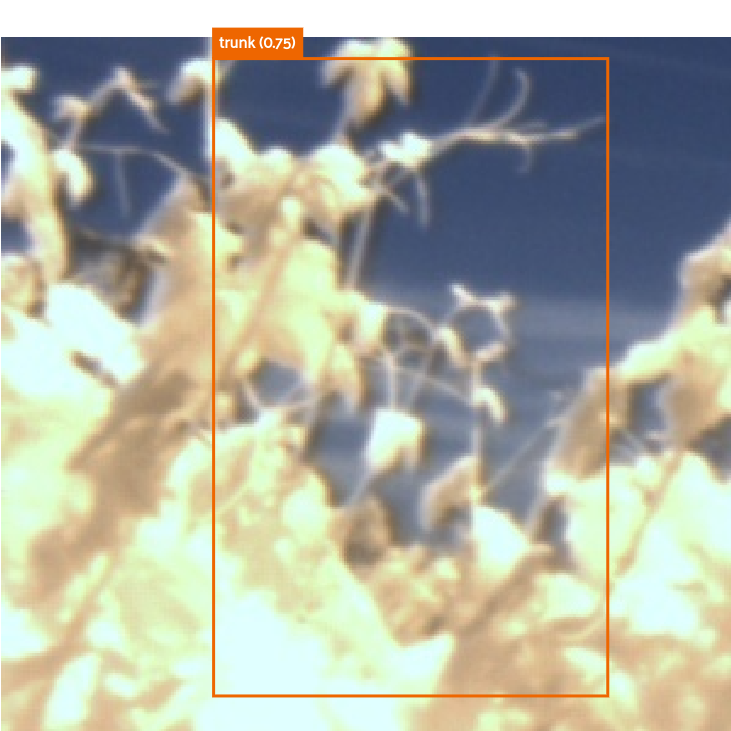}} \hfill
    \subfloat[][TF-TRT FP16]{\includegraphics[width=0.24\textwidth]{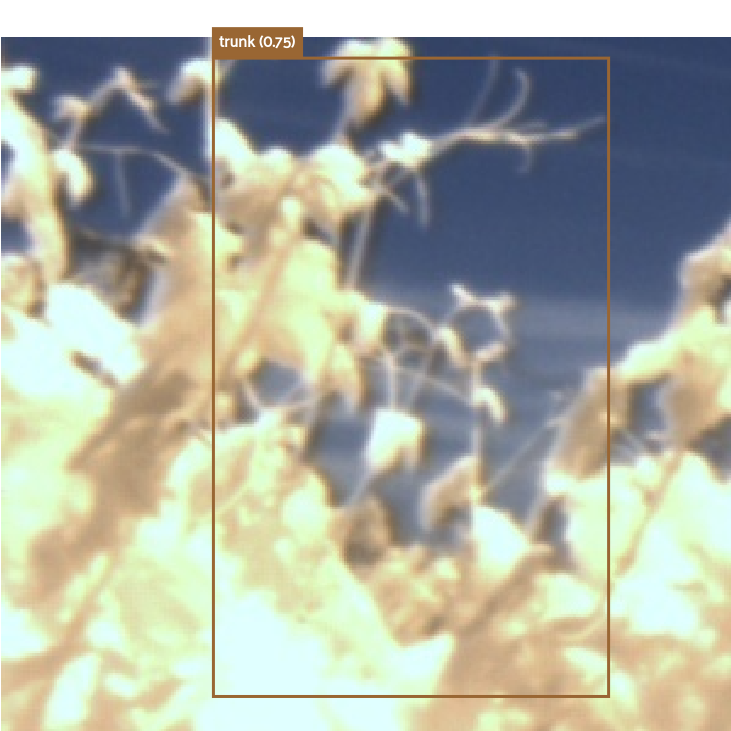}} \hfill
    \subfloat[][TF-TRT INT8]{\includegraphics[width=0.24\textwidth]{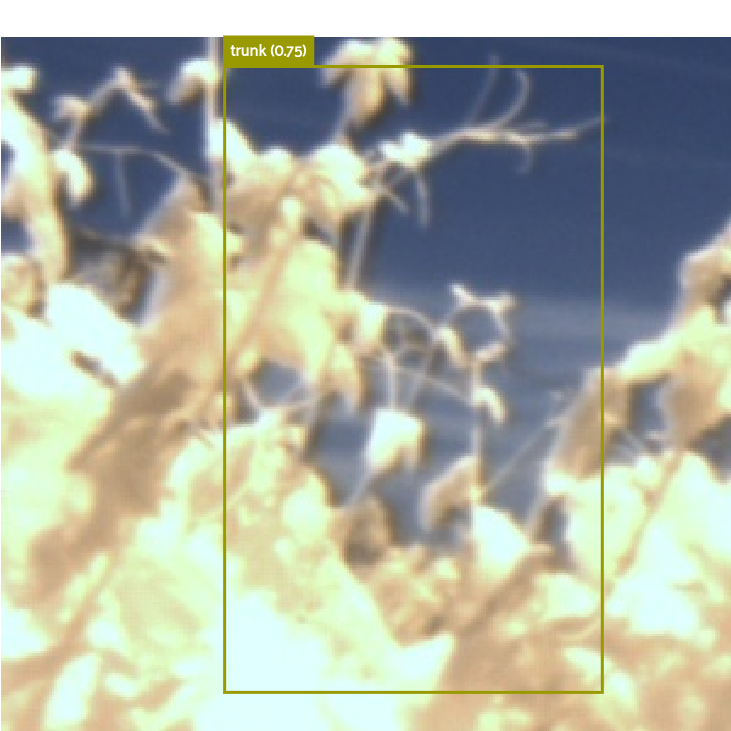}} \hfill
    \subfloat[][KV260]{\includegraphics[width=0.24\textwidth]{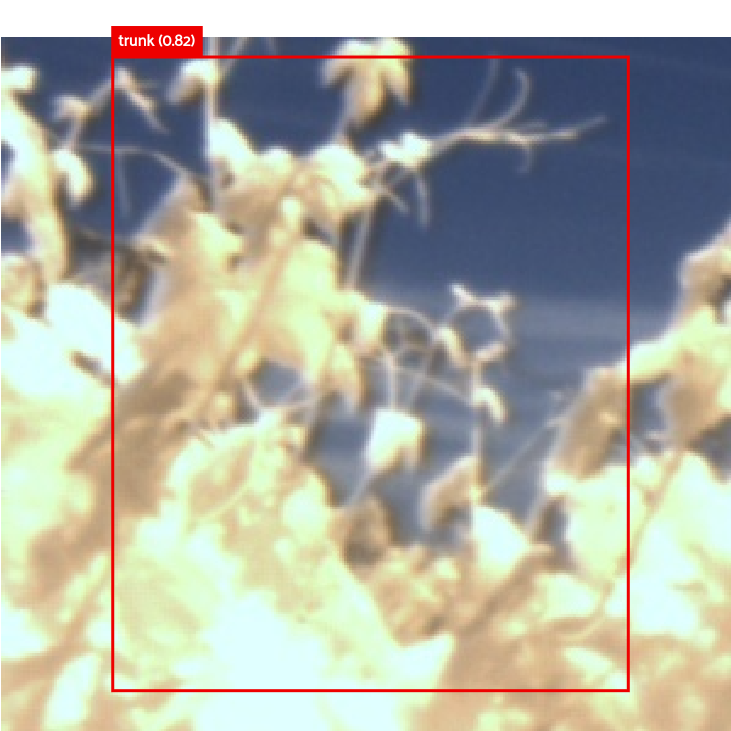}} \hfill
    \subfloat[][ZCU104]{\includegraphics[width=0.24\textwidth]{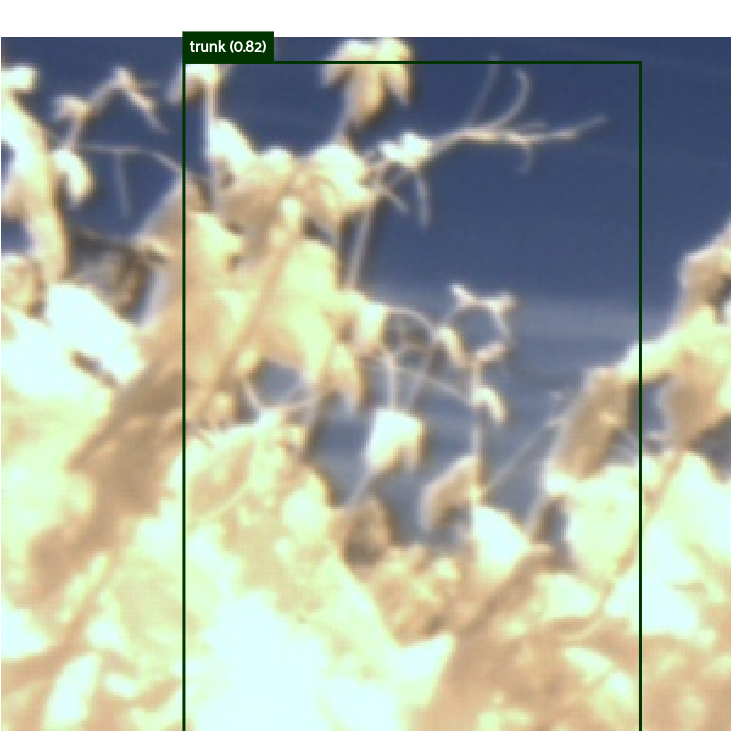}} \hfill
    \subfloat[][TPU]{\includegraphics[width=0.24\textwidth]{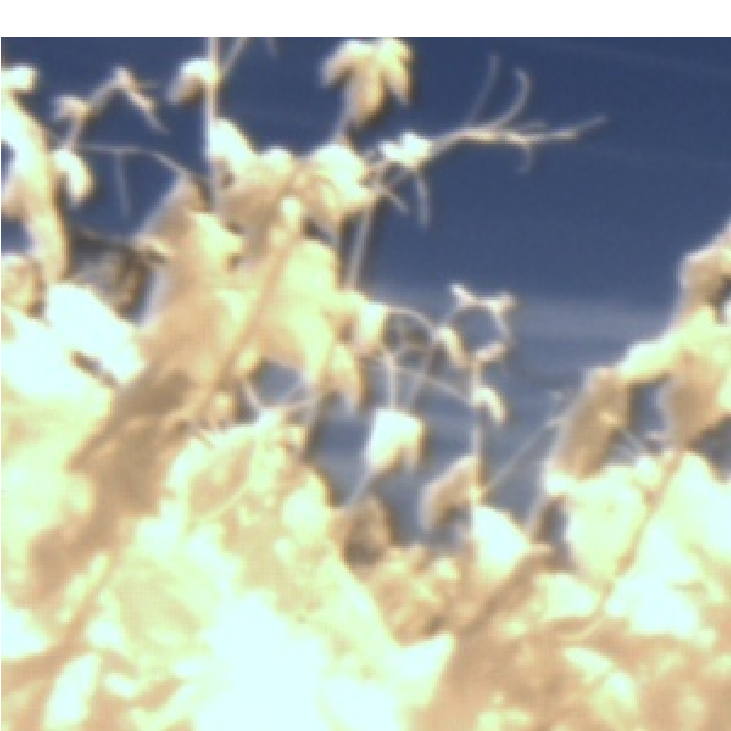}} \hfill
    \caption{Detailed sample image \ref{fig:r8} from figure \ref{fig:image_results} from  Blue -- ground-truth; light green -- NVIDIA RTX3080 TF2; orange -- NVIDIA RTX3090 \ac{tf-trt} \ac{fp32}; brown -- NVIDIA RTX3090 \ac{tf-trt} \ac{fp16}; dark yellow -- NVIDIA RTX3090 \ac{tf-trt} \ac{int8}; red -- AMD-Xilinx Kria KV260; dark green -- AMD-Xilinx ZCU104; pink -- Coral Dev Board \ac{tpu}}
    \label{fig:11h}
\end{figure*}

\begin{figure*}[!htb]
    \centering
    \subfloat[][Ground truth]{\includegraphics[width=0.24\textwidth]{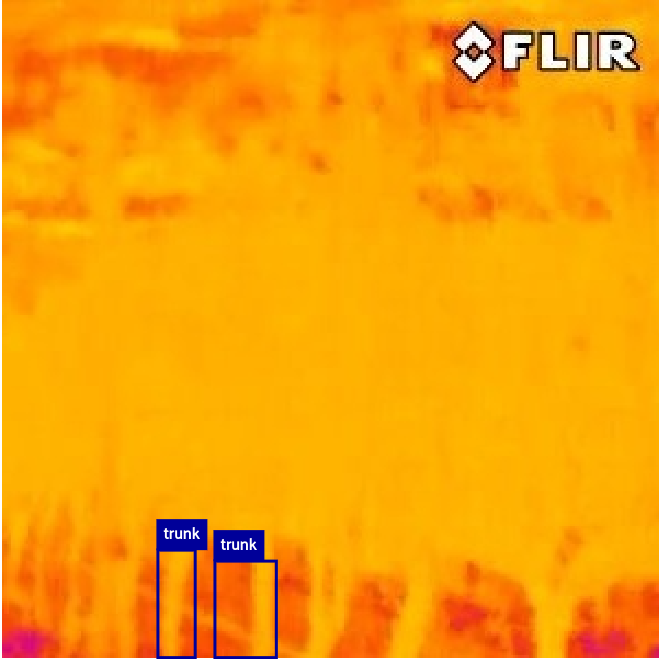}} \hfill
    \subfloat[][RTX3090 TF2]{\includegraphics[width=0.24\textwidth]{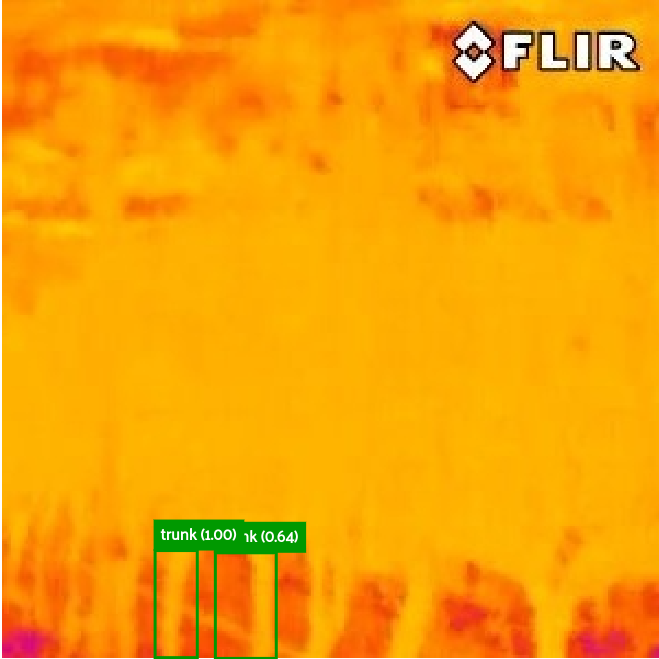}} \hfill
    \subfloat[][TF-TRT FP32]{\includegraphics[width=0.24\textwidth]{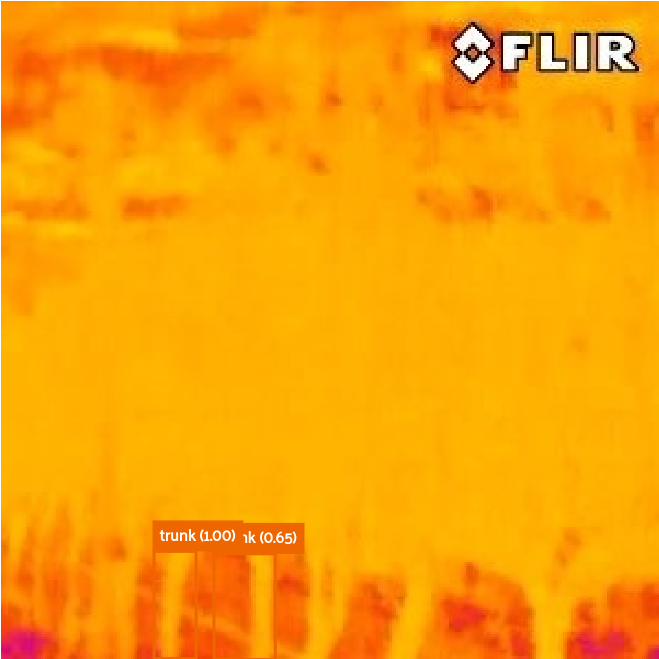}} \hfill
    \subfloat[][TF-TRT FP16]{\includegraphics[width=0.24\textwidth]{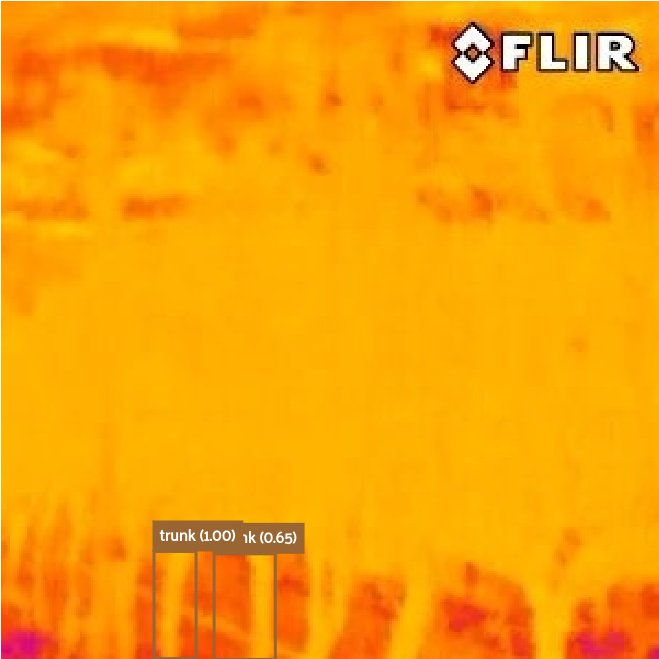}} \hfill
    \subfloat[][TF-TRT INT8]{\includegraphics[width=0.24\textwidth]{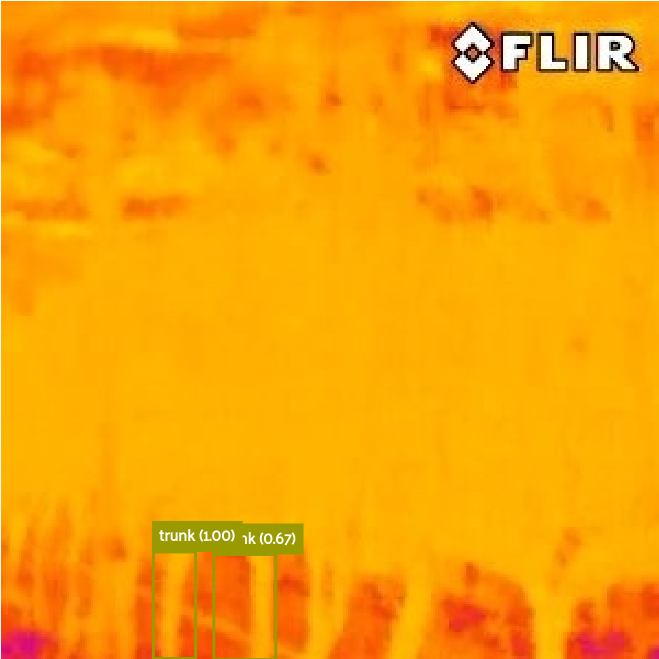}} \hfill
    \subfloat[][KV260]{\includegraphics[width=0.24\textwidth]{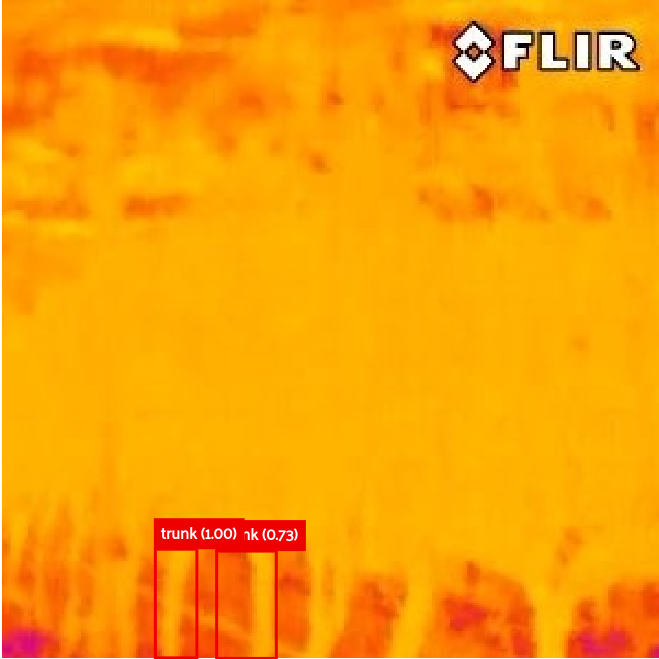}} \hfill
    \subfloat[][ZCU104]{\includegraphics[width=0.24\textwidth]{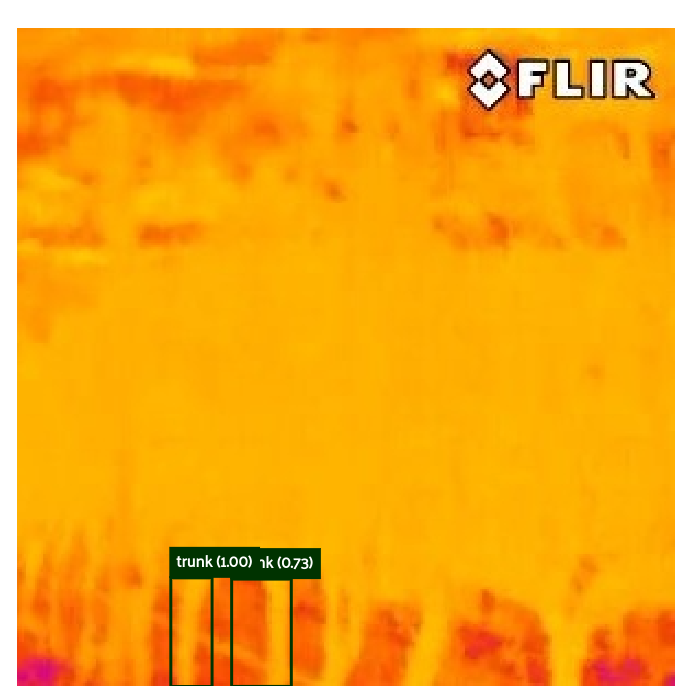}} \hfill
    \subfloat[][TPU]{\includegraphics[width=0.24\textwidth]{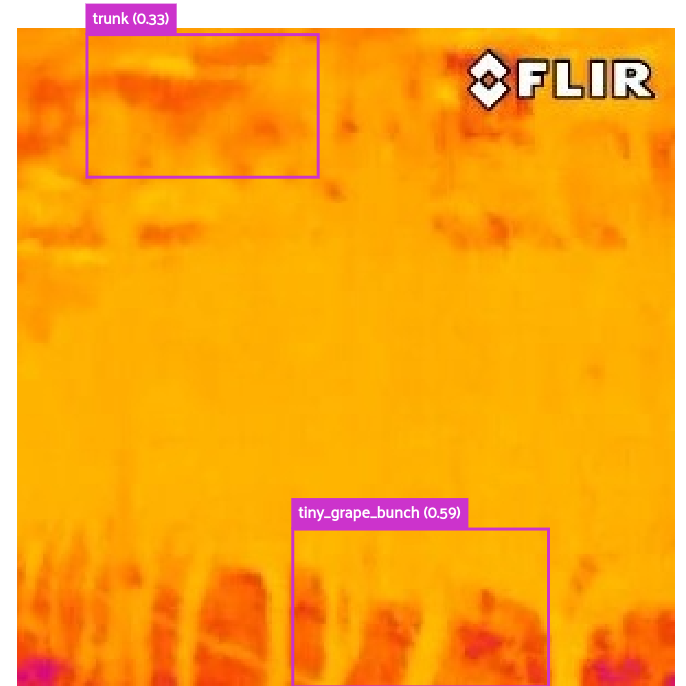}} \hfill
    \caption{Detailed sample image \ref{fig:r9} from figure \ref{fig:image_results} from  Blue -- ground-truth; light green -- NVIDIA RTX3080 TF2; orange -- NVIDIA RTX3090 \ac{tf-trt} \ac{fp32}; brown -- NVIDIA RTX3090 \ac{tf-trt} \ac{fp16}; dark yellow -- NVIDIA RTX3090 \ac{tf-trt} \ac{int8}; red -- AMD-Xilinx Kria KV260; dark green -- AMD-Xilinx ZCU104; pink -- Coral Dev Board \ac{tpu}}
    \label{fig:11i}
\end{figure*}

\begin{figure*}[!htb]
    \centering
    \subfloat[][Ground truth]{\includegraphics[width=0.24\textwidth]{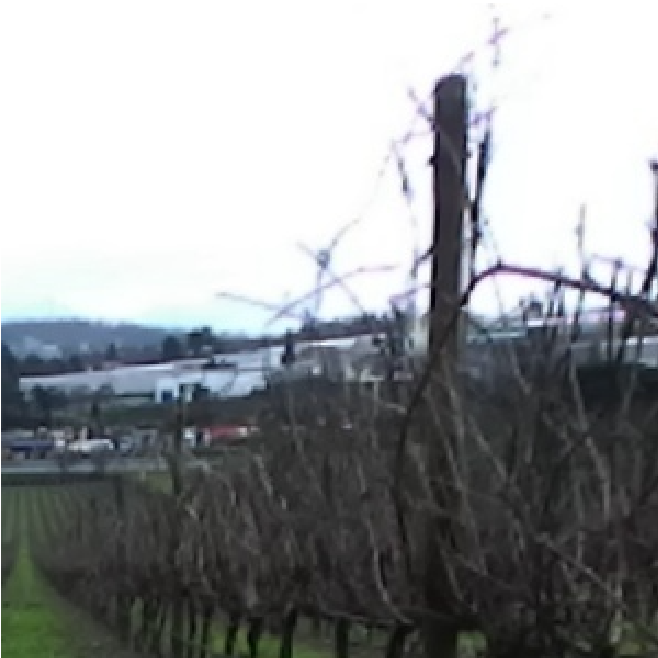}} \hfill
    \subfloat[][RTX3090 TF2]{\includegraphics[width=0.24\textwidth]{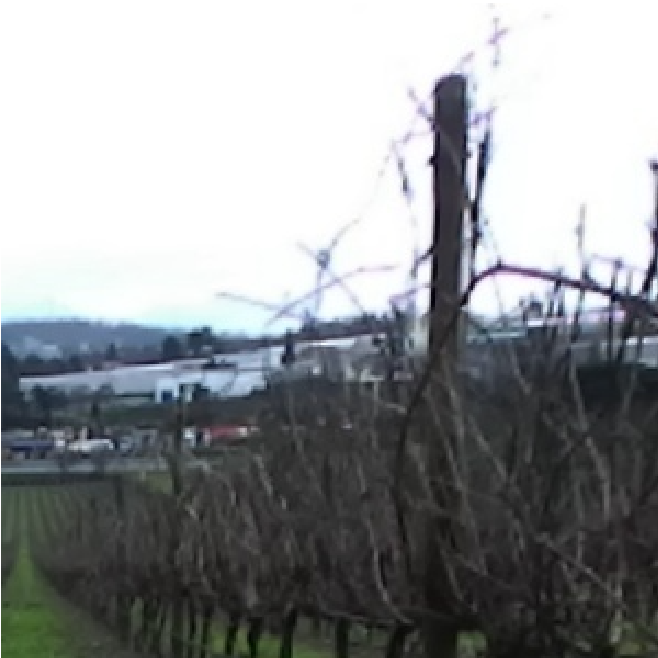}} \hfill
    \subfloat[][TF-TRT FP32]{\includegraphics[width=0.24\textwidth]{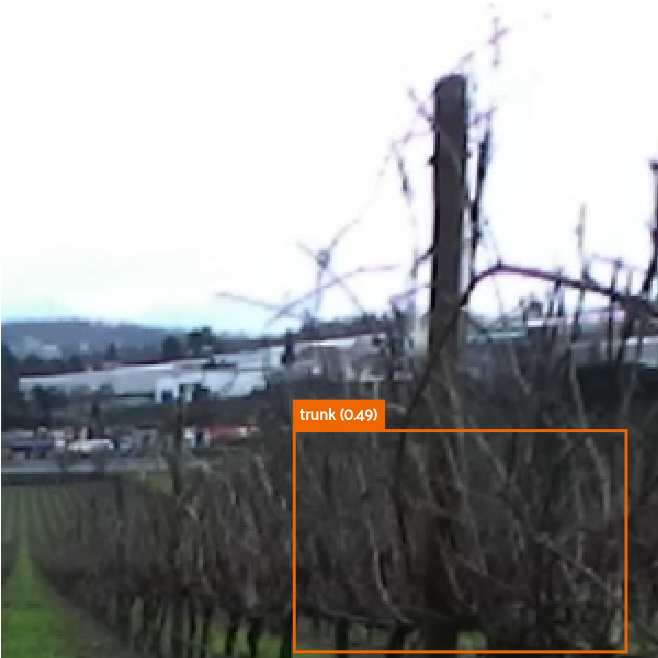}} \hfill
    \subfloat[][TF-TRT FP16]{\includegraphics[width=0.24\textwidth]{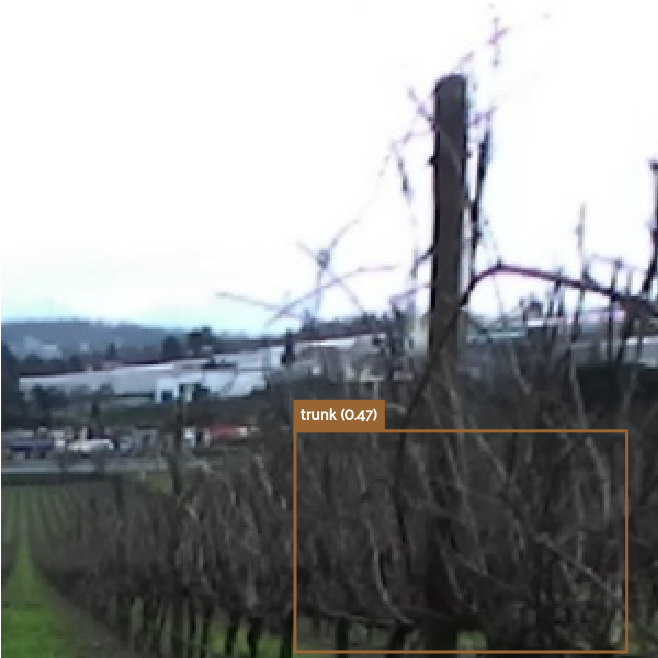}} \hfill
    \subfloat[][TF-TRT INT8]{\includegraphics[width=0.24\textwidth]{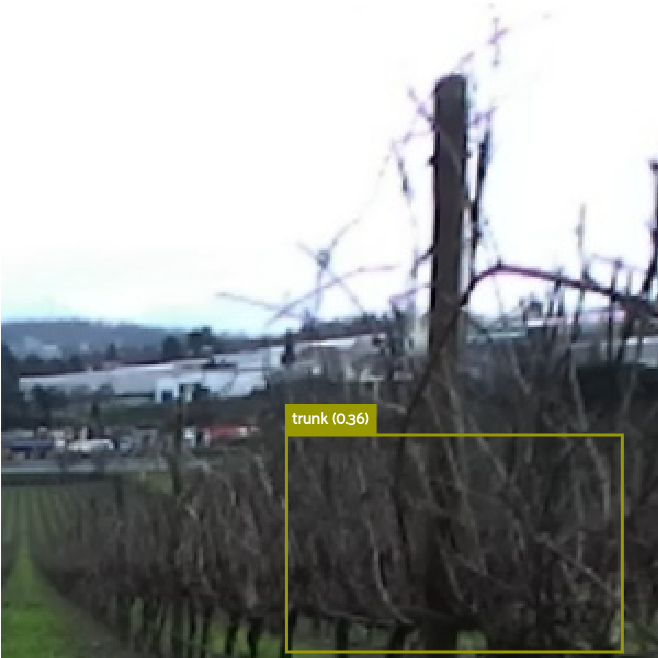}} \hfill
    \subfloat[][KV260]{\includegraphics[width=0.24\textwidth]{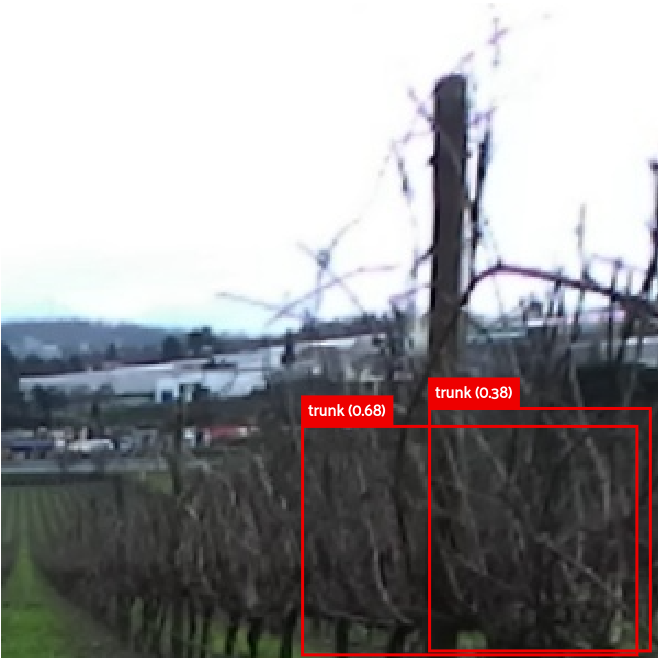}} \hfill
    \subfloat[][ZCU104]{\includegraphics[width=0.24\textwidth]{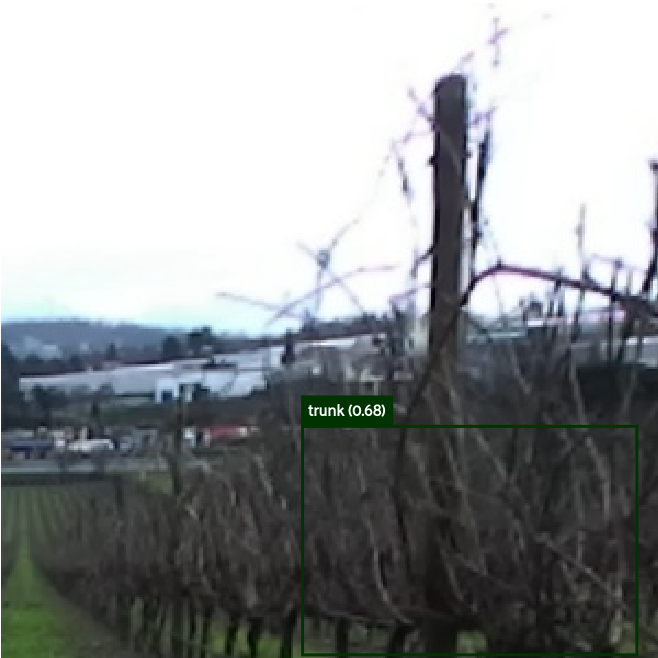}} \hfill
    \subfloat[][TPU]{\includegraphics[width=0.24\textwidth]{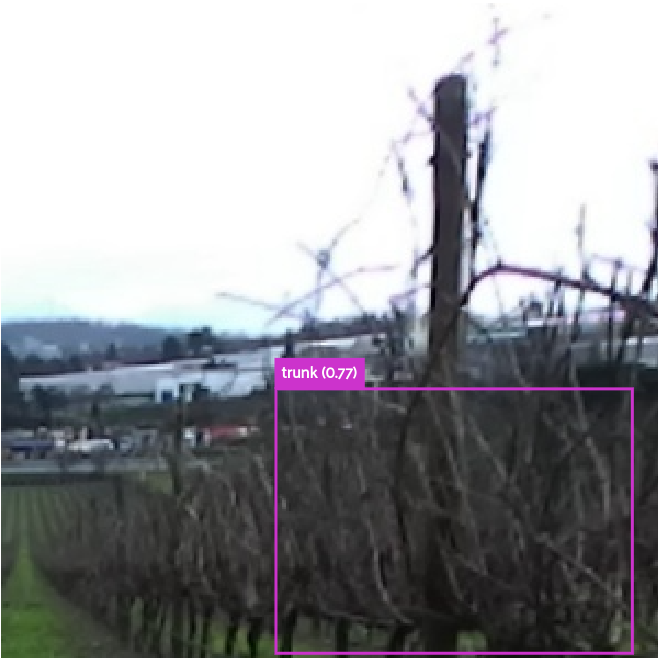}} \hfill
    \caption{Detailed sample image \ref{fig:r10} from figure \ref{fig:image_results} from  Blue -- ground-truth; light green -- NVIDIA RTX3080 TF2; orange -- NVIDIA RTX3090 \ac{tf-trt} \ac{fp32}; brown -- NVIDIA RTX3090 \ac{tf-trt} \ac{fp16}; dark yellow -- NVIDIA RTX3090 \ac{tf-trt} \ac{int8}; red -- AMD-Xilinx Kria KV260; dark green -- AMD-Xilinx ZCU104; pink -- Coral Dev Board \ac{tpu}}
    \label{fig:11j}
\end{figure*}

For better readability of the figure \ref{fig:image_results}, this appendix supplements the same figure with one annotation kind per images in the figures \ref{fig:11a} to \ref{fig:11j}

\end{document}